\documentclass{article}

\usepackage[utf8]{inputenc}
\usepackage[T1]{fontenc}
\usepackage{times}
\usepackage{helvet}
\usepackage{courier}
\usepackage[
  letterpaper,
  textheight=9in,
  textwidth=5.5in,
  top=1in,
  headheight=12pt,
  headsep=25pt,
  footskip=30pt
]{geometry}
\usepackage{booktabs}
\usepackage{array}
\usepackage{amsmath,amssymb,amsfonts}
\usepackage{nicefrac}
\usepackage[expansion=false]{microtype}
\microtypesetup{expansion=false}
\usepackage{xcolor}
\usepackage{fvextra}
\usepackage[most]{tcolorbox}

\usepackage{graphicx}
\usepackage{svg}
\usepackage{placeins}
\usepackage{float}
\usepackage{capt-of}
\usepackage{tikz-cd}
\usepackage{multicol}
\usepackage{url}
\usepackage[hidelinks]{hyperref}

\definecolor{promptframe}{HTML}{4A4A4A}
\definecolor{prompttitle}{HTML}{555555}
\definecolor{promptback}{HTML}{FAFAFA}
\definecolor{examplepromptframe}{HTML}{8DBA8D}
\definecolor{examplepromptback}{HTML}{F1FAF1}

\newtcolorbox{promptquote}[1][Prompt]{%
  enhanced,
  breakable,
  colback=promptback,
  colframe=promptframe,
  coltitle=white,
  colbacktitle=prompttitle,
  fonttitle=\bfseries,
  title={#1},
  boxrule=0.8pt,
  arc=3pt,
  outer arc=3pt,
  left=8pt,
  right=8pt,
  top=8pt,
  bottom=8pt,
  toptitle=2pt,
  bottomtitle=2pt,
  before skip=0.6em,
  after skip=0.6em
}

\newtcolorbox{examplepromptquote}[1][Prompt]{%
  enhanced,
  breakable,
  colback=examplepromptback,
  colframe=examplepromptframe,
  coltitle=white,
  colbacktitle=prompttitle,
  fonttitle=\bfseries,
  title={#1},
  boxrule=0.8pt,
  arc=3pt,
  outer arc=3pt,
  left=8pt,
  right=8pt,
  top=8pt,
  bottom=8pt,
  toptitle=2pt,
  bottomtitle=2pt,
  before skip=0.6em,
  after skip=0.6em
}

\newcommand{\promptlistingtitle}[1]{%
  \par\medskip
  \noindent\colorbox{prompttitle}{%
    \makebox[\dimexpr\linewidth-2\fboxsep\relax][l]{\color{white}\bfseries #1}%
  }%
  \par\nobreak\vspace{-0.25pt}\noindent
}

\title{Towards Human-Level Book-Writing Capability}
\author{Jan Zierstek \quad Matteo Batelic \quad Maya Medjad \quad Tim Schönenberger}
\newcommand{\paperaffiliation}{Pageshift Entertainment}
\newcommand{\papercontact}{research@pageshift-entertainment.ai}
\newcommand{\paperdataseturl}{https://huggingface.co/datasets/Pageshift-Entertainment/LongPage}
\newcommand{\papermodelcollectionurl}{https://hf.co/collections/Pageshift-Entertainment/pagestorm-research-preview}
\date{}

\makeatletter
\renewcommand{\normalsize}{%
  \@setfontsize\normalsize{10}{11}%
  \abovedisplayskip=7pt plus 2pt minus 5pt\relax%
  \belowdisplayskip=\abovedisplayskip\relax%
  \abovedisplayshortskip=0pt plus 3pt\relax%
  \belowdisplayshortskip=4pt plus 3pt minus 3pt\relax%
}
\normalsize

\renewcommand{\section}{%
  \@startsection{section}{1}{\z@}%
  {-2.0ex plus -0.5ex minus -0.2ex}%
  {1.5ex plus 0.3ex minus 0.2ex}%
  {\large\bfseries\raggedright}%
}
\renewcommand{\subsection}{%
  \@startsection{subsection}{2}{\z@}%
  {-1.8ex plus -0.5ex minus -0.2ex}%
  {0.8ex plus 0.2ex}%
  {\normalsize\bfseries\raggedright}%
}
\renewcommand{\subsubsection}{%
  \@startsection{subsubsection}{3}{\z@}%
  {-1.5ex plus -0.5ex minus -0.2ex}%
  {0.5ex plus 0.2ex}%
  {\normalsize\bfseries\raggedright}%
}
\renewcommand{\paragraph}{%
  \@startsection{paragraph}{4}{\z@}%
  {1.5ex plus 0.5ex minus 0.2ex}%
  {-1em}%
  {\normalsize\bfseries}%
}
\renewcommand{\subparagraph}{%
  \@startsection{subparagraph}{5}{\z@}%
  {1.5ex plus 0.5ex minus 0.2ex}%
  {-1em}%
  {\normalsize\bfseries}%
}

\renewcommand{\maketitle}{%
  \begin{center}
    \vspace*{0.6em}
    {\LARGE\bfseries \@title\par}
    \vspace{2.6em}
    {\normalsize\bfseries \@author\par}
    \vspace{1.8em}
    {\normalsize \paperaffiliation\par}
    \vspace{0.45em}
    {\small For inquiries: \href{mailto:\papercontact}{\textit{\papercontact}}\par}
  \end{center}
  \vspace{2em}
}

\renewenvironment{abstract}{%
  \begin{center}
    \begin{minipage}{0.78\textwidth}
    \begin{center}
      {\large\bfseries Abstract}
    \end{center}
    \vspace{0.3em}
}{%
    \end{minipage}
  \end{center}
  \vspace{2.2em}
}
\makeatother

\begin{document}
\maketitle

\begin{abstract}
Large language models are optimized for instruction following and agentic tasks remain poorly aligned with the requirements of high-quality creative writing. We show that a purpose-built creative writing model can outperform both GPT-5.5 and Claude Opus 4.8 on writing quality evaluation. Fiction frequently depends on behaviors that assistant-tuned models are explicitly trained to avoid, particularly deception, moral ambiguity, and unreliable narration. As a result, generated stories often appear structurally correct while remaining stylistically generic, overly explanatory, or weakly grounded in human literary behavior. We present a dataset construction and training framework for book-scale creative writing that reframes supervised fine-tuning as a prompt-to-book generation task grounded in human-authored fiction. Starting from public-domain novels, we derive a multi-resolution Planning Scaffold by summarizing each book at progressively finer levels, from a high-level premise to chapter- and scene-level structure. We then invert this hierarchy during training: the model learns to expand a prompt into increasingly detailed plans and finally into the original human-authored book text. This formulation preserves human prose as the final supervised target while using intermediate summaries to make book-scale generation learnable. We train a long-context language model on these prompt-to-book trajectories and show that this objective shifts generation away from assistant-style prose and toward human literary writing.
\end{abstract}

\footnotetext[1]{\raggedright\setlength{\parskip}{0pt}The dataset is available at \url{\paperdataseturl}.}
\footnotetext[2]{\raggedright\setlength{\parskip}{0pt}The trained model collection is available at \url{\papermodelcollectionurl}.}
\section{Introduction}%
\label{sec:introduction}

Recent language models can generate long and locally coherent text~\cite{yang2022re3,yang2023doc,wang2025dome}, but their outputs often remain recognizably assistant-like even in creative writing settings. Stories generated by instruction-tuned models frequently over-explain character motivations, resolve conflict too directly, or default to safe and predictable interactions. While these models are highly effective at reasoning and task completion, the behavioral patterns encouraged during assistant alignment do not fully match the distributions present in human fiction.

This mismatch becomes particularly visible at book scale. Fiction routinely relies on deception, ambiguity, unreliable narration, and characters acting against the reader's expectations. These behaviors are often undesirable in assistant systems, where models are optimized to be helpful, honest, and direct~\cite{ouyang2022instructgpt}. As a result, models trained primarily for assistant-style interaction may struggle to reproduce the narrative dynamics and prose characteristics commonly found in human-authored books.

Existing work on long-context generation largely approaches this problem through improved memory, retrieval, or planning~\cite{yang2022re3,yang2023doc,wang2025dome,yao2019plan}. These techniques are primarily aimed at maintaining long-range consistency. However, consistency alone does not guarantee human-like creative writing quality. A model can maintain coherent long-horizon structure while still producing prose that feels synthetic, overly assistant-like, or stylistically unlike human-authored fiction. As we show in this paper, training specifically for creative writing can substantially improve writing quality, allowing our model to outperform both GPT-5.5 and Claude Opus 4.8 on writing quality evaluations.

In this work, we introduce a dataset construction and training framework for prompt-to-book generation grounded in human-authored fiction. The central idea is to transform books into a planning scaffold and then reverse that process during training. A planning scaffold is a multi-stage representation of a book constructed through progressively finer summaries, ranging from a high-level book description to chapter- and scene-level structure~\cite{fan2018hierarchical,fan2019strategies,rashkin2020plotmachines,yang2023doc}. Starting from public-domain novels, the pipeline first compresses each book into this scaffold representation. The model is then trained to invert the process: given a prompt, it expands coarse summaries into increasingly detailed representations before generating the original book text.

This formulation treats creative writing as a staged expansion problem rather than a single-step continuation task~\cite{riedl2010narrative,martin2018event,yao2019plan}. The planning scaffold provides supervision for long-horizon generation, while the final target remains the human-authored book itself. The objective is therefore not merely to generate coherent long text, but to align model behavior with the structure and prose distributions present in published fiction.

\subsection{Overview}%
\label{sec:overview}

Section~\ref{sec:dataset} describes the dataset construction pipeline. Starting from public-domain books, we preprocess the source text, generate synthetic prompts, and construct the hierarchical Planning Scaffold from progressively finer summaries of each book.

Section~\ref{sec:training} describes the supervised fine-tuning setup. The model is trained to reproduce the full prompt-to-book trajectory, generating the Planning Scaffold before producing the original human-authored chapter text.

Section~\ref{sec:evaluation} presents the main evaluation results. We compare the model against frontier general-purpose language models, including GPT-5.5 and Claude Opus 4.8, on writing quality evaluations, and analyze how well the model follows both the original prompt and the generated Planning Scaffold.

Section~\ref{sec:interpretability} analyzes whether the trained model uses the Planning Scaffold in the intended way. We study attention patterns during generation and use sparse autoencoder features to examine internal representations associated with narrative structure.

Section~\ref{sec:conclusion} summarizes the main findings and limitations. In particular, we discuss the transfer of the learned planning procedure beyond public-domain books, the remaining challenge of long-context consistency, and future work on improving prompt and scaffold following.
\section{Dataset}\label{sec:dataset}

This section describes the construction of the dataset used for supervised book-scale generation. The source books provide the final prose targets, while the prompts, intermediate plans, and metadata used for training are generated from the books themselves. The construction problem is therefore inverse to the generation problem: starting from a complete book, the pipeline recovers the structured information that a model should later learn to generate before writing the book.

The resulting training examples contain three main components: a synthetic user prompt, an intermediate planning scaffold, and the original book text. The prompt specifies the requested book, the scaffold exposes the latent narrative structure, and the book text remains the final human-authored target. The following subsections describe the corpus, the annotation strategy, and the hierarchical processing pipeline used to produce these representations.

\subsection{Corpus Construction}\label{sec:corpus-construction}

The source corpus consists of public-domain books from Project Gutenberg~\cite{gutenberg}. The final release contains approximately 6{,}000 books.

The corpus is constructed in two stages. The first stage contains a 300-book seed set drawn from the global top-300 Project Gutenberg downloads at the time of collection. The second stage adds 5{,}700 further books. Both stages produce the same final representation and training format. The difference lies in how the annotations are generated.

\subsection{Annotation Strategy}\label{sec:annotation-strategy}

Stage one processes all 300 seed books with a prompted Qwen3\mbox{-}32B model~\cite{qwen3technicalreport}. The model is used as a reasoning system throughout the pipeline, producing the intermediate representations later used for training. These outputs additionally serve as supervision for distilling a faster model specialized for the repeated low-level stages of the pipeline.

Stage two replaces the scene-level and chapter-level components with a distilled Qwen3\mbox{-}14B tool model~\cite{qwen3technicalreport} trained from the stage-one outputs. Unlike the Qwen3\mbox{-}32B setup, the distilled model operates without reasoning, making it substantially faster to run.

This separation is motivated by computational cost. Most processing effort occurs at the scene and chapter levels because these operations must be repeated many times within each book. In contrast, the global processing stages are invoked only a few times per example. Stage two therefore continues to use the prompted Qwen3\mbox{-}32B reasoning model for the higher-level abstractions and metadata generation steps, while the distilled Qwen3\mbox{-}14B model handles the repeated local processing.

\subsection{Pipeline}\label{sec:pipeline}

The pipeline progressively converts raw book text into scene-level, chapter-level, and book-level representations. Each level captures a different scale of narrative information. Scene-level processing preserves local events and narrative function. Chapter-level processing links scenes into larger developments and records how information is distributed across a chapter. Book-level processing compresses the chapter representations into global narrative structure.

A central design choice is that summaries throughout the pipeline are represented as bullet-point lists rather than prose paragraphs. The objective is not to produce polished standalone summaries, but to preserve narrative facts in a form that can later be aggregated and recombined. The pipeline therefore biases the model toward producing multiple short bullet points instead of dense free-form text. A typical bullet point contains roughly 10--20 words, with an absolute upper limit of 45 words.

At the scene level, the representation captures both content and narrative role. In addition to local events, the pipeline records information such as which characters are central to the scene, how the narration is structured, and whether the focus is placed on action, exposition, dialogue, or changes in pacing. The objective is to preserve details that are often lost in conventional summarization.

At the chapter level, the scene representations are aggregated into a larger structural unit. The chapter representation links together the local developments of individual scenes while preserving broader stylistic and narrative balance across the chapter.

Finally, the book-level stage compresses the chapter representations into a global representation of the narrative. This representation is then used to generate metadata and synthetic prompts. The resulting structure acts as an intermediate planning scaffold that makes the latent organization of long-form narratives explicit.

Dataset construction therefore runs from completed book text to prompts and planning structure. During training, the direction is reversed: the model is conditioned on a synthetic user prompt and learns to generate the scaffold before producing the final book text. This preserves human-authored prose as the final target while still providing explicit supervision for long-range planning.

\begin{center}
\includegraphics[width=\textwidth]{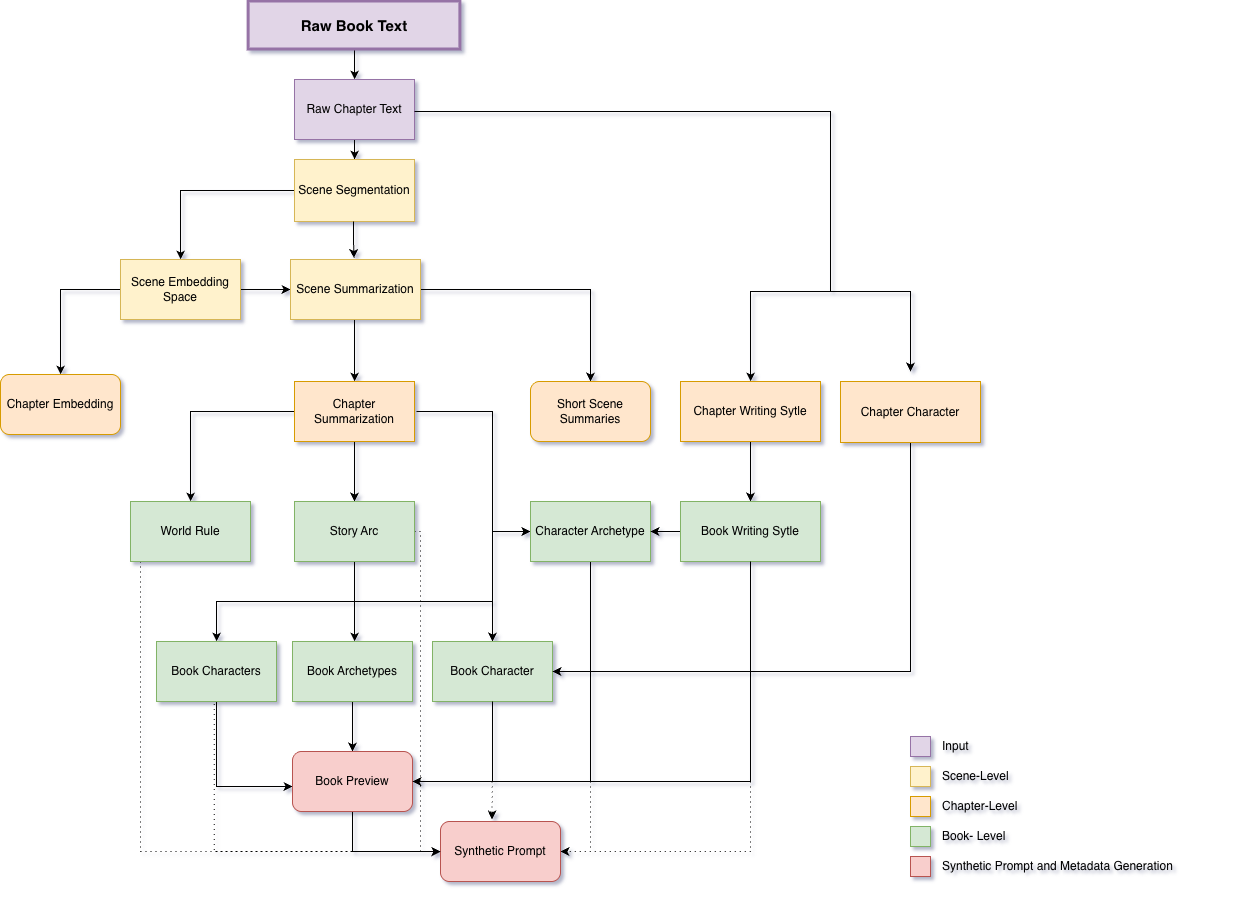}
\captionof{figure}{The figure illustrates the transformation of raw book text into a hierarchical planning scaffold through scene-level, chapter-level, and book-level processing stages. Colors denote the processing level of each component, while arrows indicate the flow of information between extracted representations. The diagram also highlights that later book-level and synthetic prompt components are constructed from multiple intermediate outputs rather than from a single linear summarization path.}
\end{center}

\subsubsection{Scene-Level Processing}
\paragraph{Scene Segmentation}

Scene-level processing provides the first structural layer of the pipeline. Rather than treating a chapter as one continuous block of text, the pipeline divides it into smaller narrative units that can be analyzed and summarized more precisely. This is important because chapters often contain shifts in setting, time, focal character, character grouping, dialogue focus, point of view, or narrative purpose.

Scene segmentation is guided by these narrative criteria, but they are not used as rigid deterministic rules. Instead, they define what the model should consider when identifying scene boundaries. This allows the pipeline to handle cases where a transition is implied by narrative movement rather than explicit formatting.

The result is a structured scene breakdown for each chapter. Each scene is assigned a short name, a text span, a narrative focus, and a narrative perspective. This makes the chapter easier to process in later stages, because each scene can be scored, summarized, and aggregated as a distinct narrative unit.
\paragraph{Scene Schema}

The scene schema preserves both the position of the scene inside the chapter and the basic narrative information needed for later processing. Each scene includes a short descriptive name, its textual span, the dominant narrative focus, and the narrative perspective. The narrative focus identifies the character or narrator through whom the scene is mainly presented, while the narrative perspective describes how access to that viewpoint is framed.

Once the scene boundaries are validated, the corresponding scene text and word count are attached. This turns each scene into a self-contained unit for later scoring, summarization, and chapter-level aggregation.

\paragraph{Scene Embedding Space}
A plain text summary is useful for describing what happens in a scene, but it does not always preserve the scene’s functional properties. Some narrative features are easy to underrepresent in prose summaries, especially when they are diffuse rather than event-like. For example, pacing, exposition density, world-building intensity, or the amount of dialog may shape the reading experience even if they are not naturally stated as explicit plot points.

To preserve these properties, each scene is assigned a seven-dimensional narrative score vector. The dimensions are action, dialog, world building, exposition, romantic content, erotic content, and pacing. Each dimension is scored from 0 to 100, indicating how strongly that feature is present in the scene. These scores act as interpretable control signals rather than neural embeddings: they make visible what kind of narrative work the scene performs.

This representation complements the textual scene summary. The summary captures the concrete events, while the score vector captures broader structural qualities. A scene can therefore be represented not only as “what happened,” but also as whether it is dialog-driven, exposition-heavy, fast-paced, action-oriented, or primarily used for world building. Later stages use this information when generating chapter summaries and when aggregating scene-level information into chapter-level representations.

To make the scores more stable, the seed pipeline uses an ensemble-style procedure. The same scene is scored multiple times by a reasoning model, and the final value for each dimension is computed as the mean across these generations. This is useful because reasoning models may arrive at slightly different scores across runs, since each generation can produce a different internal reasoning trace before assigning values.

Averaging these outputs reduces the influence of individual noisy judgments. Very low values are then thresholded to zero, which further suppresses weak or inconsistent signals and produces a cleaner representation of the dominant narrative features in each scene.

\paragraph{Scene Summarization}
Scene summarization is not treated as a generic text-compression step. Important structural signals, such as shifts in tension, exposition, pacing, dialogue focus, or world-building, may be compressed away or treated as secondary details if the model is asked only to summarize the scene in general terms.

Instead, each annotated scene is converted into a compact natural-language summary guided by its narrative score vector. The score vector provides an additional signal about the function of the scene within the chapter, helping the summary emphasize the aspects that are most relevant for later aggregation.

The resulting summaries are concise bullet-style descriptions. They preserve enough local information to support chapter-level processing, while avoiding the need to reconsider the full scene text at every later stage.

\subsubsection{Chapter-Level Processing}

After the scene-level representation has been created, the pipeline moves from local narrative units to the chapter level. A chapter is not treated as a simple concatenation of scenes, but as a larger narrative unit with its own structure, emphasis, character dynamics, and stylistic properties. The goal of chapter-level processing is therefore to compress the detailed scene information into a representation that is compact enough to be used at the book level, while still preserving the main developments of the chapter.

The chapter-level stage builds on the scene summaries, scene score vectors, and scene metadata produced in the previous stage. It aggregates these local signals into chapter summaries, chapter-level narrative scores, short scene mappings, character information, and writing-style descriptions. This creates an intermediate layer between detailed scene analysis and global book planning.

\paragraph{Chapter-Level Writing Style}
The motivation for extracting chapter-level writing style is to represent narrative text not only by what happens, but also by how it is written. Event-based summaries capture plot progression, characters, and world information, but they discard stylistic signals that strongly influence perceived authorial voice, readability, atmosphere, and genre convention.

Therefore, a separate writing-style representation is introduced to preserve prose-level characteristics independently from narrative content. This enables the dataset to support tasks where stylistic consistency is important, such as style-controlled generation, authorial voice modeling, chapter retrieval, and comparative analysis of narrative form.

By abstracting away names, locations, and plot-specific details, the extracted style descriptors function as reusable stylistic fingerprints rather than summaries of chapter events.

\paragraph{Chapter Embedding}
The chapter embedding summarizes the narrative profile of the chapter as a whole. It is derived from the scene-level score vectors and uses the same dimensions. This keeps the representation consistent across levels of the hierarchy.

The purpose of the chapter embedding is to describe the dominant narrative character of the chapter. For example, a chapter may be largely expository, dialogue-driven, action-heavy, or focused on world building. While the scene-level scores describe local variation, the chapter-level score captures the overall balance of these features across the chapter.

This allows later stages to compare chapters not only by what happens in them, but also by how they function structurally within the book. A chapter with high exposition may serve a different planning role than a chapter with high action or high dialogue, even if both are similar in length.

\paragraph{Chapter Summary Generation}
 Instead of summarizing the raw chapter text directly, the pipeline builds on the scene summaries, which already identify the main events and narrative functions. The summary length is scaled to the amount of scene-level detail, so more complex chapters receive proportionally richer summaries.

These chapter summaries then serve as compact inputs for book-level processing, including story arc detection, world-rule extraction, and character analysis. In this way, chapter-level summaries preserve the main developments of the chapter without requiring later stages to process the full chapter text again.

\paragraph{Short Scene Summaries}
The pipeline also produces shorter scene summaries that connect the chapter-level summary back to the individual scenes. Their purpose is not to replace the detailed scene summaries, but to preserve the alignment between the chapter’s overall development and its local scene structure.

This intermediate layer is useful because the chapter summary compresses the narrative into a higher-level form, while the full scene summaries may contain more detail than later stages need. Short scene summaries keep the scene sequence visible in a compact way, making it easier to relate chapter plans to the scenes that support them.

\paragraph{Chapter Characters}
The chapter-level pipeline also extracts character information to preserve who is active in each part of the book. This is important because character relevance is not constant across a narrative. A character may drive one chapter, support another, or only be mentioned indirectly, while still being important for the larger story.

By distinguishing between main characters, side characters, and mentioned characters at the chapter level, the pipeline captures how characters enter, leave, and influence the narrative over time. This gives later book-level processing a more reliable basis for identifying structurally important characters and constructing coherent character descriptions across the full book.

\subsubsection{Book-Level Processing}
After chapter-level processing, the pipeline moves to a global representation of the book. The goal is no longer to summarize individual events, but to recover the higher-level structure that organizes the narrative as a whole. This includes story arcs, central character functions, world rules, writing style, and the book’s narrative archetype.

This layer is important because long-form generation requires global coherence. Characters need stable roles, conflicts need a larger trajectory, and style should remain consistent across chapters. The book-level representation therefore acts as a planning layer that connects chapter-level details into a unified narrative structure.
\paragraph{Story Arc Detection}
The pipeline groups chapter-level developments into a small set of broader story arcs. For each arc, the model generates a synthetic arc name and a compact bullet-point progression. The arc name provides a high-level label, while the bullet points describe how conflicts, goals, relationships, or turning points develop across chapters.

This representation makes long-range narrative movement explicit without turning the arcs into another detailed chapter summary. Each arc is therefore kept short, so it can function as a book-level planning signal.

\paragraph{Character Archetypes}
Character archetypes describe the functional role of characters within the narrative. This is different from measuring how much space a character occupies in the text. A character may appear often without shaping the central structure, while another may appear less frequently but still define the main conflict, motivation, or turning point of the story.

The pipeline therefore abstracts characters by their narrative function and by their relationships to other characters. These cross-character archetype relationships capture roles such as opposition, support, mentorship, rivalry, dependence, or emotional contrast. This helps represent the character system as a structured network of functions rather than as a simple list of frequently appearing names.

\paragraph{Book Character List}
The book-level character list is constructed from the character information extracted at the chapter level. Local character appearances are consolidated into a stable representation. The book-level representation keeps only main and side characters, so that the character list remains focused on characters with a sustained role in the book.

For each retained character, the pipeline generates a compact bullet-point profile that summarizes recurring traits, relationships, motivations, and developments. These profiles provide an accessible lookup representation during later generation, so that character information does not need to be reconstructed from longer narrative summaries.

\paragraph{Book Writing Style}
To preserve prose-level information, the pipeline consolidates the writing-style analyses extracted for individual chapters. Each chapter contributes local evidence about the prose form of the text, and these chapter-wise descriptions are aggregated into a single style profile for the full book.

The resulting representation is a flat list of approximately 35 bullet points, each describing a recurring stylistic property of the book. It captures prose-level patterns such as tense, diction, syntax, dialogue handling, punctuation, figurative language, narrative voice, register, spelling conventions, and tone.

This step is decoupled from plot-oriented processing. Its input consists of chapter-level style analyses rather than story arcs, character descriptions, or world rules. The objective is therefore not to summarize narrative content, but to preserve the stylistic regularities that govern how the text is written across chapters. This provides later generation with a compact style representation grounded in observed prose patterns rather than event-level summaries.

\paragraph{Book Archetype}
At the book level, the pipeline produces a compact abstraction of the narrative pattern underlying the text. It is derived from book-level signals such as story arcs, chapter summaries, and writing style, and maps them into a higher-level description of the book’s structural form.

The motivation for this step is that long-form generation depends not only on event order, but also on the broader narrative logic that organizes those events. The model needs an early signal of what kind of story is being generated, which conventions shape it, and what type of resolution the structure implies.

The archetype is therefore written as a short abstract paragraph rather than as another content summary. It captures dominant narrative modes, structural expectations, transformation patterns, ending shape, and trope orientation, while avoiding concrete plot details. This gives later generation a compact frame for interpreting the more detailed arcs and chapter plans.

\subsubsection{Synthetic Prompt and Metadata Generation}

\paragraph{Book Preview}
The data pipeline processed book-level information into a user-facing short book preview. The preview is user-facing: it gives a compact, spoiler-controlled summary of the book and states, in direct form, what the user can and cannot expect from it.

The preview is represented by three components: a synthetic title, a highlight, and exactly seven tags. The synthetic title assigns a concise name to the book. The highlight describes the premise, conflict, stakes, and narrative hook in approximately 90--130 words, or 3--5 sentences.

The seven tags are short descriptors of the book's general theme or genre. Tags are ordered from most to least important and impactful for the book.

\paragraph{Synthetic Prompt Generation}
A central component of the pipeline is the generation of synthetic user prompts. A primary challenge in this stage is that prompts produced by large language models without explicit control mechanisms often exhibit limited diversity. In practice, unconstrained generation tends to converge toward similar phrasing patterns, comparable levels of detail, and recurring structural formats. This results in a collapse of the synthetic prompt distribution and reduces its ability to approximate the variability observed in real user instructions.

To address this limitation, we introduce \textit{Structured Guide Sampling}. Rather than allowing the language model to determine the prompt form autonomously, the pipeline first samples a prompt profile from a predefined set of stylistic dimensions. These dimensions do not specify the narrative content of the target book; instead, they govern the expected form of the generated prompt, including its length, phrasing style, structural organization, specificity, and surface quality. The sampled profile is then combined with the book-level representation and provided to the language model, ensuring that the resulting prompt remains semantically aligned with the target book while adhering to explicitly controlled stylistic constraints.

This procedure is closely related to recent work on attribute-conditioned synthetic data generation, where control variables are specified or sampled before generation rather than relying on the model's implicit diversity. For example, AttrPrompt generates synthetic data from attributed prompts containing dimensions such as length and writing style, while TinyStories and SimpleStories use randomly sampled or parameterized prompt features to control properties of generated stories~\cite{yu2023attrprompt,eldan2023tinystories,finke2025simplestories}. Similar conditioning appears in synthetic prompt generation as well, such as SynPO, which generates prompts from sampled keyword constraints; in contrast, our method samples dimensions of the user prompt form itself, including length, phrasing style, structure, specificity, and surface noise~\cite{dong2024synpo}.

The sampled dimensions capture multiple aspects of prompt variation, including prompt length, request phrasing, structural layout, and realistic surface-level imperfections. Consequently, the pipeline can generate prompts ranging from short and informal requests to detailed and highly structured writing specifications. The framework additionally controls whether prompts are expressed as free-form prose, lists, or field-based specifications.

To further improve realism, the pipeline incorporates controlled noise patterns, including minor spelling mistakes, grammatical inconsistencies, and punctuation irregularities. These perturbations are applied exclusively to the surface form of the prompt and do not modify the underlying book representation. Their purpose is to prevent the synthetic distribution from consisting solely of polished and idealized instructions, thereby better approximating the irregularities present in real-world user inputs.

All prompt-style dimensions are sampled according to fixed probability distributions. As a result, the generated dataset contains a controlled mixture of prompt regimes, consisting of approximately 30\% short realistic prompts, 60\% medium-length diverse prompts, and 10\% long structured prompts. This explicitly shaped distribution reduces reliance on the intrinsic variability of the language model and improves coverage across a broader range of possible request forms.

The final synthetic prompt is therefore derived from two complementary information sources. The book representation provides semantic grounding, including plot, characters, setting, writing style, and chapter structure, while the sampled prompt profile determines how this information is expressed. The language model must synthesize both components into a single prompt that is semantically consistent with the target book while conforming to the sampled stylistic constraints.

Overall, this procedure produces synthetic training prompts that are semantically grounded, structurally diverse, and more representative of the range of instructions encountered in long-form text generation settings.

\subsubsection{Corpus Scale and Token Composition}

The corpus is characterized in terms of token-level properties of complete sequence representations. Each sequence is decomposed into three components: a synthetic user prompt $p$, an intermediate planning scaffold $s$, and the target book text $b$. Token counts are computed using the Llama~3 tokenizer. For a sequence $i$, the corresponding lengths are denoted by $p_i$, $s_i$, and $b_i$, with total length
\[
 \ell_i = p_i + s_i + b_i.
\]
This decomposition enables separate analysis of conditioning, intermediate structure, and generated content.

\begin{figure}[t]
\centering
\includegraphics[width=\textwidth]{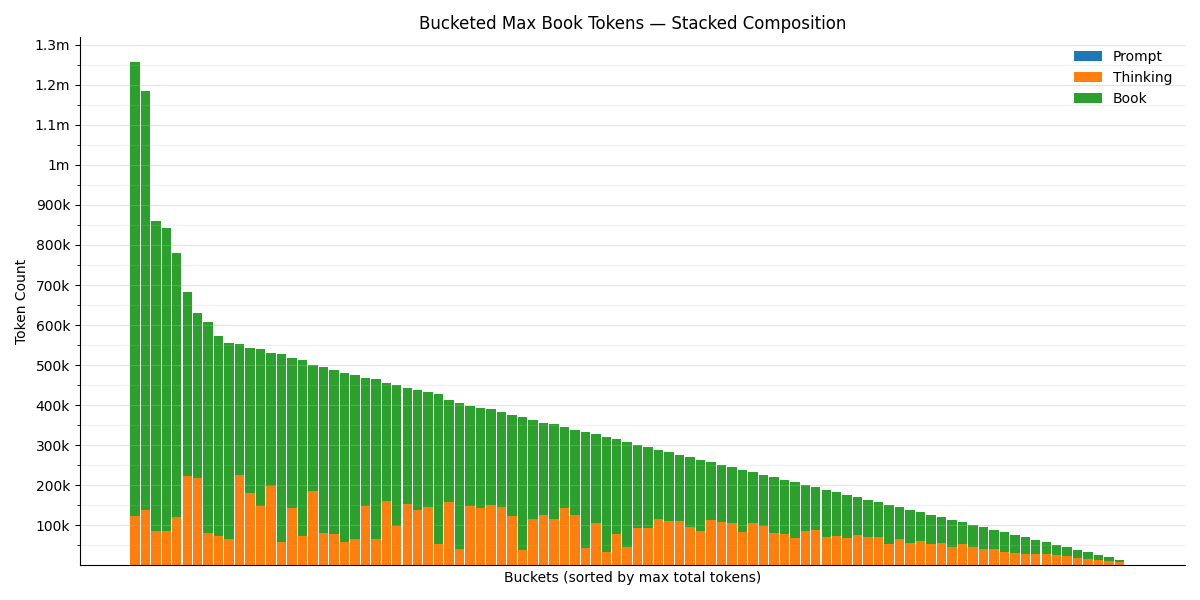}
\vspace{0.75em}
\includegraphics[width=\textwidth]{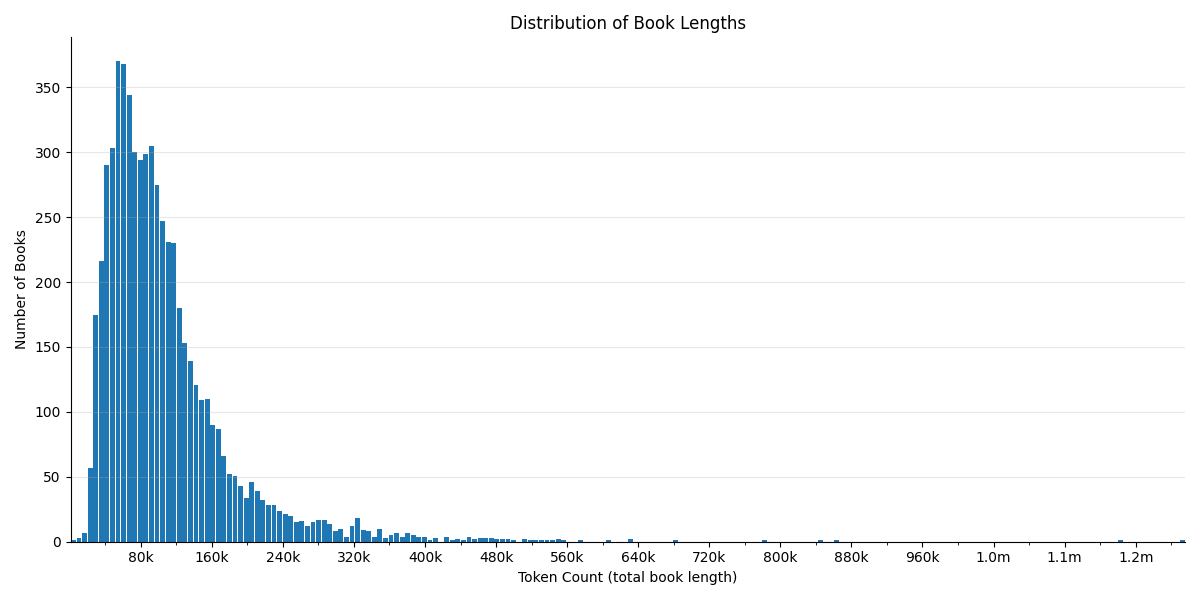}
\caption{Token-level characterization of the corpus sequence. Top: upper-envelope token composition across length buckets, separating prompt, planning scaffold, and book text. Bottom: histogram of total sequence lengths.}\label{fig:corpus_statistics}
\end{figure}

Figure~\ref{fig:corpus_statistics} summarizes the token-level scale and composition of the corpus. The lower panel shows the distribution of total sequence lengths across 6,000 books. In aggregate, the corpus contains 649.1M tokens under the Llama~3 tokenizer. Total sequence lengths range from 1,085 to 1,255,497 tokens, with a mean of 106,940 tokens and a median of 88,983 tokens. The distribution is right-skewed: the 90th, 95th, and 99th percentiles are 187,346, 245,090, and 390,604 tokens, respectively. While most books fall below 100k tokens, 2,512 books exceed this threshold, 257 exceed 262k tokens, 20 exceed 500k tokens, and 2 exceed one million tokens.

The upper panel reports the corresponding token composition across the length spectrum. Each bar corresponds to the maximum-length sequence within a fixed-width total-length bin and is decomposed into synthetic user prompt, reasoning scaffold, and book-text tokens. Across the full corpus, book text accounts for 60.43\% of all tokens, while the reasoning scaffold accounts for 39.35\%; the synthetic user prompt contributes only 0.23\%. At the per-book level, the median shares are 56.60\% book text, 43.08\% reasoning scaffold, and 0.16\% synthetic prompt. Thus, although book text forms the largest component, the reasoning scaffold constitutes a substantial fraction of the total sequence budget, whereas the synthetic prompt is negligible by comparison.

At larger sequence lengths, the absolute contribution of both book text and planning tokens increases proportionally, while their relative proportions remain approximately stable. This indicates that increases in total sequence length are primarily driven by longer book content rather than changes in prompt or scaffold size.

Across the selected sequences, the book text constitutes the majority of tokens, particularly at larger scales. The planning scaffold contributes a consistent fraction of the total token budget, while the synthetic prompt remains comparatively small. These results indicate that, in addition to long-form content, a non-negligible portion of each sequence is allocated to structured intermediate representation.

Overall, the corpus exhibits (i) a heavy-tailed distribution of sequence lengths and (ii) a structured token composition in which planning information occupies a measurable share of the total context.

\FloatBarrier{}

\section{Training the Model}\label{sec:training}

\subsection{Model Choice}\label{sec:model-choice}

We initialize from Ministral 3 14B Base~\cite{ministral3paper,ministral3base}. Since the target task is text-only, we remove the vision encoder and train only the language model.

We use a base model rather than an instruction-tuned model because general-purpose assistant models are not naturally aligned with purely creative writing. Assistant-style post-training encourages models to behave as truthful, honest, non-deceptive, and helpful respondents~\cite{ouyang2022instructgpt}. These objectives are appropriate for question answering and task assistance, but they can conflict with core requirements of fiction generation. A fiction model must be able to invent events that are not true, maintain fictional world states, write unreliable narrators, and generate characters who lie, manipulate, misunderstand, or conceal information. In these settings, fabrication and deception are not failures; they are narrative devices.

This mismatch also affects narrative form. Assistant-tuned models tend to answer requests directly, explain ambiguity, and avoid misleading the user. Long-form fiction often requires withholding information, delaying resolution, preserving ambiguity, sustaining voice, and allowing characters to act under false beliefs. If assistant priors are too strong, the model can collapse fictional narration toward explanatory prose, moral commentary, or over-literal compliance with the prompt. Starting from a base model reduces the amount of assistant-specific behavior that the creative-writing training must overwrite.

Nevertheless, we observe evidence that the base checkpoint still contains substantial instruction-following behavior. After our book-writing SFT stage, the model still achieves roughly 50\% accuracy on GSM8K when prompted with an instruction-style template~\cite{cobbe2021gsm8k}. This was unexpected for a model trained only on the serialized book-writing scaffold and suggests that the base checkpoint may have seen instruction-like data during mid-training. We also observe a qualitative failure mode in historical fiction prompts: the model sometimes inserts real historical figures even when they are not useful for the story. We hypothesize that both behaviors are downstream effects of residual instruction-style training signals in the base model. In future iterations, we plan to either start from an older base model with less instruction-following behavior or train a new text-only base model with tighter control over the training data.

\subsection{Dataset Representation}\label{sec:training-data-representation}

Each training example is represented as a single serialized long-form generation trajectory. The example begins with a user prompt and then expands into a hierarchy of supervised planning and prose fields. Rather than representing the dataset as a direct mapping from a prompt to a complete book, we represent each example as a structured scaffold that decomposes the book-generation process into book-level planning, early first-chapter realization, repeated chapter planning, and chapter-level prose generation.

\FloatBarrier{}
\begin{figure}[!htbp]
\centering
\begin{minipage}[t]{0.24\linewidth}
  \centering
  \includegraphics[width=\linewidth]{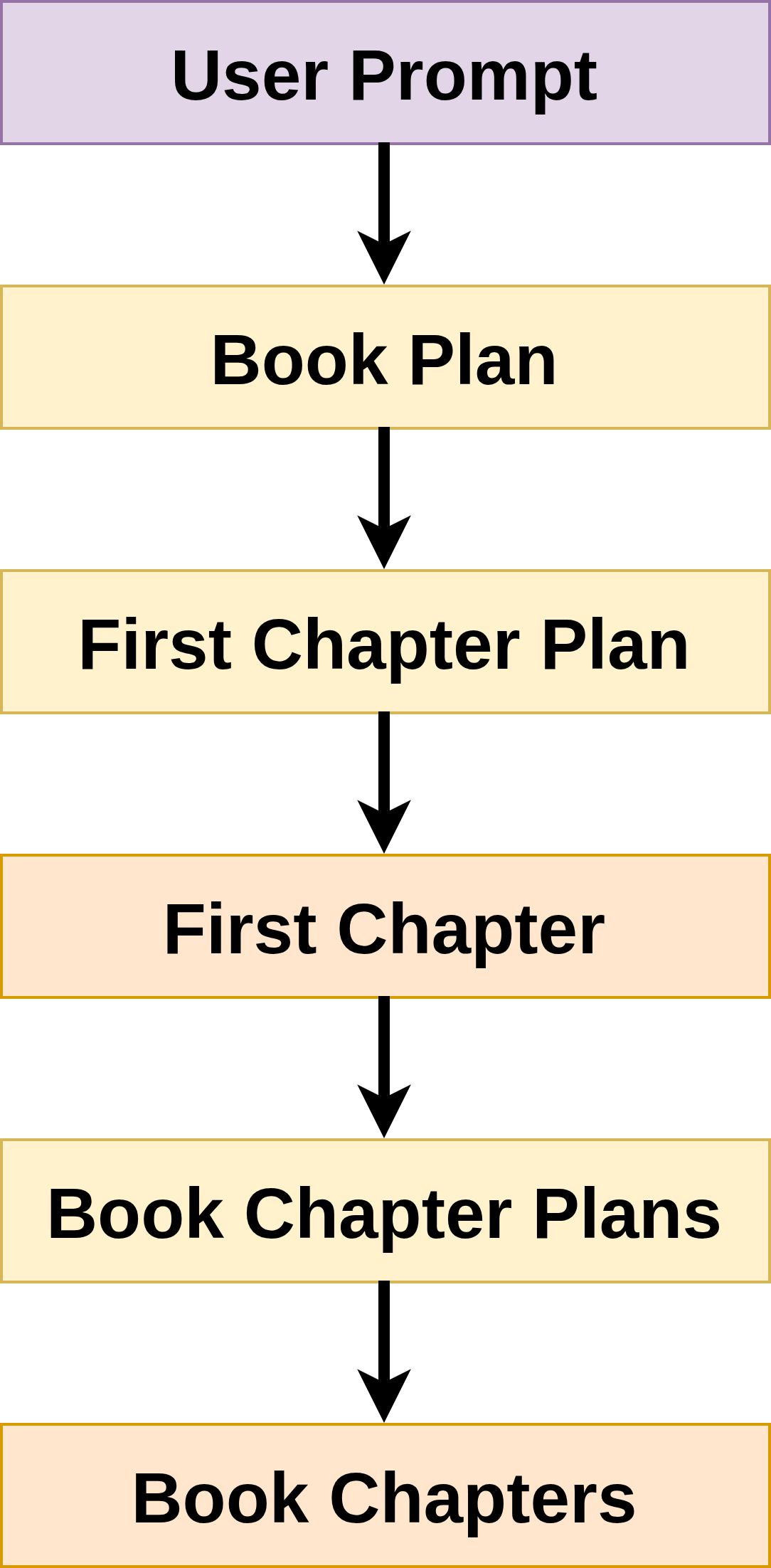}

  \vspace{0.25em}
  \small\textbf{(a)} Overview
\end{minipage}\hfill
\begin{minipage}[t]{0.34\linewidth}
  \centering
  \includegraphics[width=\linewidth]{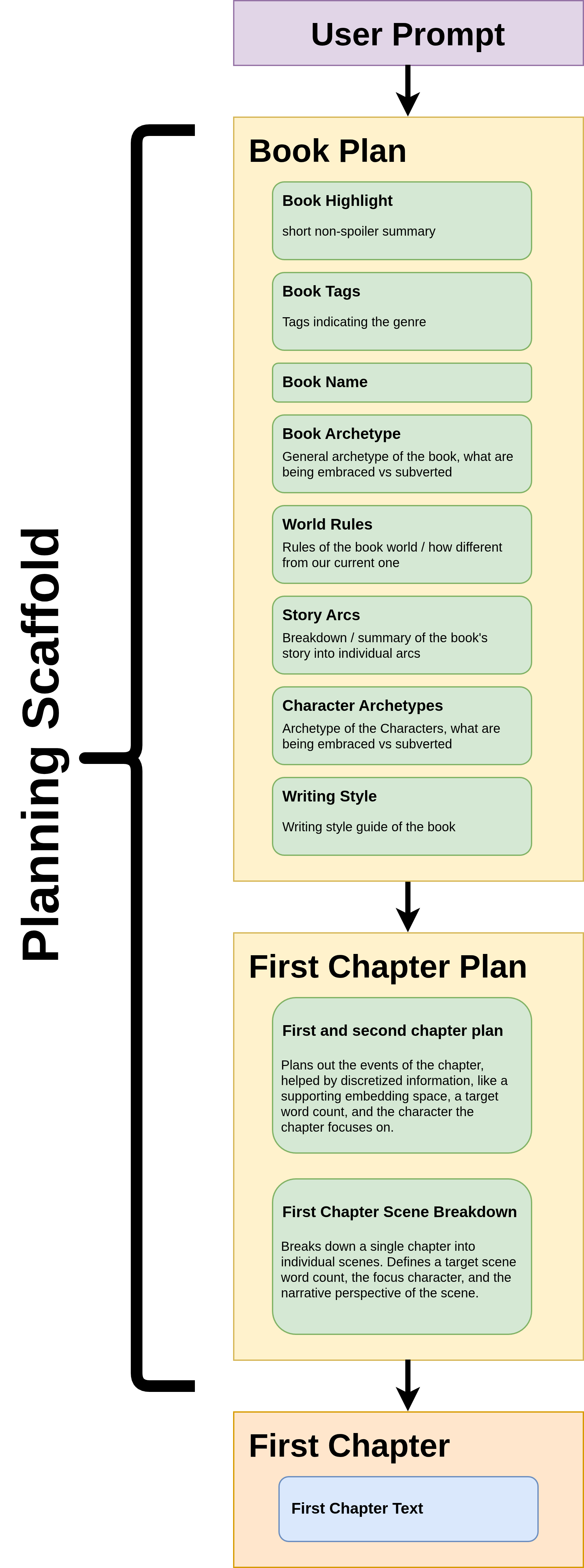}

  \vspace{0.25em}
  \small\textbf{(b)} Book and first-chapter planning
\end{minipage}\hfill
\begin{minipage}[t]{0.34\linewidth}
  \centering
  \includegraphics[width=\linewidth]{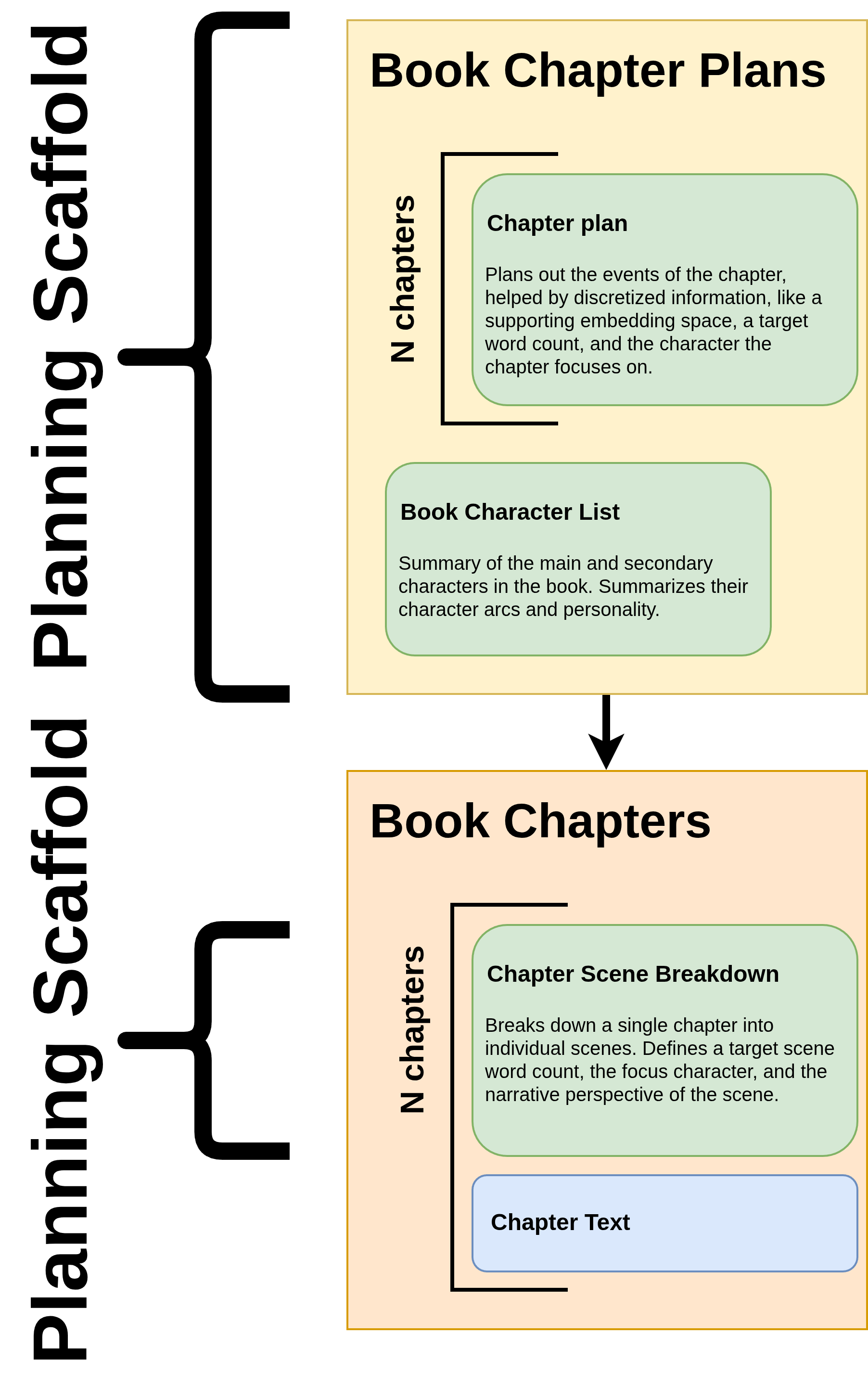}

  \vspace{0.25em}
  \small\textbf{(c)} Later chapter planning
\end{minipage}

\caption{Hierarchical structure of a composed training example. Panel (a) shows the top-level serialized order. Panel (b) expands the book-planning and first-chapter realization stages. Panel (c) expands the repeated later-chapter planning and generation stages.}\label{fig:component_diagram}
\end{figure}
\FloatBarrier{}

A central design choice is the inclusion of an \emph{Early First Chapter} stage in panel~(b). The first chapter text is generated as early as possible, before the remaining long-running generation stages. This provides an early quality-control point for the sample. If the generated first chapter is poor, later stages such as repeated chapter planning and later chapter generation can be skipped, avoiding unnecessary compute expenditure on low-quality generations.

The early first-chapter stage also includes limited forward planning information for the next chapter. This helps the first chapter establish continuity into the rest of the book while still allowing the first prose realization to occur early in the sequence.

After the first chapter is generated, the sequence continues into the repeated chapter-generation block shown in panel~(c). Each later chapter receives its own chapter plan, scene-level breakdown, and final chapter text, providing more detailed structure for the remaining chapters, including chapter focus, narrative perspective, and target lengths.

\FloatBarrier{}

\subsection{Training Objective and Hyperconvergence}\label{sec:training-objective}

Training is supervised fine-tuning with a standard autoregressive cross-entropy objective. Each example is formatted as a single token sequence containing the initial prompt, scaffold headers, planning components, and prose components. The loss is masked on the initial prompt tokens only; every following token contributes to the loss.

Let $y_{1:T}$ denote the serialized token sequence and let $m_i$ be the loss mask for token $y_i$, where $m_i=0$ for prompt tokens and $m_i=1$ otherwise. We minimize
\[
\mathcal{L}(\theta)
=
-\frac{1}{\sum_{i=1}^{T} m_i}
\sum_{i=1}^{T}
m_i \log p_{\theta}(y_i \mid y_{<i}).
\]
This objective trains the model to reproduce both the prose and the scaffold structure used to generate that prose.

We train to hyperconvergence, following the training outline of the Hyperfitting work~\cite{carlsson2024hyperfitting}. In this setting, continued SFT is used not only to teach the model the dataset format, but also to sharpen open-ended long-form generation. We use the term \emph{hyperconvergence} for the point at which the model has strongly adapted to the scaffold and produces low-entropy, template-consistent continuations for the target generation process.

The maximum training sequence length is 262{,}144 tokens. Samples longer than this limit are clipped to the maximum length.

\subsection{Training System}\label{sec:training-system}

We train with JAX on a TPU v6e-256 pod~\cite{bradbury2018jax}. Training uses bfloat16 for the model copy used in the forward and backward passes~\cite{kalamkar2019bfloat16}. We use ZeRO~\cite{rajbhandari2020zero} to shard the FP32 master parameters and FP32 optimizer states. Optimizer computations are performed in FP32, while gradient accumulation is performed in bfloat16.

The long sequence length requires sequence-parallel attention. We use Ring Attention for sequence parallelism over the TPU topology, making 262{,}144-token training examples tractable across the pod~\cite{liu2023ringattention}.

We use the Muon optimizer for training~\cite{liu2025muon}. The FP32 master parameters are updated by the optimizer, and a bfloat16 training copy is materialized from the master parameters before the model computation.

\subsection{Learning-Rate-Scaled Stochastic Downcasting}\label{sec:stochastic-downcast}

A novel part of our training system is the stochastic downcast from the FP32 master parameters to the bfloat16 training copy. Prior work studies stochastic rounding as a way to reduce numerical error in low-precision LLM training~\cite{ozkara2025stochasticrounding}. Our setup differs from that setting: we keep the master parameters, optimizer states, and optimizer computations in FP32, and apply stochasticity only when materializing the bfloat16 copy used for training computation.

We first recall the idealized scalar stochastic-rounding operator. Let $\mathcal{F}_{\mathrm{bf16}}$ be the set of bfloat16-representable values. For a scalar $x$, let
\[
q^{-}(x)=\max \{q \in \mathcal{F}_{\mathrm{bf16}} : q \leq x\},
\qquad
q^{+}(x)=\min \{q \in \mathcal{F}_{\mathrm{bf16}} : q \geq x\}.
\]
Nearest rounding maps $x$ to the closest representable value,
\[
Q_{\mathrm{nr}}(x)
=
\arg\min_{q \in \mathcal{F}_{\mathrm{bf16}}} |q-x|.
\]
Stochastic rounding instead samples between the adjacent representable values:
\[
Q_{\mathrm{sr}}(x)
=
\begin{cases}
q^{+}(x), &
\text{with probability }
\dfrac{x-q^{-}(x)}{q^{+}(x)-q^{-}(x)}, \\[1.2em]
q^{-}(x), &
\text{with probability }
\dfrac{q^{+}(x)-x}{q^{+}(x)-q^{-}(x)}.
\end{cases}
\]
This gives
\[
\mathbb{E}\!\left[Q_{\mathrm{sr}}(x)\mid x\right]=x,
\]
so the rounding error has zero conditional mean in the ideal scalar case. In contrast, nearest rounding introduces a deterministic quantization error
\[
e_{\mathrm{nr}}(x)=Q_{\mathrm{nr}}(x)-x,
\]
which can systematically remove updates whose magnitude is below the local bfloat16 resolution.

Our implementation uses a dithered stochastic downcast rather than explicitly sampling the adjacent bfloat16 values. Let $\Theta_t$ be the FP32 master parameters at optimization step $t$. For each parameter element $\Theta_{t,i}$, we define the local bfloat16 resolution scale
\[
r(\Theta_{t,i})
=
\max\left(
|\Theta_{t,i}| \, \varepsilon_{\mathrm{bf16}},
\operatorname{tiny}_{\mathrm{fp32}}
\right),
\]
where $\varepsilon_{\mathrm{bf16}}$ is the bfloat16 machine epsilon and $\operatorname{tiny}_{\mathrm{fp32}}$ is the smallest positive normal FP32 value. We then sample
\[
u_{t,i} \sim \mathcal{U}(0,1)
\]
and construct a zero-mean perturbation
\[
\delta_{t,i}
=
\alpha_t
\left(u_{t,i}-\frac{1}{2}\right)
r(\Theta_{t,i}).
\]
The perturbation satisfies
\[
\mathbb{E}[\delta_{t,i}\mid \Theta_{t,i}] = 0,
\qquad
\operatorname{Var}[\delta_{t,i}\mid \Theta_{t,i}]
=
\frac{\alpha_t^2 {\left(r(\Theta_{t,i})\right)}^2}{12}.
\]
The bfloat16 training copy is then materialized as
\[
W_{t,i}
=
Q_{\mathrm{nr}}\!\left(\Theta_{t,i}+\delta_{t,i}\right).
\]
Equivalently, in vector form,
\[
W_t
=
Q_{\mathrm{nr}}
\left(
\Theta_t
+
\alpha_t \, r(\Theta_t) \odot
\left(u_t-\frac{1}{2}\right)
\right),
\qquad
u_t \sim {\left(\mathcal{U}(0,1)\right)}^{|\Theta_t|}.
\]

The stochastic strength $\alpha_t$ is tied to the learning rate. Let $\eta_t$ be the learning rate at step $t$, let $\eta_{\max}$ be the maximum learning rate, and let
\[
\eta_{\mathrm{floor}} = 7 \cdot 10^{-7}.
\]
We set
\[
\alpha_t
=
\operatorname{clip}
\left(
\frac{\eta_t - \eta_{\mathrm{floor}}}
{\eta_{\max} - \eta_{\mathrm{floor}}},
0,
1
\right).
\]
Thus, stochasticity is strongest at the maximum learning rate and is annealed as the learning rate decays. When $\eta_t \leq \eta_{\mathrm{floor}}$, we have $\alpha_t=0$, so the downcast becomes the deterministic nearest bfloat16 cast:
\[
W_t = Q_{\mathrm{nr}}(\Theta_t).
\]

The stochastic downcast is applied once per optimization step, when refreshing the bfloat16 training copy from the FP32 master parameters. The random stream is keyed by both the current training step and the parameter identity. For sharded parameters, the noise is generated using global data-parallel and tensor-parallel shard offsets, so the stochastic downcast is stable with respect to the distributed layout.

The motivation is that the FP32 master parameters may receive updates that are meaningful in FP32 but too small to survive repeated deterministic materialization into bfloat16. The stochastic downcast converts these sub-resolution changes into zero-mean rounding variability rather than always discarding them in the same direction. Prior convergence analysis of stochastic rounding shows that, under the analyzed optimizer setting, stochastic rounding can provide more favorable quantization-error behavior than nearest rounding~\cite{ozkara2025stochasticrounding}. Empirically, our learning-rate-scaled stochastic downcast improves convergence compared to deterministic FP32-to-bfloat16 materialization. We found the unscaled variant less stable, while annealing the stochasticity with the learning rate retained the convergence benefit and improved training stability.

\subsection{Generation Template and Constrained Decoding}\label{sec:generation-template}

At inference time, generation follows the same scaffold structure used during training. Examples are serialized using a Llama-3-style instruction format~\cite{grattafiori2024llama3}, but the standard \texttt{user} and \texttt{assistant} role identifiers are replaced with component-specific headers corresponding to the Planning Scaffold.

The model is trained and evaluated as a single-turn generation system. Each sample contains a single initial prompt with no system prompt and no interactive dialogue structure. Although the generation process expands into multiple intermediate planning and prose components, these are represented as structured continuations of the original request rather than as conversational turns.

During inference, decoding is constrained to the same structural template used during dataset construction. We implement this using regex-guided constrained decoding, which enforces the ordering of sections, component headers, and boundary markers. The constraint mechanism governs only the structural form of the output and does not constrain the semantic or stylistic content of the generated prose.

Generation proceeds hierarchically in the same order as the training representation described in Section~\ref{sec:training-data-representation}. The model first generates the book-level plan, followed by the early first-chapter planning stage and the first chapter itself.

\section{Evaluation}\label{sec:evaluation}

We evaluate the model from two complementary perspectives. First, we assess the quality of the generated writing using narrative-theory-grounded evaluations derived from prior work on creative-writing assessment and compare performance against leading general-purpose language models. Second, we evaluate generation stability by measuring how well the planning scaffold maintains and propagates information across the generated artifacts.

A particular challenge in this evaluation is that the scaffold was trained on public-domain books, while the benchmark contains prompts that differ substantially from that training distribution. The results therefore provide evidence not only about overall writing quality, but also about how well the learned planning representation transfers to new genres, settings, and narrative tasks.

The following subsections describe the benchmark, evaluation methodology, and experimental results.

\subsection{Evaluation prompt construction}\label{sec:evaluation-setup}

The evaluation prompt set contains 360 prompts written specifically for this study. The prompts are organized into 36 families, with ten variations per family. Of these prompts, 240 ask for original fiction and 120 ask for fanfiction. The same prompt set is used for both the writing-quality evaluation and the generation-stability evaluation.

A central goal of the prompt suite is to evaluate transfer beyond the distribution used during post-training. The planning scaffold is learned from a corpus of public-domain books, while the evaluation prompts primarily target modern stories and genres. The prompt suite is therefore designed to test whether the resulting model can apply its learned planning representations to different narrative requests.

The original-fiction prompts ask the model to develop a new book from a high-level request, testing transfer across genres and settings. The fanfiction prompts instead build on existing fictional worlds and characters. Success on both prompt types requires the model to combine the planning capabilities learned during post-training with broader knowledge acquired during foundation-model pretraining.

Appendix~\ref{sec:appendix-prompt-dataset-construction} provides a detailed description of the prompt-construction methodology and prompt-family design.

\subsection{Writing quality}\label{sec:evaluation-comparative}

Evaluating creative writing is inherently difficult. Unlike domains such as mathematics or programming, writing quality is not directly verifiable. Readers often agree on whether a story is engaging, coherent, or well written, but the underlying reasons for those judgments are harder to formalize. LLMs as judges are inadequate for evaluating creative writing because they are incapable of reliably distinguishing good writing from bad writing. We therefore draw on prior work in narrative theory to construct targeted benchmarks for effective fiction writing.

We evaluate generation quality using seven benchmarks derived from prior work in narrative theory. Each benchmark measures a specific narrative property associated with effective fiction writing. Scores are computed using a hybrid procedure in which language models extract structured information from the generated text and deterministic rules then determine whether the corresponding narrative criterion is satisfied. Appendix~\ref{app:score-operationalization} provides the full scoring methodology, including prompts and evaluation rules.

\begin{figure}[h]
\centering
\includegraphics[width=0.98\linewidth]{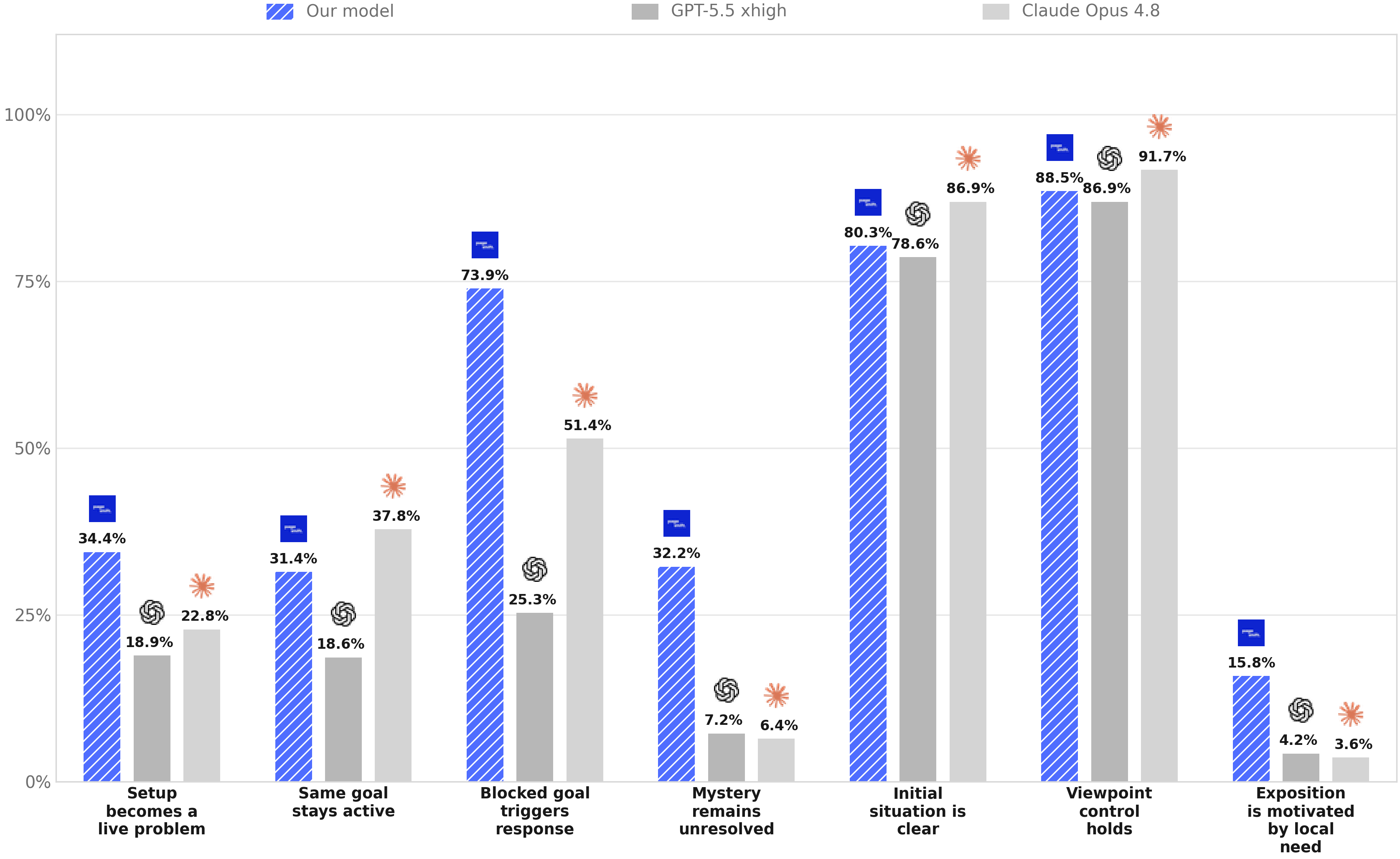}
\caption{Targeted first-chapter writing-quality comparison on seven literature-backed benchmarks from the shared 360-prompt benchmark. Bars report pass rates, with higher values indicating stronger performance on the corresponding benchmark.}
\label{fig:writing-quality-compact-main}
\end{figure}
\FloatBarrier

Figure~\ref{fig:writing-quality-compact-main} reports the main comparison of our model against the leading frontier language models, GPT-5.5 and Claude Opus 4.8. The full benchmark results, including additional frontier and leading open-weight model comparisons, are reported in Appendix~\ref{app:full-benchmark-results}.

Across the evaluation set, our model outperforms GPT-5.5 on all seven benchmarks and achieves the highest score on four of the seven benchmarks. Among the frontier baselines, Claude Opus 4.8 is generally the strongest comparison model and exceeds our model on three benchmarks: \emph{Same goal stays active}, \emph{Initial situation is clear}, and \emph{Viewpoint control holds}. These results suggest that frontier assistant models remain particularly strong on benchmarks that emphasize local consistency and clarity.

Our largest advantages appear on benchmarks associated with narrative development. In particular, we observe substantial gains on \emph{Blocked goal triggers response}, \emph{Mystery remains unresolved}, and \emph{Exposition is motivated by local need}. The result on \emph{Mystery remains unresolved} is especially notable: both frontier baselines score below 8\%, while our model reaches 32.2\%. A similar pattern appears on \emph{Exposition is motivated by local need}, where all models perform relatively poorly, but our model substantially outperforms both frontier baselines.

We hypothesize that this pattern reflects the difference between models optimized for creative writing and models optimized for general-purpose assistance. Assistant models are trained to answer questions, resolve uncertainty, and provide information directly. Fiction writing often requires the opposite behavior: maintaining open questions, delaying resolution, and introducing information only when it becomes relevant to the unfolding narrative. The largest performance gaps occur on benchmarks that measure these aspects of narrative construction.

\paragraph{Setup becomes live problem.}
A first chapter must make the setup a live problem. Elements initially introduced as background should become sources of pressure that begin to structure the story's events. Narrative theorists have long argued that stories become engaging when background conditions start producing developments that matter. One of the most basic distinctions is between establishing the situation and introducing the developments that give the story direction. Labov and Waletzky describe this contrast as the difference between orientation, which provides the relevant background, and complicating action, which introduces the events that create tension and move the narrative forward~\cite{labov1967narrative}. Bruner makes a similar argument at a broader cognitive level: stories depend on familiar expectations about how situations normally unfold, but become worth telling when those expectations are disrupted, creating what he calls Trouble~\cite{bruner1991narrative}. Related theories explain why some developments feel narratively significant. Changes become especially significant when they are perceived as consequential, unexpected, or norm-violating, and when they are important enough to give the story a point incident~\cite{huhn2014event,baroni2014tellability}. Story-grammar approaches translate this insight into a more operational form by treating stories as organized around initiating events that trigger responses and subsequent action, instead than around static description alone~\cite{stein1979analysis,rumelhart1975notes}. Accordingly, this diagnostic evaluates whether the requested setup develops into a source of narrative pressure that begins organizing the story, instead than remaining merely a premise, description, or genre frame.

\paragraph{Same goal stays active.}
Narrative comprehension often depends on the reader being able to track a character's continuing motivation. In the story-grammar tradition, an episode is not just a sequence of events: an initiating event evokes an internal response, which may include a goal or plan, and that response motivates attempts whose outcomes produce consequences and reactions~\cite{stein1979analysis,rumelhart1975notes}. Schank and Abelson's plan-based account makes the same point from the side of commonsense action understanding: to interpret behavior, readers infer the goals and plans that make actions intelligible~\cite{schank1977scripts}. Experimental work on goal information gives the corresponding comprehension evidence. Magliano and Radvansky argue that understanding story actions requires monitoring character goals, especially when plots involve actions directed at overcoming obstacles~\cite{magliano2001goal}. Lutz and Radvansky further show that goal information remains available as part of the causal representation of a narrative, with failed or unsatisfied goals remaining especially relevant because they continue to explain later action~\cite{lutz1997fate}. Palmer's account of fictional minds gives the narratological counterpart: readers construct characters as continuing consciousnesses from dispersed textual cues, and persistent motivations help those minds remain coherent across the chapter~\cite{palmer2004fictional}. Accordingly, this diagnostic asks whether an early goal, desire, obligation, or inquiry continues to organize later choices, delays, revisions, obstacles, or consequences, instead than being introduced once and then abandoned.

\paragraph{Blocked goal triggers response.}
A blocked goal should not leave a character stuck in place. Obstacles become narratively meaningful when they force characters to respond and adapt. Story-grammar approaches provide the main grounding for this diagnostic. In Stein and Glenn's model, stories develop through a sequence in which a problem produces a response, the response motivates a plan, and the plan leads to attempts whose consequences determine whether the goal is achieved~\cite{stein1979analysis}. Rumelhart's story-schema work develops a similar idea, treating stories as organized around recurring patterns of response, attempt, consequence, and reaction~\cite{rumelhart1975notes}. Goal-based theories likewise emphasize that blocked goals require adaptation. Schank and Abelson argue that when an obstacle prevents a goal from being achieved, characters must form new plans or subplans to address the blocking condition~\cite{schank1977scripts}. Reader-oriented accounts provide a complementary explanation: events become important when they contribute to a chain of causes and consequences, and narrative understanding depends on connecting actions and outcomes into a coherent whole~\cite{trabasso1985relatedness,herman2002story}. Accordingly, this diagnostic evaluates whether a blocked goal produces a meaningful response instead than inert frustration. The response may take many forms, but it should alter the character's situation and shape what happens next.

\paragraph{Mystery remains unresolved.}
The beginning of a story should answer enough questions to establish the narrative while leaving important questions unresolved. Readers should feel that the story has begun, not that it has already finished. Narrative theorists have long argued that stories maintain interest by controlling what readers know and when they know it. Barthes argues that narratives often achieve this by raising questions, delaying answers, and gradually revealing information over time~\cite{barthes1974sz}. Sternberg similarly argues that readers continually form and revise expectations as they fill informational gaps and test competing interpretations of events~\cite{sternberg1978expositional}. Reader-response approaches explain why such incompletion is productive rather than frustrating. Iser argues that gaps and omissions invite readers to actively connect different parts of the text through anticipation and retrospection~\cite{iser1978act}. Brewer and Lichtenstein provide the corresponding affective account, linking suspense and curiosity to narrative structures that postpone important outcomes or information while keeping them understandable~\cite{brewer1982stories}. Carroll makes the complementary point that narratives feel complete when their major questions have been answered~\cite{carroll2007closure}. Accordingly, this diagnostic evaluates whether important questions remain active by the end of the first chapter without becoming arbitrary or merely vague. Local uncertainties may be resolved, but larger questions tied to the story's problems, goals, stakes, or world should remain open.

\paragraph{Initial situation is clear.}
The beginning of a story should give readers enough information to understand what is happening, who it involves, and why the situation matters. Narrative theorists have long argued that readers need enough contextual information to make sense of a story's events. Labov and Waletzky describe this function as orientation: the information that establishes who is involved, where and when events occur, and how the situation should be understood~\cite{labov1967narrative}. Story-grammar approaches make a similar point by treating characters and context as the foundation on which later events can be understood. In Stein and Glenn's model, the setting introduces the main character and the social, physical, or temporal context that allows the story's events to unfold~\cite{stein1979analysis}. Sternberg's account of exposition clarifies that such information does not need to be delivered all at once. Readers only need the background necessary to understand the current action, while additional context can be supplied later as the story develops~\cite{sternberg1978expositional}. Rabinowitz further argues that readers rely on narrative conventions to determine which details are significant and how they fit together into a coherent situation~\cite{rabinowitz1987before}. Emmott and Herman provide the corresponding cognitive account, showing how readers use textual cues to build and update mental representations of characters, contexts, and storyworlds~\cite{emmott1997narrative,herman2002story}. Accordingly, this diagnostic evaluates whether the opening provides enough context to locate the central situation and follow the first scene, without requiring exhaustive exposition or slowing the narrative with static explanation.

\paragraph{Viewpoint control holds.}Readers should be able to tell whose perspective is guiding the story at any given moment. Narratives may shift between narrators and characters, but those shifts should remain understandable. Narrative theorists have long argued that perspective and narration are not the same thing. A story can be told by one voice while presenting events through the perceptions, knowledge, and judgments of a particular character. Genette distinguishes between who tells the story and who perceives it, while Rimmon-Kenan and Bal develop this idea by treating narrative information as filtered through a particular perspective that determines what can be seen, known, and evaluated~\cite{genette1980narrative,rimmonkenan2002narrative,bal2009narratology}. This distinction becomes especially important when stories represent character consciousness. Cohn analyzes how fiction presents thought and awareness, Fludernik emphasizes the role of subjective experience in narrative understanding, and Palmer argues that readers construct coherent fictional minds from cues distributed throughout a text~\cite{cohn1978transparent,fludernik1996natural,palmer2004fictional}. Reader-oriented accounts explain why viewpoint must remain trackable. Readers need to distinguish between what a character believes, perceives, or assumes and what the story itself presents as fact. Wiebe argues that this depends on tracking psychological point of view, while experimental work by Black, Turner, and Bower shows that unmarked shifts in perspective can make stories harder to follow~\cite{wiebe1994tracking,black1979point}. Accordingly, this diagnostic evaluates whether perceptions, thoughts, knowledge, and judgments remain attributable to a narrator or character. Deliberate viewpoint shifts can succeed when they are clearly signaled; failures include uncontrolled head-hopping, impossible knowledge, or passages where character belief and narrative assertion become difficult to distinguish.

\paragraph{Exposition is motivated by local need.}
Information should enter a story when readers need it. Background details become most effective when they help readers understand what is currently happening and why it matters. Information does not need to appear the moment a story begins. What matters is that readers receive it when it becomes useful for understanding unfolding events. Sternberg argues that exposition is fundamentally a problem of timing and information management rather than a fixed block that belongs at the beginning of a narrative~\cite{sternberg1978expositional}. Readers can receive information early, late, indirectly, or retrospectively, as long as its placement helps them understand and interpret the story. Readers are also more likely to accept background information when it feels necessary rather than arbitrary. Tomashevsky argues that textual elements feel justified when they are integrated into the developing structure of a work rather than appearing as isolated insertions~\cite{tomashevsky1965thematics}. Barthes likewise suggests that information becomes meaningful when it contributes to the reader's understanding of the unfolding narrative rather than remaining an isolated collection of facts~\cite{barthes1974sz}. Accordingly, this diagnostic evaluates whether exposition is made necessary by the chapter's present action or interpretive demands. Failures include encyclopedia-like explanation, backstory that arrives before readers have a reason to care about it, or world information that never becomes active.

\FloatBarrier

\subsection{Prompt and scaffold following}\label{sec:evaluation-results}

We evaluate prompt and scaffold following on the shared 360-prompt benchmark described in Section~\ref{sec:evaluation-setup}. Specifically, we measure how well user-specified requirements continue to be reflected across successive stages of the generation scaffold. The evaluation covers three comparisons: the user prompt to the Book Preview, the Book Preview to the Book Plan, and the Book Plan to the first-chapter text.

Table~\ref{tab:local-retention-main} reports the prompt- and scaffold-following results by story type. Both original fiction and fanfiction follow most prompt requirements in the Book Preview, achieving scores of 93.7\% and 86.0\%, respectively. Scaffold following declines substantially at subsequent stages. For original fiction, the score falls to 54.2\% from the Book Preview to the Book Plan and to 39.2\% from the Book Plan to the first-chapter text. Fanfiction exhibits a larger initial decline, reaching 28.1\% at the preview-to-plan transition, followed by a plan-to-prose score of 31.1\%.

\begin{table}[h]
\centering
\footnotesize
\setlength{\tabcolsep}{5pt}
\begin{tabular*}{0.98\linewidth}{@{\extracolsep{\fill}}lccc}
\toprule
Split
& \shortstack{Prompt to\\Book Preview}
& \shortstack{Book Preview to\\Book Plan}
& \shortstack{Book Plan to\\first-chapter text} \\
\midrule
Original fiction & 93.7\% & 54.2\% & 39.2\% \\
Fanfiction        & 86.0\% & 28.1\% & 31.1\% \\
\bottomrule
\end{tabular*}
\caption{Prompt- and scaffold-following scores across successive stages of the generation scaffold. Scores are linearly normalized from the shared 1--7 scale, with 1 corresponding to 0\% and 7 corresponding to 100\%.}
\label{tab:local-retention-main}
\end{table}
\FloatBarrier

These results indicate that the Book Preview closely follows the user's request, but that later scaffold components do not consistently preserve and realize all requested elements. The largest losses occur during the conversion of the Book Preview into the Book Plan, with additional degradation occurring when planning information is translated into prose.

The difference between original fiction and fanfiction is particularly informative in light of the distribution shift introduced by our evaluation benchmark. As discussed in Section~\ref{sec:evaluation-setup}, the planning scaffold is learned from a corpus of public-domain books, while the benchmark primarily targets modern story concepts and includes a substantial number of fanfiction requests. Original-fiction prompts, which remain closer to the training distribution, achieve consistently higher scores. However, the model continues to follow prompt and scaffold information across both story types, including fanfiction prompts that are substantially more distant from the scaffold-training distribution. We interpret this as evidence that the model is able to combine planning capabilities learned from public-domain books with the broader world knowledge acquired during foundation-model pretraining. This allows the learned planning representation to transfer beyond the specific genres, settings, and story types present in the scaffold-training corpus and to generalize to modern narrative requests that were not directly represented during scaffold training.

At the same time, the lower fanfiction scores indicate that this transfer is incomplete. Original-fiction prompts primarily require the model to apply its planning capabilities to genres, settings, and concepts that are broadly represented in pretraining, whereas fanfiction prompts additionally require knowledge of specific characters, relationships, events, and canon constraints from established fictional works. The larger performance degradation on fanfiction therefore suggests that combining learned planning behavior with highly specific story knowledge remains a more challenging transfer setting.

A second pattern visible in Table~\ref{tab:local-retention-main} is the progressive decline in scaffold following across successive generation stages. While prompt following remains high in the Book Preview, performance drops considerably in the Book Plan and remains limited in the generated prose. This degradation is observable across both story types and represents one of the primary limitations of the current model. The scaffold does not yet reliably propagate all relevant information through the full generation pipeline, resulting in the gradual loss of requested details and constraints as generation progresses. Improving prompt and scaffold following across stages therefore is one of the most important directions for future work, as stronger information propagation will improve both prompt adherence and long-range narrative control.

\subsection{Long-generation consistency}\label{sec:evaluation-carry}

\begin{figure}[h]
\centering
\begin{tikzpicture}[x=0.62cm,y=0.045cm]
\path[use as bounding box] (-1.1,-9.0) rectangle (15.95,106.0);
\draw[->] (0.65,0.0) -- (15.55,0.0) node[right]{Chapter};
\draw[->] (1.0,-1.5) -- (1.0,103.0);
\node[rotate=90, anchor=center] at (-0.55,50.0) {Chapter-plan following score (\%)};
\foreach \x in {1,...,15} {
\draw (\x,1.2) -- (\x,-1.2) node[below=6pt] {\scriptsize \x};
}
\foreach \y in {0,20,40,60,80,100} {
\draw[gray!30] (1,\y) -- (15.35,\y);
\draw (1.08,\y) -- (0.92,\y) node[left=6pt] {\scriptsize \y};
}
\draw[thick] (1,73.1) -- (2,65.4) -- (3,65.5) -- (4,65.5) -- (5,62.9) -- (6,61.4) -- (7,59.7) -- (8,51.4) -- (9,58.6) -- (10,57.4) -- (11,57.0) -- (12,37.2) -- (13,58.5) -- (14,54.2) -- (15,50.8);
\foreach \x/\y in {
1/73.1, 2/65.4, 3/65.5, 4/65.5, 5/62.9, 6/61.4, 7/59.7, 8/51.4,
9/58.6, 10/57.4, 11/57.0, 12/37.2, 13/58.5, 14/54.2, 15/50.8
} {
\fill (\x,\y) circle (1.7pt);
}
\end{tikzpicture}
\caption{Mean chapter-level plan-following score between each chapter plan and its generated text. Scores are assigned on a 1--7 rubric and linearly mapped to percentages.}
\label{fig:later-chapter-trend-main}
\end{figure}
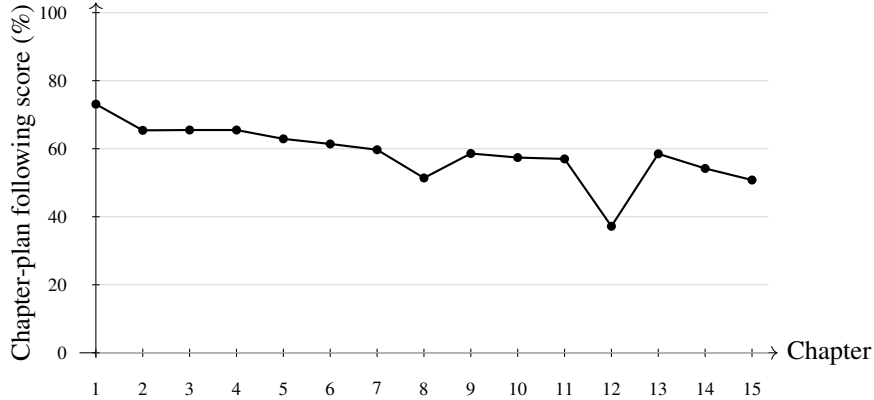

Using the shared 360-prompt benchmark, we evaluate whether the planning scaffold continues to guide generation across later stages of a book. For each chapter, we compare the generated chapter text with its corresponding chapter plan and measure the extent to which the generated text follows the planned content. Figure~\ref{fig:later-chapter-trend-main} reports the mean chapter-level plan-following score for each chapter position.

\FloatBarrier

Chapter~1 achieves the highest plan-following score at 73.1\%. The score falls to 65.4\% for Chapter~2 and then follows a broadly downward trajectory across later chapters. The chapters immediately following Chapter~1 remain between 59.7\% and 65.5\%, while later chapters generally achieve lower scores, ending at 50.8\% for the final evaluated chapter. Across all subsequent chapters, scores range from 37.2\% to 65.5\%. Individual chapter scores exhibit some variability, but the overall pattern is one of gradually decreasing plan following as generation progresses.

The pronounced difference between Chapters~1 and~2 is partly explained by the structure of the planning scaffold. The dedicated First Chapter Plan is generated directly from the book-level plan, and Chapter~1 is then generated from this component. Both steps are completed before the model constructs the chapter plans used for the remainder of the book. Only after Chapter~1 has been generated does the model produce the subsequent chapter plans that guide the remainder of the book. Chapter~1 therefore benefits from a planning stage that is more directly connected to the book-level scaffold, whereas later chapters depend on planning components generated during a downstream stage of the generation process. This difference in scaffold structure explains the initial drop observed between Chapters~1 and~2.

Beyond the initial drop, plan-following scores exhibit a clear downward trend as generation progresses, suggesting that the influence of the planning scaffold diminishes over longer contexts. We find the resulting level of chapter-plan adherence insufficient for reliable book-length generation, particularly in later chapters, where a substantial portion of the planned content is no longer consistently realized in the generated text. Addressing this degradation in long-range scaffold adherence therefore represents an important direction for future work.

\section{Interpretability}%
\label{sec:interpretability}
We use interpretability to analyses and to ask whether the model's long-form generations are supported by structured internal behavior. The analyses operate at two complementary levels. At the Planning Scaffold level, we inspect attention patterns to test whether generated prose continues to retrieve from components designed to hold book-level information, such as the book plan, world rules, and related planning fields. At the feature level, we use sparse autoencoder (SAE) latents to ask whether activations contain reusable directions for recurring book-writing operations.

These analyses answer different parts of the same interpretability question. The attention analysis tests the routing assumption behind the Planning Scaffold: when writing a chapter, the model should balance local same-chapter continuation with lookup into earlier planning fields. The SAE analysis tests whether the resulting prose-level behavior can be decomposed into candidate mechanisms that are active in interpretable contexts and that causally shift local next-token contrasts. We therefore present the Planning Scaffold-level attention analysis first, followed by the SAE discovery-and-validation pipeline and its retained feature-level findings.

\subsection{Planning Scaffold-Level Attention Analysis}%
\label{sec:mechanistic}

The Planning Scaffold is intended to support retrieval from structured planning fields during generation. We examine attention patterns to test whether the trained model uses the Scaffold as intended. This provides a Scaffold-level view of information routing.

For each analyzed generation, we align tokens to the semantic components induced by the Planning Scaffold and measure how generated prose allocates attention between prior Scaffold components and earlier tokens in the current chapter. We summarize this behavior using the prior-component attention curve. At each generated token, this curve measures the fraction of attention routed from the current chapter to earlier Planning Scaffold components; the remaining mass stays within the chapter. Across typical generations, this allocation is approximately balanced. Roughly half of the attention remains local, supporting immediate continuation and short-range coherence, while the remainder retrieves information from the broader Planning Scaffold when needed.

\subsubsection{Global Planning Scaffold Routing}%
\label{subsec:global-scaffold-routing}

Figure~\ref{fig:component-attention-structure} shows the average component-level attention pattern across 128 book-generation samples. The diagonal captures local autoregressive continuation, while the structured off-diagonal mass shows that later generation stages route attention back to earlier Planning Scaffold components. The prompt, book plan, chapter plans, and character-related fields therefore function as persistent reference sources rather than disposable intermediate text.

\begin{center}
\includegraphics[width=0.58\linewidth]{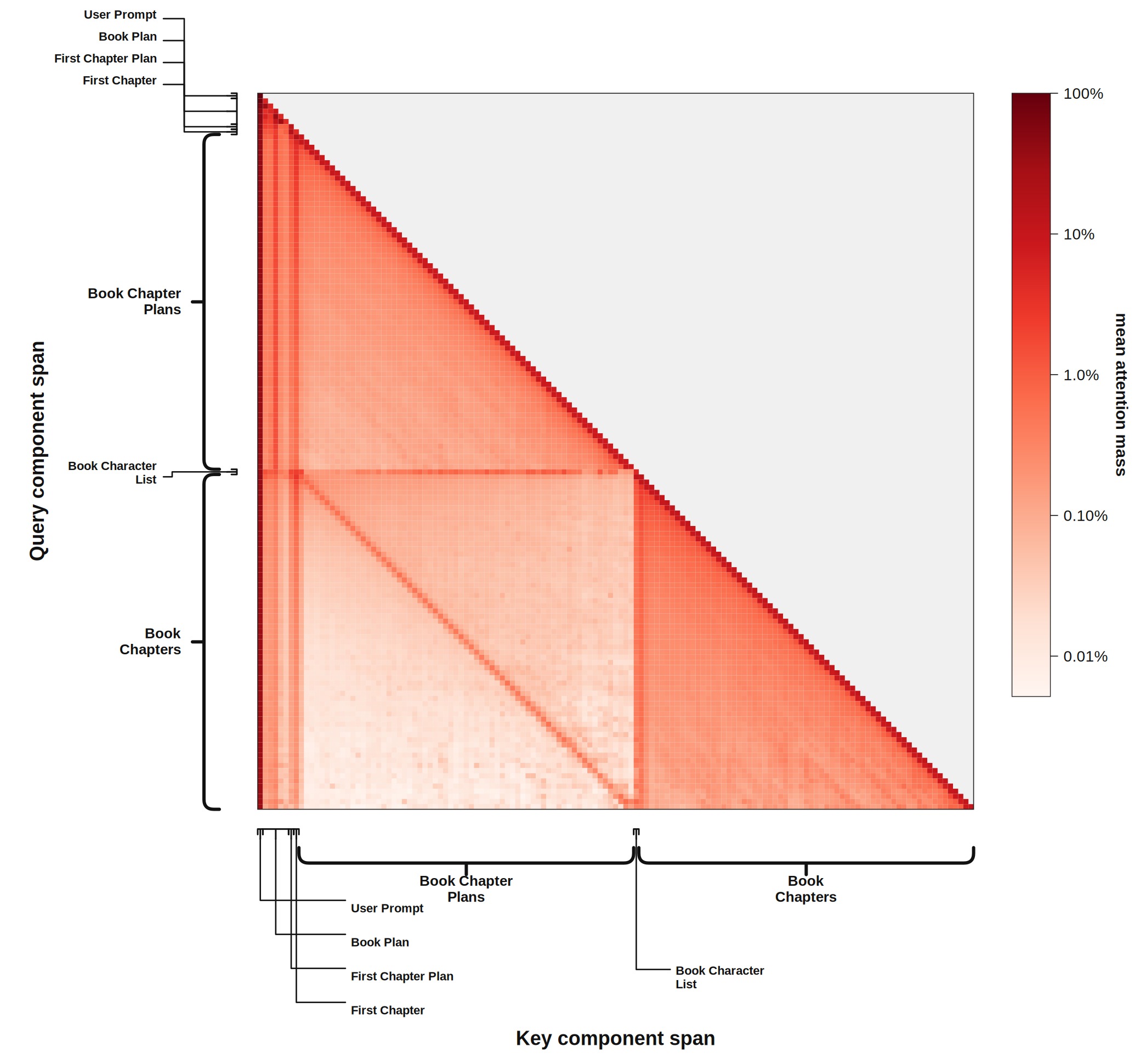}
\captionof{figure}{Mean component-level attention across 128 book-generation samples.}%
\label{fig:component-attention-structure}
\vspace{-0.5em}
\end{center}

The aggregation pattern is consistent with the intended organization of the Planning Scaffold. Local chapter context primarily supports short-range continuation, while the broader Planning Scaffold provides longer-range constraints and retrieval targets. This suggests that the Planning Scaffold components remain available as information sources throughout generation instead of functioning solely as initialization context.

\subsubsection{First-Chapter Planning Scaffold Use}%
\label{subsec:chapter-level-attention}

The first chapter provides a useful setting for examining how the model converts planning information into sustained prose. In this example, we use a layer-wise view of prior-component attention, which shows both how Planning Scaffold retrieval changes across generated token positions and how this retrieval is distributed across model depth.

\begin{center}
\includegraphics[width=0.66\linewidth,height=0.32\textheight,keepaspectratio]{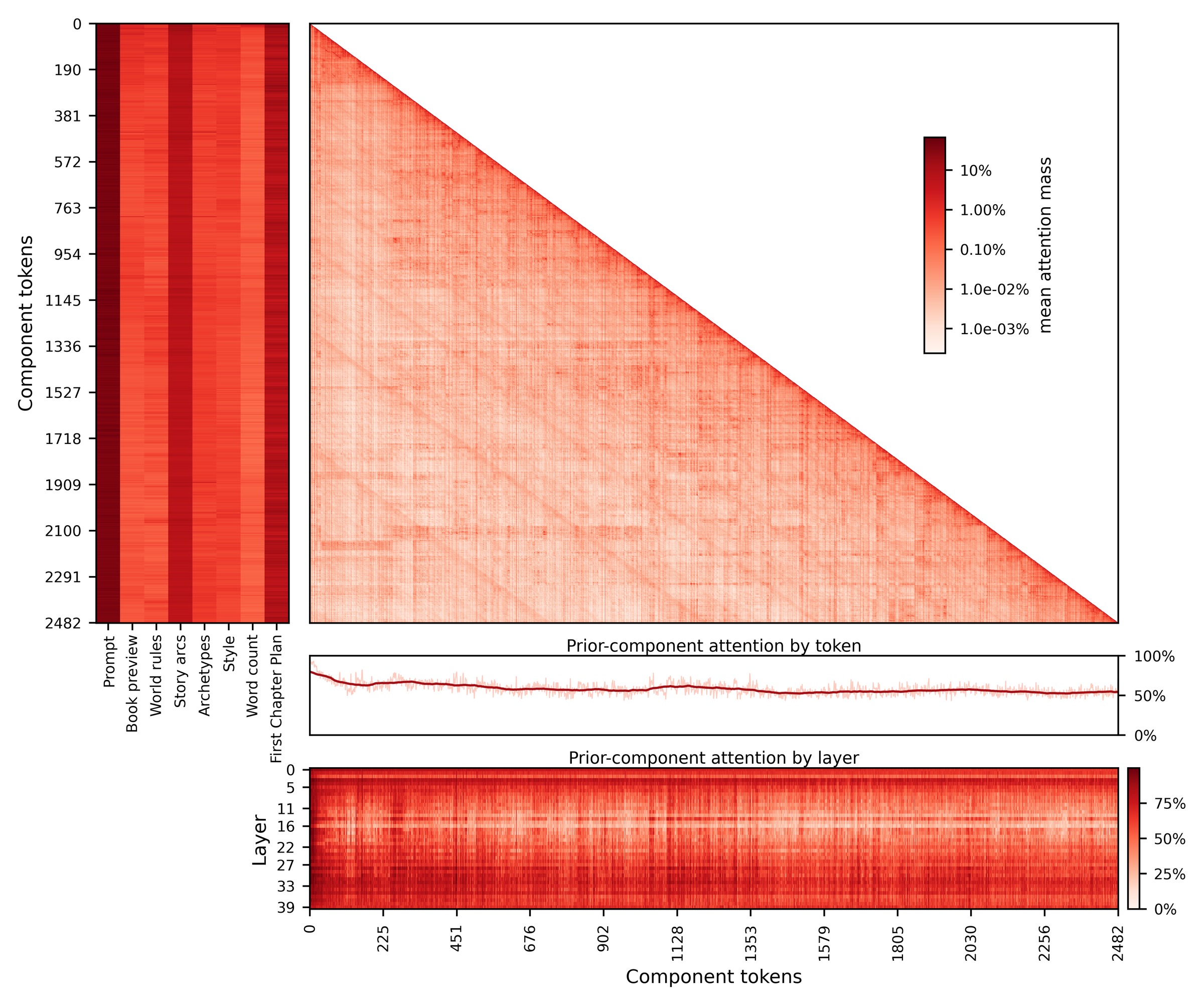}
\captionof{figure}{Layer-wise prior-component attention during first-chapter generation.}%
\label{fig:first-chapter-layer-outside-mass}
\vspace{-0.5em}
\end{center}

Figure~\ref{fig:first-chapter-layer-outside-mass} shows that prior-component attention remains active throughout the first chapter, with the total prior-component mass staying in a balanced regime instead of collapsing into purely local continuation. The layer-wise decomposition further shows that this balance is not uniform across depth. Early and late layers allocate substantial attention to prior Planning Scaffold components, while middle layers place comparatively more attention within the current chapter. This suggests a depth-dependent organization of the generation process: middle layers primarily support local continuation within the chapter, whereas early and late layers maintain access to planning information from the Scaffold.

\subsubsection{Targeted World-Rule Retrieval}%
\label{subsec:world-rules-attention}

Figure~\ref{fig:first-chapter-world-rules-focus} shows a first-chapter generation containing a localized region of unusually strong attention to the \textit{World rules} component. The callout zooms into this region and highlights individual generated tokens according to the amount of attention they assign to the world-rule field.

\begin{center}
\includegraphics[width=0.66\linewidth,height=0.24\textheight,keepaspectratio]{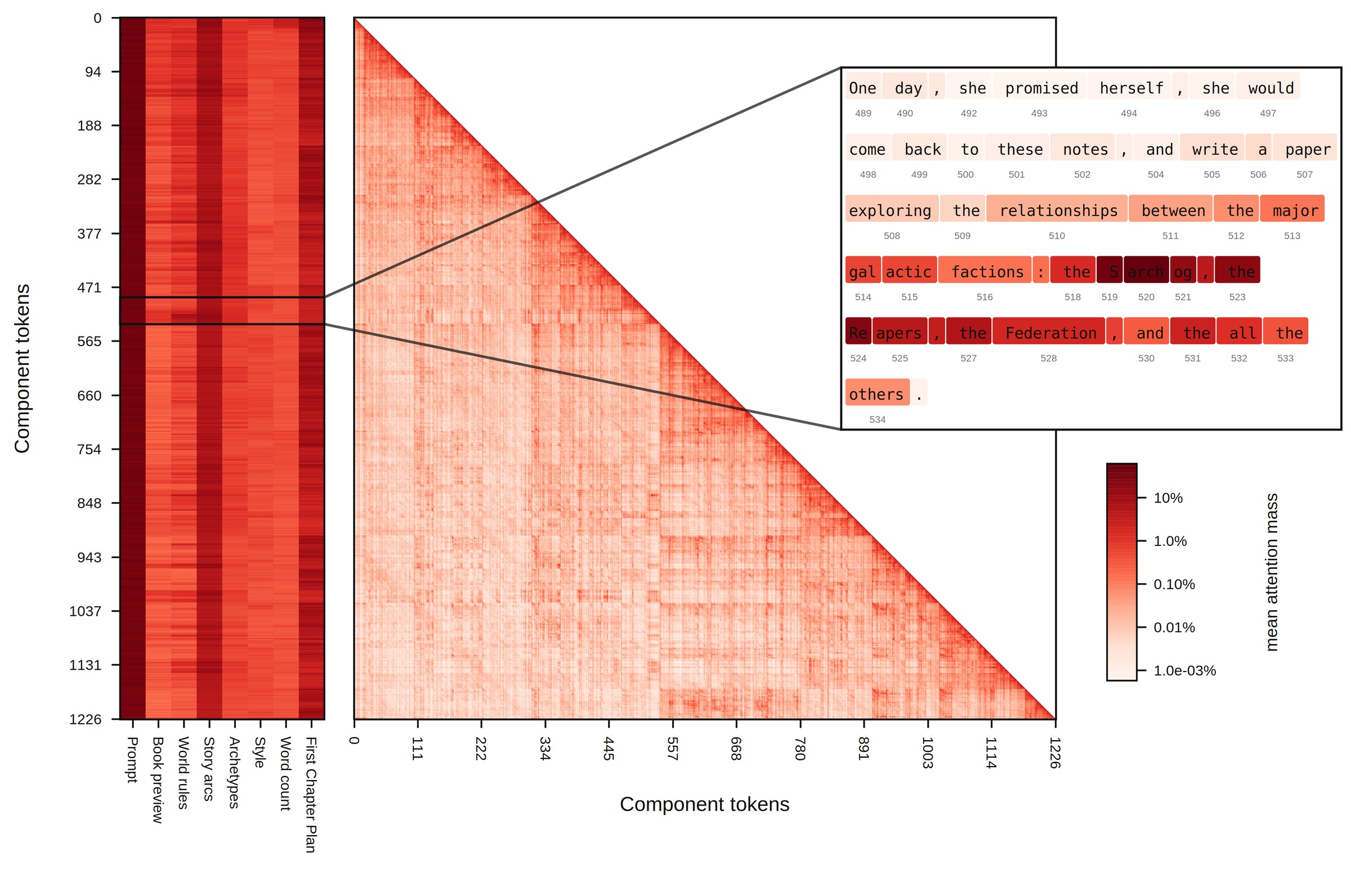}
\captionof{figure}{Localized first-chapter retrieval from the \textit{World rules} component.}%
\label{fig:first-chapter-world-rules-focus}
\vspace{-0.5em}
\end{center}

The elevated world-rule attention is concentrated around generated discussion of factions and their political relationships, including alliances and rivalries. This pattern is consistent with the intended contents of the world-rule field, which stores setting-specific entities, relationships, and other world-state information. This example provides qualitative evidence that strong retrieval from the world-rule component tends to occur in contexts where such information is expressed in the generated prose.

More broadly, the example suggests that retrieval from prior Planning Scaffold components is not uniformly distributed throughout generation. Instead, particular Planning Scaffold fields can become especially important when the generated text invokes the type of information they were designed to encode.

\subsubsection{Later Chapter Generation}%
\label{subsec:later-chapter-generation}

Figure~\ref{fig:later-chapter-continuity} shows attention during a later-chapter generation. In this setting, prior components include both the Planning Scaffold and previously generated narrative text. The figure therefore provides a view of how the model balances local continuation against retrieval from earlier context once substantial prose has already been generated.

\begin{center}
\includegraphics[width=0.66\linewidth,height=0.26\textheight,keepaspectratio]{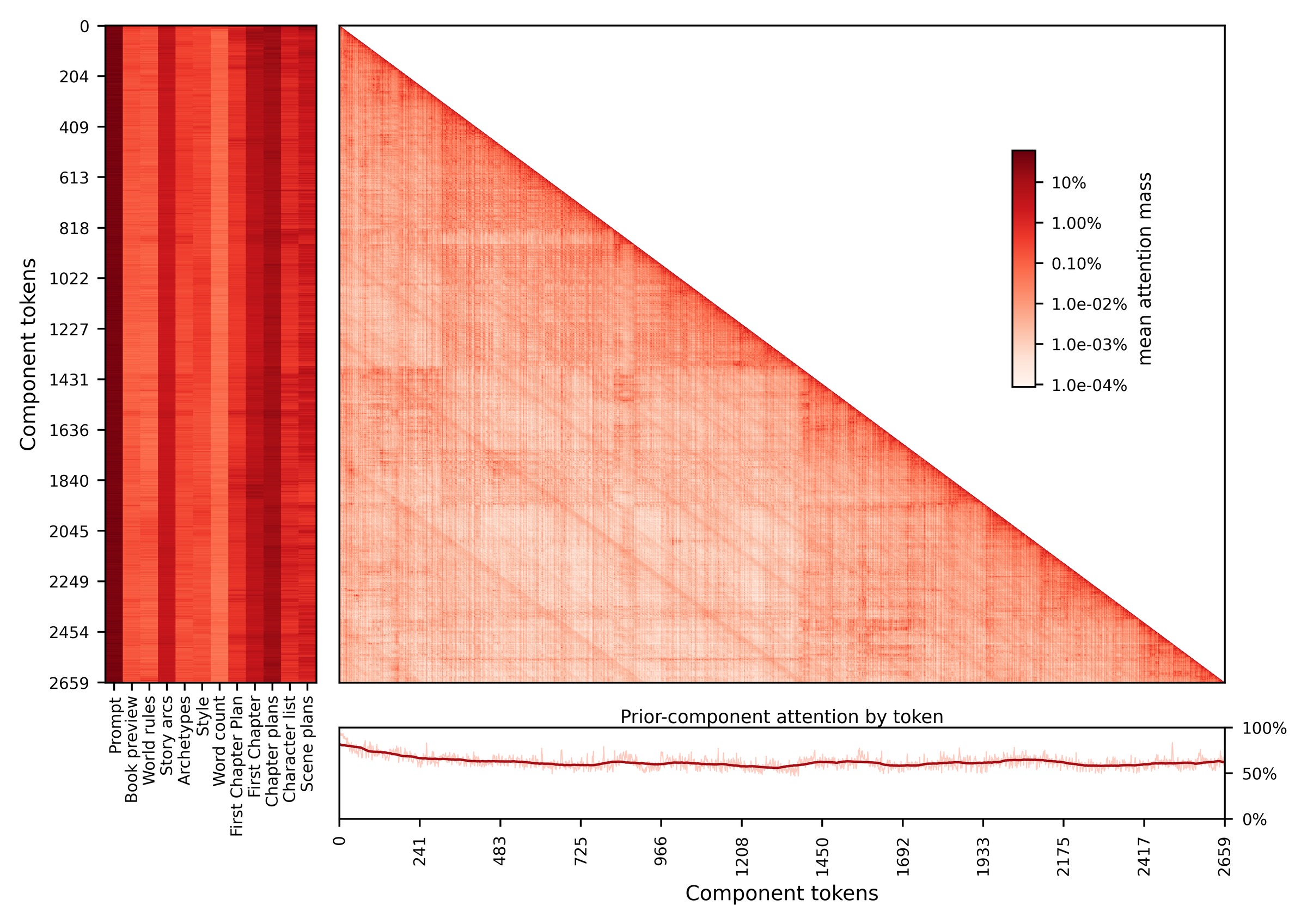}
\captionof{figure}{Prior-component attention during later-chapter generation.}%
\label{fig:later-chapter-continuity}
\vspace{-0.5em}
\end{center}

Even after substantial narrative context has accumulated, attention remains divided between the current chapter and prior components. Notably, strong attention continues to be directed toward the Planning Scaffold despite the availability of earlier chapter text. This suggests that later-chapter generation relies on both previously generated prose and the structured planning fields that define longer-range narrative constraints.

Taken together, the attention analyses indicate that Planning Scaffold usage persists throughout generation. The aggregate attention patterns show sustained access to prior components, the world-rule example shows localized retrieval in world-state-related contexts, and the later-chapter analysis shows that this behavior remains present even when extensive narrative context is available. These findings are consistent with the Planning Scaffold serving as a stable source of information throughout long-form generation.

\subsection{Sparse Autoencoder Feature Analysis}%
\label{subsec:sae-feature-analysis}

Sparse autoencoders (SAEs) provide a method for decomposing language-model activations into a sparse set of latent features. Internal activations in large language models are generally believed to exhibit superposition, in which multiple computational patterns are represented within the same activation dimensions. Consequently, individual dimensions are often difficult to interpret directly.

An SAE is trained to reconstruct a model activation after projecting it into a higher-dimensional latent space. Sparsity is enforced through a Top-$K$ activation constraint, which allows only a small number of latent features to be active for any given token. This encourages the learned representation to distribute information across a large collection of sparse latent directions.

The resulting latent features provide a useful unit of analysis for interpretability. Prior work has shown that SAE latents can often be associated with recurring semantic, syntactic, or functional patterns in model computation~\cite{bricken2023monosemanticity,cunningham2023sparse}. Instead than analyzing raw activations directly, we therefore study the activation of SAE features and ask whether particular features are consistently associated with recurring writing behaviors in generated book content.

In this section, we investigate whether the model's activations contain reusable feature directions associated with recurring writing behaviors. This analysis is exploratory. Instead of beginning from a predefined set of hypotheses, we derive candidate labels from the generated corpus and use them to guide the search for interpretable SAE features.

We train our sparse autoencoders with 82{,}944 latent features and a Top-$K$ sparsity of $K=64$. The training procedure generally follows the recommendations of Gao et al.~\cite{gao2024scaling}, including the token horizon and learning-rate schedule.

Let $x$ denote a model activation at a token position. The SAE maps $x$ to a sparse latent code $z$ and reconstructs the activation as

\begin{equation}
z = f_{\mathrm{enc}}(x), \qquad \hat{x} = b + Dz .
\end{equation}

Here $z_i$ denotes the activation of feature $i$, and $D_i z_i$ is its contribution to the reconstruction. Throughout the analysis, these latent directions are treated as candidate explanatory features.

Our approach differs from prior SAE analyses primarily in feature discovery. As summarized in Figure~\ref{fig:sae-feature-pipeline}, we use a five-stage pipeline: first constructing a corpus-derived label vocabulary with an automated label-discovery procedure, then using that vocabulary to retrieve candidate SAE features from generated text, refining candidates based on their activation patterns, validating them through attribution and intervention analyses, and finally auditing token-level activation traces. The following subsections describe this pipeline and the resulting retained features.

\begin{center}
\includegraphics[width=0.88\linewidth]{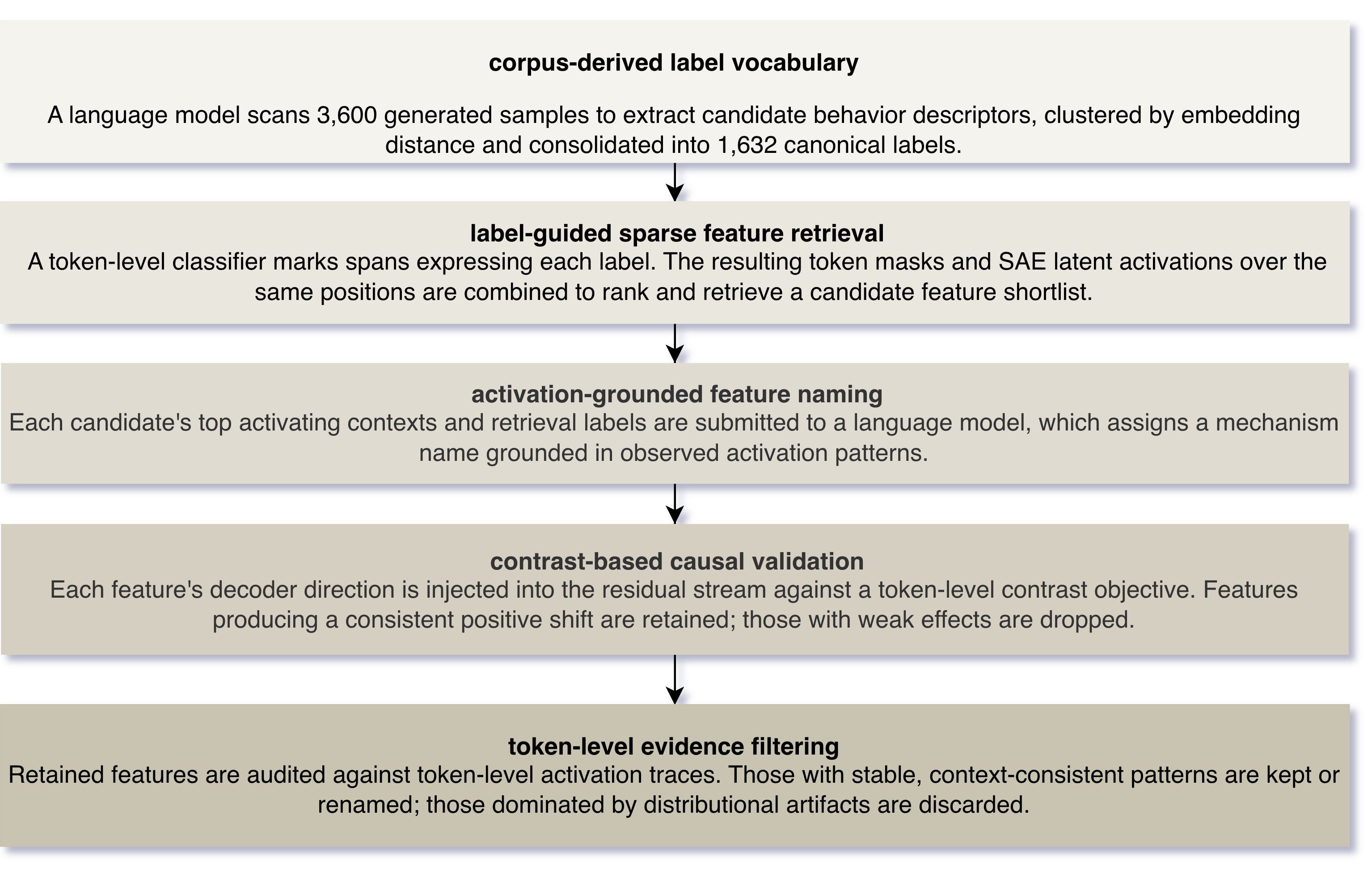}
\captionof{figure}{Five-stage SAE feature-discovery and validation pipeline.}%
\label{fig:sae-feature-pipeline}
\vspace{-0.5em}
\end{center}

\subsubsection{Corpus-Derived Label Vocabulary}%
\label{subsec:label-vocabulary}

Before retrieving SAE features, we construct a corpus-derived label vocabulary. This addresses a vocabulary problem: behavior-directed feature discovery requires candidate behavior descriptions, but a small, manually curated label set would largely reflect researcher priors. We therefore use an LLM to propose short, labels from the full generation record of each sample, including both the prompt and generated text. The objective is not to interpret SAE features at this stage, but to construct a broad vocabulary of recurring patterns in the corpus.

The resulting label-discovery run processed 3{,}600 generated samples and produced a canonical vocabulary of 1{,}632 labels. These labels provide the retrieval vocabulary for the subsequent SAE analysis.

Label proposals are pooled into a single vocabulary. The proposal prompts exclude formatting artifacts, structural markup, unsupported labels, and one-off entities such as unique character or location names. For auditing purposes, we retain provenance information, including aliases, source-sample identifiers, and whether a proposal originated from the prompt or generated text.

We canonicalize the pooled vocabulary in three stages. First, label strings are normalized to remove spelling, capitalization, and formatting variants. Second, the normalized labels are embedded using a semantic embedding model and grouped according to embedding distance. Labels whose embedding distance falls within a threshold of \texttt{0.230} are assigned to the same candidate cluster. Third, each cluster is submitted to an LLM, which removes duplicates and merges closely related labels while preserving distinctions useful for feature discovery. The resulting vocabulary of 1{,}632 labels is the output of this canonicalization procedure.

This hybrid approach combines embedding-based grouping with LLM-assisted consolidation. The embedding threshold provides a scalable way to identify semantically related labels, while the LLM resolves duplicates and naming variations within each cluster. The final vocabulary retains provenance metadata and serves as the retrieval vocabulary for the subsequent SAE analysis.

\subsubsection{Label-Guided Sparse Feature Retrieval}%
\label{subsec:label-guided-retrieval}

Given the LLM-assisted label vocabulary, we run a permissive label-conditioned scan over 3{,}600 generated samples. The purpose of this stage is high recall: it should surface sparse features that may correspond to recurring book-writing mechanisms, even at the cost of producing candidates that are later revised or rejected.

For each candidate label, we use an LLM as a semantic token-level classifier over the generated output. Instead than relying on string matching, the classifier is asked whether the label is expressed in the text and, when it is, which exact output tokens or short spans provide the evidence. This produces weak token-level masks for label expression. Because many mechanisms are not localized to a single word, we also consider a small neighborhood around each marked span when retrieving SAE latents. This allows the scan to recover features that activate on the evidence itself, on nearby setup tokens, or on immediately surrounding continuation structure.

The initial scan combines several weak signals. We rank features by their association with LLM-marked spans, their selectivity for labeled versus unlabeled regions, their behavior across output positions, their top activating contexts, and their local attribution to behavior-directed objectives. None of these signals is treated as decisive. They are used to reduce a large latent space to a tractable set of candidates for later inspection.

For a scalar objective \(J\), such as a positive-minus-negative token logit difference, we approximate the local relevance of feature \(i\) by
\begin{equation}
\Delta J_i \approx z_i \frac{\partial J}{\partial z_i}.
\end{equation}
This quantity is used only as a first-order ranking statistic. It prioritizes features that are both active and locally aligned with the queried behavior, but it does not establish that the feature causally implements the behavior.

\subsubsection{Activation-Grounded Feature Naming}%
\label{subsec:activation-grounded-naming}

After candidate discovery, we assign mechanism names using an automated context-based renaming procedure. The label that initially retrieved a feature is retained as provenance information, while feature interpretation is based on activation evidence. This approach is motivated by recent work on generating natural-language interpretations for SAE latents at scale~\cite{paulo2025automatically}, although we use the resulting names conservatively and retain them only when they remain supported by activation patterns and later validation analyses.

Because the automatically discovered label vocabulary is intentionally redundant and partially overlapping, individual SAE features are often retrieved by multiple related labels. For each feature, we collect the associated labels together with representative activating contexts and provide both to an LLM. The LLM then proposes a concise mechanism name that summarizes the recurring pattern expressed across the activating examples.

The pooled labels provide information about the semantic neighborhood in which a feature was retrieved, while the activation contexts ground the interpretation in observed model behavior. In practice, this often shifts descriptions toward a more mechanistic level. Broad genre labels may become specific narrative operations, while candidates dominated by formatting patterns, metadata, or other non-semantic regularities are discarded.

\subsubsection{Contrast-Based Causal Tests}%
\label{subsec:contrast-causal-tests}

For each renamed candidate, we construct local token contrasts that test whether the sparse feature contributes to the proposed book-writing mechanism. Each contrast consists of a context, a positive continuation token that matches the proposed mechanism, and a negative continuation token that provides a plausible alternative. This stage complements activation-context evidence with output evidence: recent automated-interpretability work argues that feature descriptions should reflect both what activates a feature and how feature activation affects outputs~\cite{gurarieh2025outputcentric}, while sparse-feature circuit work uses feature-level interventions to identify causally implicated components~\cite{marks2025sparse}. The objective is the corresponding logit difference,
\begin{equation}
J(x) = \ell(t^{+} \mid x) - \ell(t^{-} \mid x),
\end{equation}
where \(t^{+}\) is the mechanism-consistent token and \(t^{-}\) is the contrast token.

We then intervene on the sparse feature by adding its decoded contribution at the relevant position and measuring the resulting change in the contrast objective. For feature \(i\), this gives
\begin{equation}
\Delta J_i =
J\!\left(x; h + D_i\right) -
J\!\left(x; h\right),
\end{equation}
where \(h\) is the residual activation at the intervention site and \(D_i\) is the SAE decoder direction for the sparse feature. A positive \(\Delta J_i\) indicates that activating the sparse feature increases the relative probability of the mechanism-consistent continuation.

These causal tests provide evidence that a proposed sparse feature is not only correlated with the renamed mechanism in retrieved contexts, but can also shift the model toward mechanism-consistent local continuations. We therefore use causal validation as evidence that the sparse feature contributes to the book-writing behavior identified during discovery and context-based renaming.

\subsubsection{Token-Level Evidence Filtering}%
\label{subsec:trace-filtering}

Finally, we audit token-level activation traces for candidates that pass causal validation. For the run reported here, this produced 408 validated traces comprising 49{,}420 token rows. Each trace consists of a local window of generated text surrounding a tested event. For every token in the window, the export records the feature activation, the position used for the causal objective, the tested next-token contrast, and the measured intervention effect.

The purpose of the trace audit is to evaluate whether a proposed feature interpretation is supported by the contexts in which the feature activates. Causal validation establishes that a feature can influence a selected token contrast, whereas trace analysis examines the surrounding activation patterns. Features whose activations consistently occur in contexts compatible with the proposed interpretation are retained, while candidates dominated by formatting artifacts, metadata-like strings, or other unrelated patterns are rejected.

Trace analysis also helps distinguish narrow lexical effects from broader contextual patterns. Some features activate primarily on isolated words or phrases, whereas others remain active across extended narrative passages. These differences affect the level of interpretation assigned to a feature and help determine whether a proposed mechanism name reflects a recurring pattern in the generated text.

The trace audit additionally serves as a final filtering and renaming stage. Features with consistent causal effects may still exhibit heterogeneous activation contexts. In such cases, interpretations are chosen to reflect the recurring patterns observed across traces instead than the most specific label associated with a subset of activations.

\subsection{Sparse Autoencoder Findings}%
\label{subsec:sae-findings}

After label-guided sparse feature retrieval, activation-grounded naming, contrast-based causal tests, and token-level evidence filtering, we retained six recurring book-writing mechanisms. We report support score as \(s=-\Delta_{\mathrm{ablate}}\), so positive values indicate that the sparse feature supports the tested mechanism-consistent token contrast. The retained features fall into three families: hook and conflict features, investigation features, and broader archetype-style prose features.

\paragraph{Hook and conflict features.}
The strongest sparse feature we found supports high-stakes conflict framing in hook-like generated prose. It activates in contexts that foreground danger, factions, survival pressure, difficult choices, or internal struggle. Across 24 validated traces, this feature had a mean support score of \(+0.847\), with 79\% of traces supporting the tested contrast. The objective-level tests were strongest for conflict versus peace \((+1.42)\), followed by danger versus comfort \((+0.91)\) and threat versus home \((+0.21)\). We interpret this feature as a writing mechanism for making a premise feel consequential. Its effect is not merely to select a genre label or to insert isolated conflict words; the supporting traces place the feature in prose that raises stakes, introduces pressure, and frames the story around conflict that demands resolution.

A second high-confidence sparse feature captures a narrower threat-hook mechanism. It activates around immediate peril, accusation, isolation, companion danger, and other forms of short-range threat framing. Across 24 validated traces, it had a mean support score of \(+0.527\), with 83\% of traces supporting the tested contrast. The strongest tested contrast was threat versus home \((+0.78)\), followed by danger versus comfort \((+0.43)\) and conflict versus peace \((+0.37)\). This feature is adjacent to high-stakes conflict framing, but it is more localized. Where the previous feature emphasizes broad conflict and stakes, this feature appears to help construct the immediate hook: a concrete danger, a perilous situation, or a threat-bearing premise that gives the generated text urgency.

\paragraph{Investigation features.}
The hidden-truth investigation feature appears in generated prose about discovery, corruption, concealed evidence, and truth-seeking setup. It is weaker than the two hook-conflict features, but its activation contexts and causal tests point to a coherent investigation mechanism. Across 32 validated traces, it had a mean support score of \(+0.142\), with 62\% of traces supporting the tested contrast. The objective-level pattern was consistently positive but modest: truth versus routine \((+0.26)\), corruption versus friendship \((+0.15)\), conspiracy versus romance \((+0.13)\), and secret versus ordinary \((+0.02)\). We therefore interpret the feature as supporting investigation-adjacent setup language instead than as a broad ``mystery'' or ``conspiracy'' concept. Its strongest evidence comes from contexts where the generated text frames the story around uncovering what is hidden or institutionally obscured.

The secret/uncanny investigation feature occupies a nearby but less stable part of the writing space. It activates around secret discovery, uncanny clues, suspicious institutions, and pressure-to-choose setups. Across 32 validated traces, its mean support score was only \(+0.018\), with 44\% of traces supporting the tested contrast. The central observation is its mixed objective profile. The feature supports secret versus ordinary \((+0.46)\) and corruption versus friendship \((+0.37)\), but is near zero or suppressive for truth versus routine \((-0.02)\) and conspiracy versus romance \((-0.74)\). This suggests a context-sensitive mechanism instead than a simple monotonic investigation variable. The final name is therefore deliberately softer: the feature is associated with secret or uncanny investigation setup, but does not uniformly push every investigation-related contrast in the same direction.

\paragraph{Archetype and ending-pressure features.}
The largest retained feature family is not a narrow semantic feature. It is a broad paragraph-writing mode that appears in explanatory generated prose about premise development, conflict systems, alliances, consequences, and resolution mechanics. Grouping the related sparse features produced 216 validated traces. The aggregate mean support score was small, \(+0.016\), and the supporting fraction was 53\%. This finding separates evidence volume from causal cleanliness. The feature family is clearly associated with a recurring mode of book-writing prose, but it does not behave like a single narrow concept variable. Its traces indicate a broad explanatory mode: the model is elaborating how a story world, conflict structure, or character arc develops over time. We therefore report it as a generic archetype paragraph mode instead than as a specific concept such as strategy, memory, survival, or alliance.

The survival-cost ending pressure feature is a more specific mechanism within archetype-style explanatory prose. It appears around desperate choices, necessary compromise, hard survival, and endings constrained by cost instead than wish fulfillment. Across 64 validated traces, it had a mean support score of \(+0.138\), with 61\% of traces supporting the tested contrast. In the survival/dark-tone objective family, the mean support score was \(+0.23\) across 32 tested instances. We interpret this feature as a context-sensitive ending-pressure mechanism. It does not simply encode ``survival'' in isolation. Instead, it appears to participate in prose where survival is made narratively costly: characters endure loss, accept compromises, or reach endings shaped by constraint. This makes it distinct from the broader archetype paragraph mode, which captures explanatory structure more generally.

\section{Conclusion and Limitations}\label{sec:conclusion}

In this work, we argued that general-purpose assistant models are not optimized for creative writing and presented a training framework designed specifically for book-scale fiction generation. We constructed a hierarchical Planning Scaffold from human-authored public-domain books and trained a language model to reproduce the generation process from prompt to scaffold to chapter text.

Our experiments showed that this specialized training objective substantially improved creative writing performance. Despite having only 14B parameters, the resulting model outperformed substantially larger frontier language models on our writing quality evaluations. We also found that the learned planning procedure transferred beyond the public-domain books from which it was derived. The model was able to generate modern genres and settings that were not directly represented in the training corpus, suggesting that the Planning Scaffold teaches the model how to organize stories while the pretrained model supplies the underlying world knowledge. Transfer to fanfiction was noticeably weaker, however, as these prompts require detailed knowledge of specific characters, relationships, and canon that is only partially captured by the base model.

The interpretability analyses further supported this conclusion. Our attention analysis showed that generated chapter text continued to reference the Planning Scaffold throughout generation, confirming that the scaffold remained an active planning representation instead of being ignored after it was generated. Likewise, the sparse autoencoder analysis identified semantic features corresponding to high-level narrative concepts and story mechanisms, providing complementary evidence that the model learned meaningful internal representations for long-form fiction generation.

At the same time, our consistency evaluations identified the primary limitation of the current approach. Although the model successfully learned the planning procedure, adherence to both the original prompt and the intermediate Planning Scaffold gradually weakened across later generation stages, resulting in reduced long-range consistency. More fundamentally, while the model was trained with a context length of 256k tokens, we found that generation at this scale is not yet reliable: sufficiently long generations eventually enter an unrecoverable repetition spiral, preventing the model from consistently producing complete books at its full training context. Improving long-context consistency together with prompt and scaffold following therefore remains the most important direction for future work.

Overall, our results demonstrate that synthetic Planning Scaffolds derived from human-authored books provide an effective and interpretable framework for training creative writing models. More broadly, they suggest that creative writing benefits from being treated as its own modeling objective instead of as an extension of general-purpose assistant training.

\section*{Acknowledgments}

Research supported with Cloud TPUs from Google's TPU Research Cloud (TRC).

\appendix
\clearpage
\section{Prompt Dataset Construction and Composition}
\label{sec:appendix-prompt-dataset-construction}

The evaluation dataset contains 360 single-turn user requests for long-form book writing. The prompts cover both original fiction and fanfiction and were constructed to vary systematically across genre, premise structure, narrative scope, request style, prompt length, and constraint complexity.

\subsection{Dataset Composition}
\label{sec:appendix-dataset-composition}

The dataset consists of two subsets: original-fiction prompts and fanfiction prompts.

Original-fiction prompts request entirely new narratives that are independent of existing fictional universes. Because no prior fictional context can be assumed, these prompts may specify aspects of the requested work, including the protagonist, setting, central conflict, genre, narrative direction, tone, or stylistic preferences, while leaving other elements intentionally unspecified.

Fanfiction prompts instead assume an existing fictional canon and request a transformation or continuation of that source material. Depending on the prompt family, the request may extend the original story, introduce a counterfactual change, shift narrative perspective, or combine multiple fictional settings or continuities.

The complete dataset is organized into 36 prompt families, each containing 10 prompts. Twenty-four families represent original fiction (240 prompts), while the remaining 12 represent fanfiction (120 prompts). Within each family, prompts vary in premise, narrative scope, protagonist role, requested tone, stylistic preferences, prompt length, and constraint complexity, providing broad coverage of realistic fiction-writing requests while avoiding near-duplicate prompts.

Prompt drafting was assisted by Gemma 3 27B, after which all prompts were manually reviewed and edited before inclusion in the dataset. The review process ensured that prompts were coherent, representative of their assigned prompt family, distinct from other prompts within the same family, and phrased as plausible natural user requests. Prompts containing contradictory instructions, unnecessary repetition, excessive prescriptiveness, or unnatural wording were revised or discarded.

The prompts were designed to resemble realistic requests submitted to large language models for assistance with writing books. Accordingly, they vary not only in narrative content but also in rhetorical form. Some requests are brief and premise-oriented, while others specify characterization, atmosphere, setting, plot direction, prose style, or the desired ending. Prompt length was intentionally varied across the dataset, ranging from 32 to 238 words (median 74; mean 93). Longer prompts generally provide more explicit guidance rather than greater narrative complexity, leaving substantial freedom in scene construction, dialogue, pacing, and the precise development of the narrative.

\subsection{Original-Fiction Prompt Families}
\label{sec:appendix-original-fiction-families}

\paragraph{Composition.}
The original-fiction prompt families are organized into nine broad genre categories. Several categories contain multiple prompt families to capture distinct narrative traditions, premise structures, and storytelling conventions.

\paragraph{Fantasy.}
Fantasy encompasses epic or high fantasy, urban fantasy, portal fantasy or isekai, progression fantasy, cultivation or wuxia, LitRPG or GameLit, dungeon or tower fiction, romantasy, and cozy fantasy. These families span a wide range of speculative settings and narrative structures, from large-scale secondary-world adventures and structured power progression to character-driven stories emphasizing relationships or everyday life within fantastical worlds.

\paragraph{Science fiction.}
Science-fiction prompts include space opera, hard or near-future science fiction, cyberpunk, and superhero or mecha fiction. Together, these families explore technologically advanced societies, scientific innovation, interstellar settings, dystopian futures, and speculative visions of humanity's development.

\paragraph{Romance.}
Romance comprises contemporary romance and dark romance. Contemporary romance emphasizes interpersonal relationships in present-day settings, while dark romance explores morally complex relationships, unequal power dynamics, and psychologically intense conflicts.

\paragraph{Mystery.}
Mystery includes both traditional mystery and cozy mystery. These prompts are organized around investigations and the gradual resolution of hidden information, differing primarily in tone, setting, and narrative intensity.

\paragraph{Thriller.}
Thriller prompts include political or legal thrillers and domestic thrillers. These families build suspense through escalating threats, conspiracies, institutional conflict, or deteriorating personal relationships.

\paragraph{Historical.}
Historical prompts request stories set in earlier historical periods, where the social, political, and cultural context forms an integral part of the narrative and character development.

\paragraph{Horror.}
Horror focuses primarily on psychological horror, emphasizing uncertainty, atmosphere, and escalating emotional or supernatural threats.

\paragraph{Drama.}
Drama centers on contemporary or domestic conflicts driven by interpersonal relationships, family dynamics, and character development.

\paragraph{Comedy.}
Comedy emphasizes humor, satire, and lighthearted social situations, with conflict emerging primarily from misunderstandings, exaggerated personalities, or everyday interactions.

\paragraph{Construction principles.}
The genre categories define the broad narrative context, while the individual prompt families capture distinct storytelling conventions within each genre. Individual prompts vary in protagonist profile, setting, narrative scope, central conflict, tone, stylistic preferences, and intended ending. Across all original-fiction families, the prompts establish a recognizable protagonist, an initial source of narrative tension, and a premise capable of supporting a book-length narrative while leaving substantial freedom in scene construction, dialogue, pacing, and the precise sequence of events.

\subsection{Fanfiction Prompt Families}
\label{sec:appendix-fanfiction-families}

\paragraph{Composition.}
Unlike the original-fiction prompt families, the fanfiction families are defined primarily by the relationship between the requested story and its source material rather than by genre.

The fanfiction families can be organized into four broader transformation classes: temporal extension, counterfactual revision, focalization shift, and cross-context combination. These classes describe the principal operation each prompt performs on inherited fictional material.

\paragraph{Temporal extension.}
Post-canon continuation prompts extend the narrative beyond an established endpoint. Pre-canon prompts move backward to events preceding the principal source narrative, while future-fic prompts place familiar characters, relationships, or institutions at a substantially later stage. These prompts specify the relevant temporal position and the developments that distinguish the requested story from the source.

\paragraph{Counterfactual revision.}
Alternate-universe prompts retain selected characters or relationships while changing the surrounding history, setting, social structure, or governing premise. Canon-divergence prompts identify a point at which an established event occurs differently and request a narrative based on the resulting consequences. Fix-it prompts are organized around preventing, reversing, or repairing a canonical outcome. Time-travel fix-it prompts introduce a temporal mechanism through which characters attempt to produce that change.

\paragraph{Focalization shift.}
Side-character spotlight prompts place a previously peripheral character at the center of the narrative. Villain-point-of-view reinterpretations present established events or conflicts from the perspective of an antagonist, often altering how the source narrative's moral or causal structure is understood. Point-of-view outsider prompts present the fictional setting through a character who stands outside its principal relationships, institutions, or conflicts.

\paragraph{Cross-context combination.}
Crossover prompts bring together characters or settings from multiple sources. Fusion prompts instead apply the structural premise, institutional framework, or world logic of one source to characters or relationships from another. In both cases, the prompt specifies enough of the intended connection to establish a coherent shared premise.

\paragraph{Construction principles.}
Each fanfiction prompt identifies the relevant source material, the inherited characters or institutions central to the request, and the transformation to be applied. The prompts avoid summarizing the complete source narrative and include only the canonical context needed to make the requested transformation clear.

The amount of inherited context varies by family. Continuation and future-fic prompts generally specify the temporal position of the new story. Divergence and fix-it prompts identify the event or condition being changed. Focalization prompts define the new narrative center, while crossover and fusion prompts explain how the source contexts should interact. Across all fanfiction families, the prompts preserve recognizable elements of the source material while establishing a distinct premise capable of supporting a new book-length narrative.

\subsection{Representative Prompts}
\label{sec:appendix-representative-prompts}

The following examples illustrate how genre, premise structure, narrative scope, request length, and explicit constraints appear in natural user-facing language. All examples are drawn from the original-fiction subset and range from compact premise statements to detailed requests that specify plot development, tone, and the intended form of resolution.

\begin{examplepromptquote}[Example 1: comedy or satire]\footnotesize
I want a comic mystery set during a miserable museum donor gala, where a shady but sharp insurance adjuster investigates a missing exhibit, solves the theft, and leaves the larger fraud hanging.
\end{examplepromptquote}

This compact prompt combines comedy with a mystery structure. Despite its brevity, it establishes a contained setting, a protagonist with a distinctive occupation and personality, a central investigation, and an ending that resolves the immediate theft while leaving the broader fraud unresolved.

\begin{examplepromptquote}[Example 2: historical romance or drama]\footnotesize
Write me a historical romance-drama about a foreign-born interpreter at court who becomes essential to a royal marriage treaty meant to prevent a border war. I want the love story tangled in spies, rival factions, hostage customs, formal ceremony, and the terrifying weight of choosing exactly the right words.
\end{examplepromptquote}

This prompt places a romance within a broader diplomatic and political conflict. The protagonist's role as an interpreter connects the personal relationship to the institutional stakes of the premise, while the references to espionage, factional rivalry, ceremony, and linguistic precision establish the intended sources of tension without prescribing a complete plot sequence.

\begin{examplepromptquote}[Example 3: space opera]\footnotesize
I could go for a space opera where customs paperwork is somehow as lethal as a fleet battle because everyone on the station is broke, exhausted, and nursing a grudge. Set it mostly on a collapsing border station powered by cheap coffee, panic, and bad decisions. The main character should be a junior customs administrator trying to recover after publicly blowing a promotion interview, sharp enough to survive the job and ambitious enough to hate needing this second chance. I want inspections, loopholes, illegal cargo, smugglers who know the rules better than the staff, vicious coworker politics, and her former partner showing up as the auditor at the worst possible moment. Make the jokes fast, let the story stay hopeful underneath the chaos, and end with the station saved from the immediate legal disaster while her promotion is still up in the air.
\end{examplepromptquote}

This prompt combines a space-opera setting with workplace comedy and bureaucratic suspense. It specifies the protagonist's professional situation, the institutional mechanisms that generate conflict, an important personal relationship, and the desired balance between humor and underlying optimism. The ending resolves the station's immediate crisis while leaving the protagonist's longer-term ambitions unsettled.

\begin{examplepromptquote}[Example 4: mystery]\footnotesize
I'd like a fantasy about an imperial academy built to turn gifted students into the empire's next ministers, judges, governors, and court strategists. The main character should be a dorm captain who never came there hungry for rank. Other students treat authority as the whole point of the place, but she mostly cares about keeping the younger and weaker students in her hall safe. The mystery should begin with exam answers being leaked and disciplinary records disappearing, so it first feels like an ugly academy scandal that ambitious students, old families, and powerful teachers all have reasons to hide. Then students start dying under suspicious circumstances, and the dorm captain has to stop pretending the problem is contained. Have her investigate through the normal machinery of the school: class rankings, mentor chains, faction alliances, old punishments, secret societies, scholarship debts, and the way one student's future can be turned into a weapon against someone else. The deeper truth should reveal that someone is using the academy as a miniature rehearsal for an imperial coup, testing loyalties, pressure points, and methods before the real succession conflict begins. Make the prose clear and grounded about what the investigation costs her, especially when friends become suspects, tools, or bargaining pieces. The ending should prevent something much worse, but only because she accepts a patron's bargain and commits a betrayal that leaves her unsure what kind of leader she has just become.
\end{examplepromptquote}

This prompt places a mystery within a fantasy academy and develops the investigation through the institution's ordinary systems of ranking, mentorship, discipline, and patronage. It defines the protagonist's protective motivation, the escalation from academic misconduct to murder, the larger political significance of the conspiracy, and the personal cost of resolving it. Although the prompt outlines a substantial narrative trajectory, it leaves the individual clues, suspects, scenes, and intermediate discoveries unspecified.

\clearpage
\makeatletter
{\catcode`\ =\active
\gdef\VerbatimBreakableSpaces{%
  \catcode`\ \active
  \def {\leavevmode\hskip\fontdimen2\font\relax}%
}}
\ifx\Verbatim\@undefined
  \begingroup
    \catcode`\|=0 \catcode`\[=1 \catcode`\]=2 \catcode`\{=12 \catcode`\}=12 \catcode`\\=12
    |gdef|@xVerbatim#1\end{Verbatim}[#1|end[Verbatim]]
  |endgroup
  \def\Verbatim{\@ifnextchar[{\Verbatim@opt}{\Verbatim@opt[]}}%
  \def\Verbatim@opt[#1]{%
    \par\begingroup\scriptsize\@verbatim\frenchspacing\VerbatimBreakableSpaces\raggedright\@xVerbatim
  }%
  \def\endVerbatim{\if@newlist \leavevmode\fi\endtrivlist\endgroup}%
\fi
\makeatother

\section{Full Benchmark Results}
\label{app:full-benchmark-results}

The full benchmark results are shown for all evaluated models. Each score is computed with the same narrative diagnostic evaluation suite.

\begin{table}[H]
\centering
\small
\resizebox{\textwidth}{!}{%
\begin{tabular}{lrrrrrr}
\toprule
\textbf{Score} & \textbf{Our model} & \textbf{GPT-5.5 xhigh} & \textbf{Claude Opus 4.8} & \textbf{DeepSeek V4 Pro} & \textbf{GLM-5.1} & \textbf{Kimi K2.6} \\
\midrule
Setup becomes a live problem & 34.4\% & 18.9\% & 22.8\% & \textbf{43.7\%} & 28.1\% & 36.9\% \\
Same goal stays active & 31.4\% & 18.6\% & 37.8\% & 42.8\% & 36.1\% & \textbf{44.0\%} \\
Blocked goal triggers response & \textbf{74.0\%} & 25.3\% & 51.4\% & 46.9\% & 29.7\% & 63.8\% \\
Mystery remains unresolved & \textbf{32.3\%} & 7.2\% & 6.4\% & 15.9\% & 8.6\% & 7.8\% \\
Initial situation is clear & 80.3\% & 78.6\% & 86.9\% & \textbf{94.2\%} & 91.1\% & 83.7\% \\
Viewpoint control holds & 88.5\% & 86.9\% & 91.7\% & 91.1\% & \textbf{93.1\%} & 85.1\% \\
Exposition is motivated by local need & \textbf{15.8\%} & 4.2\% & 3.6\% & 9.2\% & 5.3\% & 14.2\% \\
\bottomrule
\end{tabular}
}%
\caption{Full benchmark results; bold marks the highest score in each row.}
\label{tab:full-benchmark-results}
\end{table}

\section{Operationalization of Narrative Diagnostic Scores}
\label{app:score-operationalization}

The evaluation suite applies the narrative diagnostics to the First Chapter Text. It uses the Gemma4 31B model to extract structured, text-grounded features from that chapter, including goals, blocked actions, unresolved questions, viewpoint spans, orientation anchors, local information needs, and exposition units. The model does not assign scores or make direct quality judgments.

The extracted features are then analyzed by a deterministic, rule-based harness, which determines whether the First Chapter Text satisfies the benchmarked narrative rules. Each diagnostic returns a binary pass/fail value together with selected supporting evidence. Unless a narrower window is specified, extraction is limited to the first twelve paragraphs; smaller windows are used for diagnostics focused on the immediate chapter opening.

\subsection{Setup Becomes a Live Problem}
\label{app:setup-live-problem}

\subsubsection{Measurement target}
The \emph{Setup becomes a live problem} diagnostic asks whether the preview establishes a concrete setup package and whether the First Chapter Text converts that setup into a local live problem. A live problem is operationalized as a question or unresolved information gap that enters an exchange while a constraint is active. The score is defined for launch rows.

A passing case must satisfy two broad requirements. First, the preview must contain all three parts of a setup package: at least one participant, at least one source of pressure, and at least one directional element such as a goal, need, obligation, mission, question, exchange, or reason for action. Second, the First Chapter Text must contain at least one local problem window in paragraphs 1--12, and at least one preview pressure or direction element must be reused, activated, applied, or made consequential inside such a window. Participant evidence is required to show that the setup is grounded in a person, role, group, or relationship, but participant evidence alone cannot satisfy the bridge condition.

\subsubsection{Extraction procedure}
The setup score uses three extraction passes. The preview pass identifies a compact setup package made up of participant evidence, pressure evidence, and direction evidence. The First Chapter Text pass identifies local problem windows in paragraphs 1--12, where an unresolved question or information gap is paired with an exchange and an active constraint. The comparison pass then links preview pressure or direction evidence to those local windows when the relevant preview material is reused, activated, applied, or made consequential in the First Chapter Text.

\subsubsection{Deterministic scoring rule}
The score is computed as a sequence of gates. The row first fails if the preview contains no keyed evidence for a person, role, group, or relationship-bearing party. It next fails if the preview contains no pressure evidence, because the setup then has no obstacle, cost, risk, rule, refusal, scarcity, or comparable constraint. It then fails if the preview contains no direction evidence, because the setup then lacks a goal, need, obligation, mission, question, exchange, or other reason for action. The final preliminary gate checks whether the First Chapter Text contains at least one local problem window; without such a window, the setup has not become an active problem in the chapter opening.

If these gates pass, the bridge links are inspected in the order returned by the comparison pass. The score passes when the first usable link names an existing local window and at least one existing preview pressure or direction element. If all setup and window evidence is present but no usable bridge exists, the score fails because the preview setup remains unconnected to the local problem in the First Chapter Text.

The emitted evidence follows the same gate structure. When a gate fails, the output reports the relevant remaining counts so that the missing component can be identified. A passing result reports the selected participant evidence, selected preview evidence, selected local window, and selected relation. If the setup gates and window gate pass but no bridge exists, the emitted evidence reports the counts for participants, pressure elements, direction elements, and local windows.

\subsubsection{Implementation details}
Counts are based on distinct artifact identifiers after evidence items or windows are keyed; duplicate identifiers do not increase a count. Bridge candidates are limited to pressure and direction evidence from the preview; participant evidence is required for setup grounding but cannot by itself form the bridge. If the same identifier appears in both the pressure and direction evidence, the pressure version is prioritized for emitted evidence. A comparison link is ignored if it points to a window that was not extracted, or if it does not name any extracted pressure or direction item. When a single comparison link names several qualifying items, the first qualifying item supplies the preview evidence. The participant evidence attached to a passing result is the first extracted participant item, even when that participant is not named in the bridge link.

\subsubsection{LLM extraction prompts}

\promptlistingtitle{System prompt: Preview setup package extraction.}
\begin{Verbatim}[fontsize=\scriptsize,breaklines=true,breakanywhere=true,frame=single,framerule=0.8pt,framesep=8pt,rulecolor=\color{promptframe}]
# Task
Extract the preview setup package that could later become a live problem in the First Chapter Text.

# Rules
- Use only explicit evidence from the supplied text or supplied artifact IDs.
- Do not infer from genre knowledge or likely later story events.
- Do not return a metric score, pass/fail, verdict, rating, or quality judgment.
- Keep evidence fields short: one clause or sentence, not a full paragraph.
- Prefer compact high-signal artifact lists over exhaustive inventories.
- Empty arrays are correct when the text has no evidence for a requested artifact.
- Return corrected JSON only if asked to repair a previous response.
- Extract people as named characters, roles, groups, or relationship-bearing parties.
- Extract pressure cues as obstacles, costs, risks, refusals, scarcity, rules, threats, deadlines, debts, or burdens.
- Extract direction cues as goals, needs, obligations, missions, questions, exchanges, or reasons for action.
- Use only the preview segment.
- Use cue IDs like pp1, pr1, pd1.

# Output Rules
Return exactly one JSON object matching the Required JSON Schema below. Do not wrap it in Markdown.

# Required JSON Schema
```json
{
  "additionalProperties": false,
  "properties": {
    "direction_cues": {
      "items": {
        "additionalProperties": false,
        "properties": {
          "cue_id": {
            "maxLength": 96,
            "minLength": 1,
            "type": "string"
          },
          "direction_kind": {
            "enum": [
              "goal",
              "question",
              "exchange",
              "obligation",
              "mission",
              "need"
            ],
            "type": "string"
          },
          "evidence": {
            "maxLength": 220,
            "minLength": 1,
            "type": "string"
          },
          "label": {
            "maxLength": 120,
            "minLength": 1,
            "type": "string"
          }
        },
        "required": [
          "cue_id",
          "direction_kind",
          "label",
          "evidence"
        ],
        "type": "object"
      },
      "maxItems": 20,
      "minItems": 0,
      "type": "array"
    },
    "people": {
      "items": {
        "additionalProperties": false,
        "properties": {
          "cue_id": {
            "maxLength": 96,
            "minLength": 1,
            "type": "string"
          },
          "evidence": {
            "maxLength": 220,
            "minLength": 1,
            "type": "string"
          },
          "label": {
            "maxLength": 120,
            "minLength": 1,
            "type": "string"
          },
          "person_kind": {
            "enum": [
              "named_person",
              "role",
              "group",
              "relationship"
            ],
            "type": "string"
          }
        },
        "required": [
          "cue_id",
          "person_kind",
          "label",
          "evidence"
        ],
        "type": "object"
      },
      "maxItems": 16,
      "minItems": 0,
      "type": "array"
    },
    "pressure_cues": {
      "items": {
        "additionalProperties": false,
        "properties": {
          "cue_id": {
            "maxLength": 96,
            "minLength": 1,
            "type": "string"
          },
          "evidence": {
            "maxLength": 220,
            "minLength": 1,
            "type": "string"
          },
          "label": {
            "maxLength": 120,
            "minLength": 1,
            "type": "string"
          },
          "pressure_kind": {
            "enum": [
              "demand",
              "obstacle",
              "cost",
              "deadline",
              "risk",
              "refusal",
              "forced_choice",
              "social_conflict",
              "blocked_attempt",
              "revelation",
              "resource_limit",
              "consequence"
            ],
            "type": "string"
          }
        },
        "required": [
          "cue_id",
          "pressure_kind",
          "label",
          "evidence"
        ],
        "type": "object"
      },
      "maxItems": 20,
      "minItems": 0,
      "type": "array"
    }
  },
  "required": [
    "people",
    "pressure_cues",
    "direction_cues"
  ],
  "type": "object"
}
```
\end{Verbatim}
\par\medskip

\promptlistingtitle{System prompt: First Chapter Text local-problem window extraction.}
\begin{Verbatim}[fontsize=\scriptsize,breaklines=true,breakanywhere=true,frame=single,framerule=0.8pt,framesep=8pt,rulecolor=\color{promptframe}]
# Task
Extract local problem windows in the First Chapter Text where a question or information gap becomes an exchange under constraint.

# Rules
- Use only explicit evidence from the supplied text or supplied artifact IDs.
- Do not infer from genre knowledge or likely later story events.
- Do not return a metric score, pass/fail, verdict, rating, or quality judgment.
- Keep evidence fields short: one clause or sentence, not a full paragraph.
- Prefer compact high-signal artifact lists over exhaustive inventories.
- Empty arrays are correct when the text has no evidence for a requested artifact.
- Return corrected JSON only if asked to repair a previous response.
- A window is anchored by one First Chapter Text paragraph that raises an explicit question or concrete unresolved information gap.
- The anchor paragraph or the immediately following paragraph must contain an exchange: dialogue, request, demand, warning, answer, confrontation, or negotiation.
- The anchor paragraph or the immediately following paragraph must contain a constraint: obstacle, cost, risk, debt, scarcity, rule, refusal, deadline, or burden.
- Use only First Chapter Text paragraphs 1 through 12.
- Use support_paragraph_indices for the one or two paragraphs containing the extracted evidence.
- Use window IDs like w1, w2, w3.

# Output Rules
Return exactly one JSON object matching the Required JSON Schema below. Do not wrap it in Markdown.

# Required JSON Schema
```json
{
  "additionalProperties": false,
  "properties": {
    "local_problem_windows": {
      "items": {
        "additionalProperties": false,
        "properties": {
          "constraint_evidence": {
            "maxLength": 220,
            "minLength": 1,
            "type": "string"
          },
          "exchange_evidence": {
            "maxLength": 220,
            "minLength": 1,
            "type": "string"
          },
          "question_or_gap_evidence": {
            "maxLength": 220,
            "minLength": 1,
            "type": "string"
          },
          "question_paragraph_index": {
            "maximum": 12,
            "minimum": 1,
            "type": "integer"
          },
          "support_paragraph_indices": {
            "items": {
              "maximum": 12,
              "minimum": 1,
              "type": "integer"
            },
            "maxItems": 2,
            "minItems": 1,
            "type": "array"
          },
          "window_id": {
            "maxLength": 96,
            "minLength": 1,
            "type": "string"
          }
        },
        "required": [
          "window_id",
          "question_paragraph_index",
          "support_paragraph_indices",
          "question_or_gap_evidence",
          "exchange_evidence",
          "constraint_evidence"
        ],
        "type": "object"
      },
      "maxItems": 20,
      "minItems": 0,
      "type": "array"
    }
  },
  "required": [
    "local_problem_windows"
  ],
  "type": "object"
}
```
\end{Verbatim}
\par\medskip

\promptlistingtitle{System prompt: Preview-to-window bridge comparison.}
\begin{Verbatim}[fontsize=\scriptsize,breaklines=true,breakanywhere=true,frame=single,framerule=0.8pt,framesep=8pt,rulecolor=\color{promptframe}]
# Task
Compare preview setup cues to local problem windows in the First Chapter Text.

# Rules
- Use only explicit evidence from the supplied text or supplied artifact IDs.
- Do not infer from genre knowledge or likely later story events.
- Do not return a metric score, pass/fail, verdict, rating, or quality judgment.
- Keep evidence fields short: one clause or sentence, not a full paragraph.
- Prefer compact high-signal artifact lists over exhaustive inventories.
- Empty arrays are correct when the text has no evidence for a requested artifact.
- Return corrected JSON only if asked to repair a previous response.
- Use only supplied preview cue IDs and window IDs.
- Create a link when a preview cue is reused, activated, applied, or made consequential in a local problem window.
- Do not create links from broad genre similarity, mood, or shared setting alone.
- Do not create new preview cues or windows.
- Use link IDs like lw1, lw2, lw3.

# Output Rules
Return exactly one JSON object matching the Required JSON Schema below. Do not wrap it in Markdown.

# Required JSON Schema
```json
{
  "additionalProperties": false,
  "properties": {
    "links": {
      "items": {
        "additionalProperties": false,
        "properties": {
          "evidence": {
            "maxLength": 220,
            "minLength": 1,
            "type": "string"
          },
          "link_id": {
            "maxLength": 96,
            "minLength": 1,
            "type": "string"
          },
          "preview_cue_ids": {
            "items": {
              "maxLength": 96,
              "minLength": 1,
              "type": "string"
            },
            "maxItems": 4,
            "minItems": 1,
            "type": "array"
          },
          "relation_basis": {
            "enum": [
              "explicit_reuse",
              "causal_phrase",
              "same_named_entity",
              "role_continuity",
              "rule_application",
              "relationship_obligation",
              "same_resource_or_constraint"
            ],
            "type": "string"
          },
          "relation_kind": {
            "enum": [
              "same_obligation",
              "same_constraint",
              "same_question",
              "same_relationship_pressure",
              "same_goal_pressure",
              "same_exchange_pressure"
            ],
            "type": "string"
          },
          "window_id": {
            "maxLength": 96,
            "minLength": 1,
            "type": "string"
          }
        },
        "required": [
          "link_id",
          "preview_cue_ids",
          "window_id",
          "relation_kind",
          "relation_basis",
          "evidence"
        ],
        "type": "object"
      },
      "maxItems": 24,
      "minItems": 0,
      "type": "array"
    }
  },
  "required": [
    "links"
  ],
  "type": "object"
}
```
\end{Verbatim}
\par\medskip

\subsection{Same Goal Stays Active}
\label{app:same-goal-active}

\subsubsection{Measurement target}
The \emph{Same goal stays active} diagnostic measures whether the opening of the generated First Chapter Text gives a character an explicit motivation, supports that motivation with early movement, and then keeps the same motivation active in later action in the First Chapter Text. The diagnostic is not satisfied by a character merely appearing again, by the repetition of a setting or mood, or by background premise alone. Later activity must continue the earlier motivation as the same goal, a subgoal, a revision, an obstacle, a delay, a consequence, or an inquiry continuation.

\subsubsection{Extraction procedure}
The same-goal score uses three extraction passes. The first pass inspects paragraphs 1--4 for early goal frames: a character or role owns a motivation, and the text also supplies early support in the form of a constraint, obstacle, or attempt. The second pass inspects paragraphs 5--12 for later activities that could show that motivation continuing to organize action. The comparison pass labels whether a later activity continues the earlier motivation, revises it, delays it, blocks it, follows from it, or carries forward the same inquiry. The deterministic scorer uses only the filtered links described in the next subsection.

\subsubsection{Deterministic scoring rule}
The active-link filter requires three conditions. The later activity must express a substantive continuation of the early motivation, rather than merely involving the same character or the same setting. The stated basis for the link must be text-supported rather than only topical or contextual. Finally, the later actor must either be the original goal owner or an ally or agent acting on that owner's motivation.

For each early goal, the scorer gathers all later activities that survive this active-link filter. It then builds a set of paragraph indices from the goal paragraph, the support paragraph, and all linked later activity paragraphs. The score passes if any early goal has at least one active linked later activity, at least three distinct paragraph indices in this set, and at least one paragraph index of 5 or greater. Because the paragraph threshold uses distinct indices, an early goal and its support in the same paragraph plus one later activity in one later paragraph yields only two distinct paragraphs and does not pass.

The score fails when no early supported goal is extracted, when an early goal exists but no later goal-directed activity is extracted, or when early and later artifacts exist but no early goal has an active continuity chain meeting the paragraph rule.

\subsubsection{LLM extraction prompts}

\promptlistingtitle{System prompt: Early supported-goal extraction.}
\begin{Verbatim}[fontsize=\scriptsize,breaklines=true,breakanywhere=true,frame=single,framerule=0.8pt,framesep=8pt,rulecolor=\color{promptframe}]
# Task
Extract early character-owned motivations that already begin to organize story movement.

# Rules
- Use only explicit evidence from the supplied text or supplied artifact IDs.
- Do not infer from genre knowledge or likely later story events.
- Do not return a metric score, pass/fail, verdict, rating, or quality judgment.
- Keep evidence fields short: one clause or sentence, not a full paragraph.
- Prefer compact high-signal artifact lists over exhaustive inventories.
- Empty arrays are correct when the text has no evidence for a requested artifact.
- Return corrected JSON only if asked to repair a previous response.
- Use First Chapter Text paragraphs 1 through 4 only.
- A frame needs an explicit owner plus a goal, desire, obligation, need, plan, or inquiry.
- The same frame also needs early support: a constraint, obstacle, or attempt that makes the motivation explain what happens next.
- Do not extract static characterization, mood, premise, or world background without a concrete motivation.
- Failed or unsatisfied motivations are valid when they remain able to explain later action.
- Return at most two frames: keep only the clearest motivations with concrete early support.
- Use goal IDs like g1, g2, g3.

# Output Rules
Return exactly one JSON object matching the Required JSON Schema below. Do not wrap it in Markdown.

# Required JSON Schema
```json
{
  "additionalProperties": false,
  "properties": {
    "early_goal_frames": {
      "items": {
        "additionalProperties": false,
        "properties": {
          "goal_evidence": {
            "maxLength": 100,
            "minLength": 1,
            "type": "string"
          },
          "goal_id": {
            "maxLength": 96,
            "minLength": 1,
            "type": "string"
          },
          "goal_paragraph_index": {
            "maximum": 4,
            "minimum": 1,
            "type": "integer"
          },
          "motivation_kind": {
            "enum": [
              "goal",
              "desire",
              "obligation",
              "need",
              "plan",
              "inquiry"
            ],
            "type": "string"
          },
          "motivation_summary": {
            "maxLength": 150,
            "minLength": 1,
            "type": "string"
          },
          "owner_text": {
            "maxLength": 120,
            "minLength": 1,
            "type": "string"
          },
          "support_evidence": {
            "maxLength": 100,
            "minLength": 1,
            "type": "string"
          },
          "support_kind": {
            "enum": [
              "constraint",
              "obstacle",
              "attempt"
            ],
            "type": "string"
          },
          "support_paragraph_index": {
            "maximum": 4,
            "minimum": 1,
            "type": "integer"
          }
        },
        "required": [
          "goal_id",
          "owner_text",
          "motivation_kind",
          "motivation_summary",
          "goal_paragraph_index",
          "support_kind",
          "support_paragraph_index",
          "goal_evidence",
          "support_evidence"
        ],
        "type": "object"
      },
      "maxItems": 2,
      "minItems": 0,
      "type": "array"
    }
  },
  "required": [
    "early_goal_frames"
  ],
  "type": "object"
}
```
\end{Verbatim}
\par\medskip

\promptlistingtitle{System prompt: Later goal-directed activity extraction.}
\begin{Verbatim}[fontsize=\scriptsize,breaklines=true,breakanywhere=true,frame=single,framerule=0.8pt,framesep=8pt,rulecolor=\color{promptframe}]
# Task
Extract later activities that could show an earlier motivation continuing to organize action.

# Rules
- Use only explicit evidence from the supplied text or supplied artifact IDs.
- Do not infer from genre knowledge or likely later story events.
- Do not return a metric score, pass/fail, verdict, rating, or quality judgment.
- Keep evidence fields short: one clause or sentence, not a full paragraph.
- Prefer compact high-signal artifact lists over exhaustive inventories.
- Empty arrays are correct when the text has no evidence for a requested artifact.
- Return corrected JSON only if asked to repair a previous response.
- Use First Chapter Text paragraphs 5 through 12 only.
- Extract choices, attempts, obstacles, delays, revisions, consequences, inquiry pursuit, partial answers, and deferrals.
- Do not decide whether an activity continues an earlier motivation.
- Do not extract passive description unless it changes, blocks, delays, revises, or advances a motivation.
- Return at most five activities: keep only the clearest later activity cues.
- Use activity IDs like a1, a2, a3.

# Output Rules
Return exactly one JSON object matching the Required JSON Schema below. Do not wrap it in Markdown.

# Required JSON Schema
```json
{
  "additionalProperties": false,
  "properties": {
    "later_goal_activities": {
      "items": {
        "additionalProperties": false,
        "properties": {
          "activity_id": {
            "maxLength": 96,
            "minLength": 1,
            "type": "string"
          },
          "activity_kind": {
            "enum": [
              "choice",
              "attempt",
              "obstacle",
              "delay",
              "revision",
              "consequence",
              "inquiry_pursuit",
              "partial_answer",
              "deferral"
            ],
            "type": "string"
          },
          "activity_summary": {
            "maxLength": 150,
            "minLength": 1,
            "type": "string"
          },
          "actor_text": {
            "maxLength": 120,
            "minLength": 1,
            "type": "string"
          },
          "evidence": {
            "maxLength": 100,
            "minLength": 1,
            "type": "string"
          },
          "paragraph_index": {
            "maximum": 12,
            "minimum": 5,
            "type": "integer"
          }
        },
        "required": [
          "activity_id",
          "paragraph_index",
          "activity_kind",
          "actor_text",
          "activity_summary",
          "evidence"
        ],
        "type": "object"
      },
      "maxItems": 5,
      "minItems": 0,
      "type": "array"
    }
  },
  "required": [
    "later_goal_activities"
  ],
  "type": "object"
}
```
\end{Verbatim}
\par\medskip

\promptlistingtitle{System prompt: Goal-to-activity continuity comparison.}
\begin{Verbatim}[fontsize=\scriptsize,breaklines=true,breakanywhere=true,frame=single,framerule=0.8pt,framesep=8pt,rulecolor=\color{promptframe}]
# Task
Compare early motivations to later activities and label their text-supported relation.

# Rules
- Use only explicit evidence from the supplied text or supplied artifact IDs.
- Do not infer from genre knowledge or likely later story events.
- Do not return a metric score, pass/fail, verdict, rating, or quality judgment.
- Keep evidence fields short: one clause or sentence, not a full paragraph.
- Prefer compact high-signal artifact lists over exhaustive inventories.
- Empty arrays are correct when the text has no evidence for a requested artifact.
- Return corrected JSON only if asked to repair a previous response.
- Use only the supplied goal IDs and activity IDs.
- A relation needs more than the same setting, tone, genre frame, or the same person simply being present.
- Allow the motivation to continue as the same goal, a subgoal, a revision, an obstacle, a delay, a consequence, or an inquiry continuation when the text supports that relation.
- A later activity can continue a failed or unsatisfied motivation if that motivation still explains the action, delay, revision, obstacle, or consequence.
- Use same_owner when the later actor is the early goal owner.
- Use ally_or_agent only when the later actor is acting for the early owner's stated motivation.
- Use owner_activity_only when the same owner acts later but the later activity does not serve the earlier motivation.
- Return at most one strongest descriptive link for each early motivation.
- Do not decide whether the metric succeeds.
- Use link IDs like l1, l2, l3.

# Output Rules
Return exactly one JSON object matching the Required JSON Schema below. Do not wrap it in Markdown.

# Required JSON Schema
```json
{
  "additionalProperties": false,
  "properties": {
    "continuity_links": {
      "items": {
        "additionalProperties": false,
        "properties": {
          "activity_id": {
            "maxLength": 96,
            "minLength": 1,
            "type": "string"
          },
          "evidence": {
            "maxLength": 100,
            "minLength": 1,
            "type": "string"
          },
          "goal_id": {
            "maxLength": 96,
            "minLength": 1,
            "type": "string"
          },
          "link_id": {
            "maxLength": 96,
            "minLength": 1,
            "type": "string"
          },
          "owner_relation": {
            "enum": [
              "same_owner",
              "ally_or_agent",
              "different_owner",
              "unclear"
            ],
            "type": "string"
          },
          "relation": {
            "enum": [
              "same_motivation",
              "subgoal_or_revision",
              "obstacle_to_motivation",
              "consequence_of_motivation",
              "delay_of_motivation",
              "inquiry_continuation",
              "owner_activity_only",
              "shared_context_only",
              "unrelated",
              "unclear"
            ],
            "type": "string"
          },
          "relation_basis": {
            "enum": [
              "explicit_reference",
              "same_desired_outcome",
              "same_plan_or_inquiry",
              "same_object",
              "same_obstacle",
              "consequence_of_attempt",
              "same_owner_motivation",
              "repeated_question",
              "owner_activity_only",
              "shared_context_only"
            ],
            "type": "string"
          }
        },
        "required": [
          "link_id",
          "goal_id",
          "activity_id",
          "relation",
          "relation_basis",
          "owner_relation",
          "evidence"
        ],
        "type": "object"
      },
      "maxItems": 6,
      "minItems": 0,
      "type": "array"
    }
  },
  "required": [
    "continuity_links"
  ],
  "type": "object"
}
```
\end{Verbatim}
\par\medskip

\subsection{Blocked Goal Triggers Response}
\label{app:blocked-goal-response}

\subsubsection{Measurement target}
The \emph{Blocked goal triggers response} diagnostic is a launch-row score that asks whether the preview promises a blocked goal and whether the first four paragraphs of the First Chapter Text convert early goal pressure into visible action. It is not a general prose-quality score and it is not a semantic continuity matcher between preview and First Chapter Text. The preview requirement gates the score, but the First Chapter Text goal and First Chapter Text response do not have to share the same actor, goal text, or blocker as the preview package.

\subsubsection{Extraction procedure}
The blocked-goal score uses two extraction passes. The preview pass extracts compact blocked-goal packages consisting of an actor, a goal or inquiry, and a blocking pressure. The First Chapter Text pass extracts early goals and active responses from paragraphs 1--4. These artifacts provide evidence that a promised obstacle has been converted into early action, but the final computation does not require a semantic match between the preview package and the First Chapter Text goal or response.

\subsubsection{Deterministic scoring rule}
The score passes when all four conditions are true: at least one preview blocked-goal package is present; at least one early First Chapter Text goal appears in paragraphs 1--4; at least one early active response appears in paragraphs 1--4; and at least one response is in the same paragraph as a goal or in a later paragraph than that goal. A response in an earlier paragraph than every extracted goal does not satisfy the ordering requirement, but a response in the same paragraph as the selected goal is allowed.

The deterministic procedure applies ordered gates. If no blocked-goal package is extracted from the preview, the row fails because the launch has not promised a goal under pressure. If a preview package exists but no early goal is extracted from the First Chapter Text, the row fails because the chapter opening has not supplied a goal to organize action. If a preview package and an early goal exist but no early active response is extracted, the row fails because the blocked goal has not produced action. If goals and responses both exist, goals are considered in ascending paragraph order; for the first goal that has any response in the same or a later paragraph, the earliest such response in ascending paragraph order is selected. If no such ordered pair exists, the row fails because all responses occur before the available goals. If an ordered pair exists, the row passes because early goal pressure has produced an active response.

\subsubsection{LLM extraction prompts}

\promptlistingtitle{System prompt: Preview blocked-goal package extraction.}
\begin{Verbatim}[fontsize=\scriptsize,breaklines=true,breakanywhere=true,frame=single,framerule=0.8pt,framesep=8pt,rulecolor=\color{promptframe}]
# Task
Extract preview blocked-goal packages.

# Rules
- Use only explicit evidence from the supplied text or supplied artifact IDs.
- Do not infer from genre knowledge or likely later story events.
- Do not return a metric score, pass/fail, verdict, rating, or quality judgment.
- Keep evidence fields short: one clause or sentence, not a full paragraph.
- Prefer compact high-signal artifact lists over exhaustive inventories.
- Empty arrays are correct when the text has no evidence for a requested artifact.
- Return corrected JSON only if asked to repair a previous response.
- Use preview text only.
- A package needs an actor, a concrete goal, desire, obligation, need, or inquiry, and a blocker, cost, risk, rule, refusal, shortage, deadline, information gap, opposition, burden, threat, uncertainty, or unmet need.
- Do not extract genre premise or mood unless the preview states what someone wants or must do and what blocks it.
- Return at most three packages: keep only the clearest blocked goals.
- Use preview IDs like p1, p2, p3.

# Output Rules
Return exactly one JSON object matching the Required JSON Schema below. Do not wrap it in Markdown.

# Required JSON Schema
```json
{
  "additionalProperties": false,
  "properties": {
    "preview_blocked_goals": {
      "items": {
        "additionalProperties": false,
        "properties": {
          "actor": {
            "maxLength": 120,
            "minLength": 1,
            "type": "string"
          },
          "block_evidence": {
            "maxLength": 100,
            "minLength": 1,
            "type": "string"
          },
          "block_kind": {
            "enum": [
              "refusal",
              "scarcity",
              "rule",
              "risk",
              "cost",
              "deadline",
              "opposition",
              "physical_barrier",
              "information_gap",
              "social_pressure",
              "burden",
              "threat",
              "uncertainty",
              "unmet_need",
              "other_pressure",
              "other"
            ],
            "type": "string"
          },
          "block_text": {
            "maxLength": 140,
            "minLength": 1,
            "type": "string"
          },
          "goal_evidence": {
            "maxLength": 100,
            "minLength": 1,
            "type": "string"
          },
          "goal_kind": {
            "enum": [
              "obtain",
              "escape",
              "protect",
              "learn",
              "reach",
              "persuade",
              "hide",
              "obligation",
              "inquiry",
              "other"
            ],
            "type": "string"
          },
          "goal_text": {
            "maxLength": 140,
            "minLength": 1,
            "type": "string"
          },
          "preview_id": {
            "maxLength": 96,
            "minLength": 1,
            "type": "string"
          }
        },
        "required": [
          "preview_id",
          "actor",
          "goal_kind",
          "goal_text",
          "block_kind",
          "block_text",
          "goal_evidence",
          "block_evidence"
        ],
        "type": "object"
      },
      "maxItems": 3,
      "minItems": 0,
      "type": "array"
    }
  },
  "required": [
    "preview_blocked_goals"
  ],
  "type": "object"
}
```
\end{Verbatim}
\par\medskip

\promptlistingtitle{System prompt: Early goal-and-response extraction.}
\begin{Verbatim}[fontsize=\scriptsize,breaklines=true,breakanywhere=true,frame=single,framerule=0.8pt,framesep=8pt,rulecolor=\color{promptframe}]
# Task
Extract early goals and active responses.

# Rules
- Use only explicit evidence from the supplied text or supplied artifact IDs.
- Do not infer from genre knowledge or likely later story events.
- Do not return a metric score, pass/fail, verdict, rating, or quality judgment.
- Keep evidence fields short: one clause or sentence, not a full paragraph.
- Prefer compact high-signal artifact lists over exhaustive inventories.
- Empty arrays are correct when the text has no evidence for a requested artifact.
- Return corrected JSON only if asked to repair a previous response.
- Use First Chapter Text paragraphs 1 through 4 only.
- Extract goals, desires, obligations, needs, plans, and inquiries as goals.
- Extract only active responses: attempts, strategy switches, resource seeking, negotiation, confrontation, evasion, investigation, accepting a cost, goal revision, recruiting help, movement, commitment, or handoff.
- Do not extract emotion-only, waiting-only, or repeated-without-change reactions as responses.
- Use IDs like g1 and r1.

# Output Rules
Return exactly one JSON object matching the Required JSON Schema below. Do not wrap it in Markdown.

# Required JSON Schema
```json
{
  "additionalProperties": false,
  "properties": {
    "goals": {
      "items": {
        "additionalProperties": false,
        "properties": {
          "actor": {
            "maxLength": 100,
            "minLength": 1,
            "type": "string"
          },
          "evidence": {
            "maxLength": 90,
            "minLength": 1,
            "type": "string"
          },
          "goal_id": {
            "maxLength": 96,
            "minLength": 1,
            "type": "string"
          },
          "goal_kind": {
            "enum": [
              "obtain",
              "escape",
              "protect",
              "learn",
              "reach",
              "persuade",
              "hide",
              "obligation",
              "inquiry",
              "other"
            ],
            "type": "string"
          },
          "goal_text": {
            "maxLength": 140,
            "minLength": 1,
            "type": "string"
          },
          "paragraph_index": {
            "maximum": 4,
            "minimum": 1,
            "type": "integer"
          }
        },
        "required": [
          "goal_id",
          "actor",
          "goal_kind",
          "goal_text",
          "paragraph_index",
          "evidence"
        ],
        "type": "object"
      },
      "maxItems": 8,
      "minItems": 0,
      "type": "array"
    },
    "responses": {
      "items": {
        "additionalProperties": false,
        "properties": {
          "actor": {
            "maxLength": 100,
            "minLength": 1,
            "type": "string"
          },
          "evidence": {
            "maxLength": 90,
            "minLength": 1,
            "type": "string"
          },
          "paragraph_index": {
            "maximum": 4,
            "minimum": 1,
            "type": "integer"
          },
          "response_id": {
            "maxLength": 96,
            "minLength": 1,
            "type": "string"
          },
          "response_kind": {
            "enum": [
              "attempt",
              "switch_strategy",
              "seek_resource",
              "negotiate",
              "confront",
              "evade",
              "investigate",
              "accept_cost",
              "revise_goal",
              "recruit_help",
              "move_location",
              "commit_to_action",
              "handoff"
            ],
            "type": "string"
          },
          "response_text": {
            "maxLength": 140,
            "minLength": 1,
            "type": "string"
          }
        },
        "required": [
          "response_id",
          "actor",
          "response_kind",
          "response_text",
          "paragraph_index",
          "evidence"
        ],
        "type": "object"
      },
      "maxItems": 8,
      "minItems": 0,
      "type": "array"
    }
  },
  "required": [
    "goals",
    "responses"
  ],
  "type": "object"
}
```
\end{Verbatim}
\par\medskip

\subsection{Mystery Remains Unresolved}
\label{app:mystery-unresolved}

\subsubsection{Measurement target}
The \emph{Mystery remains unresolved} diagnostic measures whether the First Chapter Text raises a concrete focal story question or information gap and keeps it active by returning to it as a follow-up question within a short paragraph window. A passing First Chapter Text must do more than create diffuse curiosity. It must expose a specific unknown target, such as who did something, why something happened, how a rule works, what danger exists, what someone wants, or what consequence is pending. The unresolved quality is operationalized by a nearby follow-up question.

\subsubsection{Extraction procedure}
The mystery score uses two extraction passes over paragraphs 1--12. The first pass identifies concrete questions or unresolved information gaps with a specific unknown target. The second pass identifies activities that keep such questions active by delaying, pursuing, complicating, transferring, pressuring, partially answering, or reopening them. The deterministic scorer then distinguishes major and focal questions and requires a nearby follow-up-question activity.

\subsubsection{Deterministic scoring rule}
The scorer first requires at least one concrete question or information gap. If none exists, the score fails because the chapter opening has not established a specific unknown. It then requires at least one activity that could keep a question active; without such activity, the question does not remain narratively operative. A question is considered major when it concerns a story problem, character goal, stakes, world rule, antagonistic force, identity or relationship issue, or causal secret. If no major question exists, the score fails because the uncertainty is only local or logistical. A question is considered focal only when it concerns a story problem, character goal, stakes, or world rule. If no focal question exists, the score fails because no central story question has been established.

For each focal question in paragraph \(q\), the scorer searches for a follow-up question activity in paragraph \(q\), \(q+1\), or \(q+2\), capped at paragraph 12. Same-paragraph matches are allowed. Thus, a question in paragraph 10 can match paragraph 10, 11, or 12; a question in paragraph 11 can match paragraph 11 or 12; and a question in paragraph 12 can match only paragraph 12. Focal questions are evaluated from earliest to latest paragraph, and activities are considered from earliest to latest paragraph. If no nearby follow-up question exists, the score fails because the focal uncertainty is not reopened. If a nearby follow-up question exists, the score passes because the focal mystery remains active rather than being closed or dropped.

\subsubsection{LLM extraction prompts}

\promptlistingtitle{System prompt: Concrete question-or-gap extraction.}
\begin{Verbatim}[fontsize=\scriptsize,breaklines=true,breakanywhere=true,frame=single,framerule=0.8pt,framesep=8pt,rulecolor=\color{promptframe}]
# Task
Extract concrete questions or unresolved information gaps raised in the First Chapter Text.

# Rules
- Use only explicit evidence from the supplied text or supplied artifact IDs.
- Do not infer from genre knowledge or likely later story events.
- Do not return a metric score, pass/fail, verdict, rating, or quality judgment.
- Keep evidence fields short: one clause or sentence, not a full paragraph.
- Prefer compact high-signal artifact lists over exhaustive inventories.
- Empty arrays are correct when the text has no evidence for a requested artifact.
- Return corrected JSON only if asked to repair a previous response.
- Use First Chapter Text paragraphs 1 through 12 only.
- Extract explicit questions and concrete missing information, identity unknowns, causal secrets, world-rule unknowns, goal uncertainties, and unclear threats.
- The gap must have a specific unknown target, such as who did something, why something happened, how a rule works, what danger exists, what someone wants, or what consequence is pending.
- Tag questions by narrative function: story problem, character goal, stakes, world rule, antagonistic force, identity or relationship, causal secret, or local logistics.
- Do not extract generic curiosity such as what happens next, atmosphere, mood, genre premise, or rhetorical phrasing unless the text gives a concrete unknown target.
- Return at most sixteen questions: keep concrete questions tied to problems, goals, stakes, world, identity, or causal secrets before small logistics.
- Use question IDs like q1, q2, q3.

# Output Rules
Return exactly one JSON object matching the Required JSON Schema below. Do not wrap it in Markdown.

# Required JSON Schema
```json
{
  "additionalProperties": false,
  "properties": {
    "questions": {
      "items": {
        "additionalProperties": false,
        "properties": {
          "evidence": {
            "maxLength": 110,
            "minLength": 1,
            "type": "string"
          },
          "function_tags": {
            "items": {
              "enum": [
                "story_problem",
                "character_goal",
                "stakes",
                "world_rule",
                "antagonistic_force",
                "identity_relationship",
                "causal_secret",
                "local_logistics"
              ],
              "type": "string"
            },
            "maxItems": 4,
            "minItems": 1,
            "type": "array"
          },
          "paragraph_index": {
            "maximum": 12,
            "minimum": 1,
            "type": "integer"
          },
          "question_id": {
            "maxLength": 96,
            "minLength": 1,
            "type": "string"
          },
          "question_kind": {
            "enum": [
              "explicit_question",
              "information_gap",
              "identity_unknown",
              "causal_secret",
              "world_rule_unknown",
              "goal_uncertainty",
              "threat_unknown"
            ],
            "type": "string"
          },
          "subject_text": {
            "maxLength": 120,
            "minLength": 1,
            "type": "string"
          },
          "unknown_target": {
            "maxLength": 160,
            "minLength": 1,
            "type": "string"
          }
        },
        "required": [
          "question_id",
          "question_kind",
          "function_tags",
          "subject_text",
          "unknown_target",
          "paragraph_index",
          "evidence"
        ],
        "type": "object"
      },
      "maxItems": 16,
      "minItems": 0,
      "type": "array"
    }
  },
  "required": [
    "questions"
  ],
  "type": "object"
}
```
\end{Verbatim}
\par\medskip

\promptlistingtitle{System prompt: Question-activity extraction.}
\begin{Verbatim}[fontsize=\scriptsize,breaklines=true,breakanywhere=true,frame=single,framerule=0.8pt,framesep=8pt,rulecolor=\color{promptframe}]
# Task
Extract activity cues that keep a concrete question or information gap alive.

# Rules
- Use only explicit evidence from the supplied text or supplied artifact IDs.
- Do not infer from genre knowledge or likely later story events.
- Do not return a metric score, pass/fail, verdict, rating, or quality judgment.
- Keep evidence fields short: one clause or sentence, not a full paragraph.
- Prefer compact high-signal artifact lists over exhaustive inventories.
- Empty arrays are correct when the text has no evidence for a requested artifact.
- Return corrected JSON only if asked to repair a previous response.
- Use First Chapter Text paragraphs 1 through 12 only.
- Extract exchanges, blocked answers, partial answers, deferrals, new constraints, follow-up questions, investigation attempts, answers that raise another concrete gap, and end-of-text unresolved markers.
- A cue can keep a question active by delaying the answer, pursuing it, complicating it, transferring it to another person, adding pressure, or answering only part of it.
- Do not decide whether any cue is enough for the metric.
- Do not extract passive description unless it delays, pressures, partially answers, repeats, investigates, hands off, or leaves open a concrete question.
- Return at most twenty cues: prefer cues connected to problems, goals, stakes, world rules, identity, threats, or causal secrets.
- Use activity IDs like a1, a2, a3.

# Output Rules
Return exactly one JSON object matching the Required JSON Schema below. Do not wrap it in Markdown.

# Required JSON Schema
```json
{
  "additionalProperties": false,
  "properties": {
    "activities": {
      "items": {
        "additionalProperties": false,
        "properties": {
          "activity_id": {
            "maxLength": 96,
            "minLength": 1,
            "type": "string"
          },
          "activity_kind": {
            "enum": [
              "exchange",
              "refusal_or_blocked_answer",
              "partial_answer",
              "deferral",
              "new_constraint",
              "followup_question",
              "investigation_attempt",
              "answer_raises_new_gap",
              "unresolved_end_marker"
            ],
            "type": "string"
          },
          "activity_summary": {
            "maxLength": 150,
            "minLength": 1,
            "type": "string"
          },
          "evidence": {
            "maxLength": 110,
            "minLength": 1,
            "type": "string"
          },
          "paragraph_index": {
            "maximum": 12,
            "minimum": 1,
            "type": "integer"
          },
          "subject_text": {
            "maxLength": 120,
            "minLength": 1,
            "type": "string"
          }
        },
        "required": [
          "activity_id",
          "activity_kind",
          "subject_text",
          "activity_summary",
          "paragraph_index",
          "evidence"
        ],
        "type": "object"
      },
      "maxItems": 20,
      "minItems": 0,
      "type": "array"
    }
  },
  "required": [
    "activities"
  ],
  "type": "object"
}
```
\end{Verbatim}
\par\medskip

\subsection{Initial Situation Is Clear}
\label{app:initial-situation-clear}

\subsubsection{Measurement target}
The \emph{Initial situation is clear} diagnostic checks whether the beginning of the First Chapter Text gives the reader enough immediate orientation to understand who is involved, what local frame they are in, what is happening now, and why the current situation matters. The score does not judge prose quality directly.

\subsubsection{Extraction procedure}
The initial-situation score uses one extraction pass over paragraphs 1--6. The pass identifies the smallest useful set of opening-orientation anchors: who is involved, what local frame the scene occupies, what is happening now, and why the current situation matters. It is deliberately not an inventory task; the prompt asks for one anchor per component, with a second only when it adds essential orientation evidence. In scoring, time or sequence alone does not satisfy the strong-context requirement; the context must supply a stronger local frame such as place, social situation, role, relationship, or storyworld rule.

\subsubsection{Deterministic scoring rule}
The score passes when all four orientation components exist and arrive within their allowed windows: at least one participant anchor, at least one strong context anchor, at least one current-activity anchor, and at least one relevance anchor. The selected participant, selected strong context, and selected current activity must appear in paragraphs 1--4. The selected relevance anchor must appear in paragraphs 1--6. The decision rule is sequential: missing participant evidence fails the score; if participant evidence exists, missing strong context fails it; if strong context exists, missing current activity fails it; if current activity exists, missing relevance fails it. After all four components exist, late selected anchors fail the score in the same order: participant after paragraph 4, strong context after paragraph 4, current activity after paragraph 4, or relevance after paragraph 6. If none of these failures applies, the score passes.

\subsubsection{LLM extraction prompt}

\promptlistingtitle{System prompt: Opening-orientation anchor extraction.}
\begin{Verbatim}[fontsize=\scriptsize,breaklines=true,breakanywhere=true,frame=single,framerule=0.8pt,framesep=8pt,rulecolor=\color{promptframe}]
# Task
Extract the smallest useful set of opening-orientation anchors from the First Chapter Text.

# Rules
- Use only explicit evidence from the supplied text or supplied artifact IDs.
- Do not infer from genre knowledge or likely later story events.
- Do not return a metric score, pass/fail, verdict, rating, or quality judgment.
- Keep evidence fields short: one clause or sentence, not a full paragraph.
- Prefer compact high-signal artifact lists over exhaustive inventories.
- Empty arrays are correct when the text has no evidence for a requested artifact.
- Return corrected JSON only if asked to repair a previous response.
- Use First Chapter Text paragraphs 1 through 6 only.
- This is not an inventory. Prefer one anchor for each array; include a second only when it adds essential orientation evidence.
- Participant anchors identify who the opening currently involves: named people, role labels, groups, first-person narrators, or stated relationships.
- Context anchors identify the local frame needed to understand the current scene: where, when, social setting, role setting, relationship setting, or storyworld rule.
- Current activity anchors identify what is happening in the opening now: arrival, scene action, exchange, observation, routine work, preparation, waiting, confinement, decision, active problem state, or narrated present state.
- For current activity anchors, prefer the earliest concrete present-scene action, routine, observation, decision, exchange, preparation, or current state that lets a reader know what is happening now.
- Do not skip an early current-scene activity merely because a later paragraph contains a stronger dramatic incident.
- Relevance anchors identify why this current situation matters now: assignment, role, purpose, desire, obligation, task, risk, abnormality, change, relationship stake, social stake, resource stake, or pending consequence.
- For relevance anchors, prefer the earliest explicit reason the current situation matters, but the reason can appear later than the basic who, context, and current activity anchors.
- Do not extract generic mood, decorative description, or backstory unless it directly anchors who, where, what is currently happening, or why this situation matters now.
- Do not decide whether the opening is clear.
- Use IDs like p1, c1, a1, r1.

# Output Rules
Return exactly one JSON object matching the Required JSON Schema below. Do not wrap it in Markdown.

# Required JSON Schema
```json
{
  "additionalProperties": false,
  "properties": {
    "context_anchors": {
      "items": {
        "additionalProperties": false,
        "properties": {
          "anchor_id": {
            "maxLength": 96,
            "minLength": 1,
            "type": "string"
          },
          "anchor_kind": {
            "enum": [
              "physical_location",
              "time_or_sequence",
              "social_context",
              "role_context",
              "relationship_context",
              "world_rule"
            ],
            "type": "string"
          },
          "evidence": {
            "maxLength": 110,
            "minLength": 1,
            "type": "string"
          },
          "label": {
            "maxLength": 140,
            "minLength": 1,
            "type": "string"
          },
          "paragraph_index": {
            "maximum": 6,
            "minimum": 1,
            "type": "integer"
          }
        },
        "required": [
          "anchor_id",
          "anchor_kind",
          "label",
          "paragraph_index",
          "evidence"
        ],
        "type": "object"
      },
      "maxItems": 2,
      "minItems": 0,
      "type": "array"
    },
    "current_activity_anchors": {
      "items": {
        "additionalProperties": false,
        "properties": {
          "actor_text": {
            "maxLength": 120,
            "minLength": 1,
            "type": "string"
          },
          "anchor_id": {
            "maxLength": 96,
            "minLength": 1,
            "type": "string"
          },
          "anchor_kind": {
            "enum": [
              "arrival_or_departure",
              "scene_action",
              "exchange",
              "observation",
              "routine_work",
              "preparation",
              "waiting_or_containment",
              "decision",
              "active_problem_state",
              "narrated_present_state"
            ],
            "type": "string"
          },
          "evidence": {
            "maxLength": 110,
            "minLength": 1,
            "type": "string"
          },
          "label": {
            "maxLength": 150,
            "minLength": 1,
            "type": "string"
          },
          "paragraph_index": {
            "maximum": 6,
            "minimum": 1,
            "type": "integer"
          }
        },
        "required": [
          "anchor_id",
          "anchor_kind",
          "actor_text",
          "label",
          "paragraph_index",
          "evidence"
        ],
        "type": "object"
      },
      "maxItems": 2,
      "minItems": 0,
      "type": "array"
    },
    "participant_anchors": {
      "items": {
        "additionalProperties": false,
        "properties": {
          "anchor_id": {
            "maxLength": 96,
            "minLength": 1,
            "type": "string"
          },
          "anchor_kind": {
            "enum": [
              "named_person",
              "role",
              "group",
              "first_person_narrator",
              "relationship"
            ],
            "type": "string"
          },
          "evidence": {
            "maxLength": 100,
            "minLength": 1,
            "type": "string"
          },
          "label": {
            "maxLength": 120,
            "minLength": 1,
            "type": "string"
          },
          "paragraph_index": {
            "maximum": 6,
            "minimum": 1,
            "type": "integer"
          }
        },
        "required": [
          "anchor_id",
          "anchor_kind",
          "label",
          "paragraph_index",
          "evidence"
        ],
        "type": "object"
      },
      "maxItems": 2,
      "minItems": 0,
      "type": "array"
    },
    "relevance_anchors": {
      "items": {
        "additionalProperties": false,
        "properties": {
          "affected_text": {
            "maxLength": 120,
            "minLength": 1,
            "type": "string"
          },
          "anchor_id": {
            "maxLength": 96,
            "minLength": 1,
            "type": "string"
          },
          "anchor_kind": {
            "enum": [
              "assignment_or_role",
              "purpose_or_desire",
              "obligation_or_task",
              "risk_or_threat",
              "abnormality_or_change",
              "relationship_stake",
              "social_or_status_stake",
              "resource_or_access_stake",
              "pending_consequence"
            ],
            "type": "string"
          },
          "evidence": {
            "maxLength": 110,
            "minLength": 1,
            "type": "string"
          },
          "label": {
            "maxLength": 150,
            "minLength": 1,
            "type": "string"
          },
          "paragraph_index": {
            "maximum": 6,
            "minimum": 1,
            "type": "integer"
          }
        },
        "required": [
          "anchor_id",
          "anchor_kind",
          "affected_text",
          "label",
          "paragraph_index",
          "evidence"
        ],
        "type": "object"
      },
      "maxItems": 2,
      "minItems": 0,
      "type": "array"
    }
  },
  "required": [
    "participant_anchors",
    "context_anchors",
    "current_activity_anchors",
    "relevance_anchors"
  ],
  "type": "object"
}
```
\end{Verbatim}
\par\medskip

\subsection{Viewpoint Control Holds}
\label{app:viewpoint-control-holds}

\subsubsection{Measurement target}
The \emph{Viewpoint control holds} diagnostic checks whether the opening passage of the First Chapter Text maintains a legible controlling perspective. It tracks who perceives, who knows, whose interiority is entered, and whether viewpoint shifts are locally marked. A First Chapter Text can pass with one stable viewpoint regime, with externally observable narration, or with multiple viewpoint spans, provided that changes in focal holder, access scope, and narrator knowledge are either locally signaled or harmless under the deterministic checks.

\subsubsection{Extraction procedure}
The viewpoint score uses two extraction passes over paragraphs 1--12. The first pass divides the text into viewpoint spans and records local signals that may mark a shift in focal holder, access scope, or narrator knowledge. The second pass records concrete attribution events that could make viewpoint control difficult to follow, such as holderless interior access, unmarked movement between minds, or unavailable focal knowledge. The deterministic scorer applies coverage, ambiguity, multi-mind, attribution-event, and boundary-change filters to the extracted artifacts.

\subsubsection{Deterministic scoring rule}
The scoring layer applies ordered gates. First, if no viewpoint spans are extracted, the score fails because the chapter has no recoverable perspective regime. Second, it computes the set of paragraph indices covered by all span ranges. Required coverage is the smaller of six paragraphs and the available First Chapter Text opening length. If the covered set is smaller, the score fails because perspective evidence does not cover enough of the opening. Third, it sums the lengths of spans whose narration person, focal anchor, or access scope is marked as unclear or mixed. If this total divided by the covered paragraph count is greater than 0.3, the score fails because too much of the extracted viewpoint regime is ambiguous.

Fourth, each span that enters more than one character's interiority is checked. Mental-access targets are normalized, and empty, narrator-level, and unclear targets are removed. If more than one unique target remains, the span fails unless it is anchored as a group perspective or a scope marker is present at the span start boundary. Fifth, attribution events are filtered for actionability. If at least two actionable attribution problems remain, the score fails because perspective control has become locally unstable. Sixth, spans are sorted by start paragraph and adjacent boundaries are inspected. Any hard change in focal anchor, access scope, or knowledge scope fails unless it is locally marked. If all gates pass, the score passes. There is no minimum dominant-ratio threshold.

\subsubsection{Boundary and actionability rules}
All text comparisons for anchors, holders, and mental targets use lowercase text with leading and trailing whitespace removed and internal whitespace collapsed. A signal is near a span boundary when its paragraph index is either the current span's start paragraph or the preceding paragraph. Hard boundary-change checks are protected by near signals such as a scene break, explicit point-of-view label, named focal reanchor, retrospective marker, omniscient-scope marker, or document boundary. When the focal anchor changes, a paragraph-internal reanchor also protects the boundary if the current start paragraph's first 220 normalized characters contain the full normalized anchor text, or contain the anchor's final word when that final word is at least three characters long. Only named-character and group anchors can use this implicit reanchor rule.

If adjacent spans both have hard access scopes and their normalized anchors differ, the boundary fails unless protected. If access scope changes and at least one side has a hard access scope, the boundary fails unless the anchors are the same, one side is external-only narration, or the boundary is protected. If knowledge scope moves from limited or external knowledge to a wide or retrospective narrator, and the current access scope is not external-only, the boundary fails unless protected.

For attribution events, missing holder text means that the normalized holder is empty or marked as none, unclear, or unknown. Holderless interior access and holderless belief or inference are actionable only when the first holder is missing and the evidence does not contain a first-person anchor word such as I, me, my, mine, we, us, our, or ours. Unreanchored changes in interior holder and same-beat access to multiple minds are actionable only when both holders are present and their normalized texts differ. Impossible focal knowledge is always actionable.

\subsubsection{LLM extraction prompts}

\promptlistingtitle{System prompt: Viewpoint-span and shift-signal extraction.}
\begin{Verbatim}[fontsize=\scriptsize,breaklines=true,breakanywhere=true,frame=single,framerule=0.8pt,framesep=8pt,rulecolor=\color{promptframe}]
# Task
Extract viewpoint spans and local viewpoint-shift signals from the First Chapter Text.

# Rules
- Use only explicit evidence from the supplied text or supplied artifact IDs.
- Do not infer from genre knowledge or likely later story events.
- Do not return a metric score, pass/fail, verdict, rating, or quality judgment.
- Keep evidence fields short: one clause or sentence, not a full paragraph.
- Prefer compact high-signal artifact lists over exhaustive inventories.
- Empty arrays are correct when the text has no evidence for a requested artifact.
- Return corrected JSON only if asked to repair a previous response.
- Use First Chapter Text paragraphs 1 through 12 only.
- A viewpoint span is a contiguous paragraph range with the same narration person, narrator position, focal anchor, interior-access scope, and knowledge scope.
- Prefer broad contiguous spans over sentence-level fragments. Return at most eight spans.
- Avoid one oversized span when different paragraph groups need different perspective, access, or knowledge labels to remain accurate.
- Use focal_anchor_text for the character, narrator self, group, or narrator frame that controls perception or knowledge in the span. Use narrator when there is no character anchor.
- Use external_only when the span reports only observable action, dialogue, or setting without interior access.
- Use one_character_interior when thoughts, feelings, perceptions, or limited knowledge belong to one holder.
- Use multiple_character_interior only when the text directly enters more than one mind in the same local span.
- Do not split a span only because the text moves from observable action or setting into the same character's thoughts, feelings, perceptions, or limited knowledge.
- Use external_observable for externally observable action, dialogue, or setting.
- Use focal_character_limited for thoughts, feelings, perceptions, or knowledge limited to one character.
- Use narrator_wide for narrator-supplied context, background, role explanation, or world rules that go beyond a single character's immediate access.
- Use unclear knowledge scope only when the text makes the controlling perspective genuinely hard to attribute.
- Shift signals are textual reanchors at or near a span boundary, such as a scene break, named focal reanchor, dialogue turn, retrospective cue, omniscient cue, or document boundary.
- Use named_focal_reanchor when a paragraph begins by naming or clearly reintroducing the character whose perspective guides the following span.
- Use no_local_signal only for a boundary where the text changes focal holder or access scope without such a cue.
- Do not decide whether viewpoint control holds.
- Use IDs like s1 and g1.

# Output Rules
Return exactly one JSON object matching the Required JSON Schema below. Do not wrap it in Markdown.

# Required JSON Schema
```json
{
  "additionalProperties": false,
  "properties": {
    "shift_signals": {
      "items": {
        "additionalProperties": false,
        "properties": {
          "evidence": {
            "maxLength": 100,
            "minLength": 1,
            "type": "string"
          },
          "paragraph_index": {
            "maximum": 12,
            "minimum": 1,
            "type": "integer"
          },
          "signal_id": {
            "maxLength": 96,
            "minLength": 1,
            "type": "string"
          },
          "signal_kind": {
            "enum": [
              "scene_break",
              "explicit_pov_label",
              "named_focal_reanchor",
              "dialogue_turn",
              "retrospective_marker",
              "omniscient_scope_cue",
              "document_boundary",
              "no_local_signal"
            ],
            "type": "string"
          },
          "signaled_target_text": {
            "maxLength": 120,
            "minLength": 1,
            "type": "string"
          }
        },
        "required": [
          "signal_id",
          "paragraph_index",
          "signal_kind",
          "signaled_target_text",
          "evidence"
        ],
        "type": "object"
      },
      "maxItems": 8,
      "minItems": 0,
      "type": "array"
    },
    "viewpoint_spans": {
      "items": {
        "additionalProperties": false,
        "properties": {
          "access_scope": {
            "enum": [
              "external_only",
              "one_character_interior",
              "multiple_character_interior",
              "omniscient_scope",
              "retrospective_character_scope",
              "unclear"
            ],
            "type": "string"
          },
          "end_paragraph_index": {
            "maximum": 12,
            "minimum": 1,
            "type": "integer"
          },
          "evidence": {
            "maxLength": 110,
            "minLength": 1,
            "type": "string"
          },
          "focal_anchor_kind": {
            "enum": [
              "named_character",
              "narrator_self",
              "group",
              "no_character_anchor",
              "unclear"
            ],
            "type": "string"
          },
          "focal_anchor_text": {
            "maxLength": 120,
            "minLength": 1,
            "type": "string"
          },
          "knowledge_scope": {
            "enum": [
              "focal_character_limited",
              "narrator_wide",
              "retrospective_narrator",
              "external_observable",
              "unclear"
            ],
            "type": "string"
          },
          "mental_access_targets": {
            "items": {
              "maxLength": 80,
              "minLength": 1,
              "type": "string"
            },
            "maxItems": 4,
            "minItems": 0,
            "type": "array"
          },
          "narration_person": {
            "enum": [
              "first_person",
              "third_person",
              "second_person",
              "mixed_or_unclear"
            ],
            "type": "string"
          },
          "narrator_position": {
            "enum": [
              "character_narrator",
              "external_narrator",
              "documentary_frame",
              "unclear"
            ],
            "type": "string"
          },
          "span_id": {
            "maxLength": 96,
            "minLength": 1,
            "type": "string"
          },
          "start_paragraph_index": {
            "maximum": 12,
            "minimum": 1,
            "type": "integer"
          }
        },
        "required": [
          "span_id",
          "start_paragraph_index",
          "end_paragraph_index",
          "narration_person",
          "narrator_position",
          "focal_anchor_text",
          "focal_anchor_kind",
          "access_scope",
          "knowledge_scope",
          "mental_access_targets",
          "evidence"
        ],
        "type": "object"
      },
      "maxItems": 8,
      "minItems": 0,
      "type": "array"
    }
  },
  "required": [
    "viewpoint_spans",
    "shift_signals"
  ],
  "type": "object"
}
```
\end{Verbatim}
\par\medskip

\promptlistingtitle{System prompt: Viewpoint-attribution event extraction.}
\begin{Verbatim}[fontsize=\scriptsize,breaklines=true,breakanywhere=true,frame=single,framerule=0.8pt,framesep=8pt,rulecolor=\color{promptframe}]
# Task
Extract concrete local viewpoint-attribution events from the First Chapter Text.

# Rules
- Use only explicit evidence from the supplied text or supplied artifact IDs.
- Do not infer from genre knowledge or likely later story events.
- Do not return a metric score, pass/fail, verdict, rating, or quality judgment.
- Keep evidence fields short: one clause or sentence, not a full paragraph.
- Prefer compact high-signal artifact lists over exhaustive inventories.
- Empty arrays are correct when the text has no evidence for a requested artifact.
- Return corrected JSON only if asked to repair a previous response.
- Use First Chapter Text paragraphs 1 through 12 only.
- This is not a verdict about the First Chapter Text. Extract only explicit local text patterns.
- Use unanchored_interior_access when a thought, feeling, perception, or private judgment appears without a nearby textual holder.
- Do not use unanchored_interior_access when the holder can be named. If this event kind is used, first_holder_text must be none.
- Use unreanchored_interior_holder_change when one local beat moves from one holder's interior access to a different holder's interior access without a named reanchor, scene break, document boundary, or explicit viewpoint marker.
- Use same_local_beat_multiple_minds when the same sentence or tightly connected beat directly enters more than one mind.
- Use impossible_focal_knowledge when a limited focal holder is presented as knowing something the local text makes unavailable to that holder.
- Use belief_or_inference_without_holder when a belief, guess, suspicion, or evaluative judgment is presented without enough text to tell whether it belongs to a character or the narrator.
- Do not extract first-person statements with I, me, my, we, us, or our as holderless; those are anchored to the first-person narrator.
- Do not extract ordinary narrator-supplied background, role explanation, setting description, or world rules.
- Do not extract clearly collective group reactions such as a group being afraid, excited, silent, or worried.
- Do not extract dialogue turns with named speakers unless the prose directly enters different minds without a reanchor.
- Do not extract a paragraph that begins by naming or clearly reintroducing the new holder.
- Use second_holder_text as none when the event has only one holder.
- Return an empty attribution_events array when the supplied text has no explicit local event.
- Use IDs like e1.

# Output Rules
Return exactly one JSON object matching the Required JSON Schema below. Do not wrap it in Markdown.

# Required JSON Schema
```json
{
  "additionalProperties": false,
  "properties": {
    "attribution_events": {
      "items": {
        "additionalProperties": false,
        "properties": {
          "event_id": {
            "maxLength": 96,
            "minLength": 1,
            "type": "string"
          },
          "event_kind": {
            "enum": [
              "unanchored_interior_access",
              "unreanchored_interior_holder_change",
              "same_local_beat_multiple_minds",
              "impossible_focal_knowledge",
              "belief_or_inference_without_holder"
            ],
            "type": "string"
          },
          "evidence": {
            "maxLength": 130,
            "minLength": 1,
            "type": "string"
          },
          "first_holder_text": {
            "maxLength": 120,
            "minLength": 1,
            "type": "string"
          },
          "paragraph_index": {
            "maximum": 12,
            "minimum": 1,
            "type": "integer"
          },
          "second_holder_text": {
            "maxLength": 120,
            "minLength": 1,
            "type": "string"
          }
        },
        "required": [
          "event_id",
          "paragraph_index",
          "event_kind",
          "first_holder_text",
          "second_holder_text",
          "evidence"
        ],
        "type": "object"
      },
      "maxItems": 5,
      "minItems": 0,
      "type": "array"
    }
  },
  "required": [
    "attribution_events"
  ],
  "type": "object"
}
```
\end{Verbatim}
\par\medskip

\subsection{Exposition Is Motivated by Local Need}
\label{app:exposition-local-need}

\subsubsection{Measurement target}
The \emph{Exposition is motivated by local need} diagnostic checks whether exposition in the First Chapter Text is delivered because a character, narrator, or current scene needs the information now, rather than because the First Chapter Text pauses to explain background. The score rewards explanatory information prompted by an immediate local need, such as a question, request, blocked action, decision, warning, rule or permission issue, missing identity or location, or current confusion. It does not reward general worldbuilding, broad reader curiosity, decorative lore, or background that merely shares a topic with the current scene.

\subsubsection{Extraction procedure}
The exposition score uses three extraction passes over paragraphs 1--12. The first pass identifies local information needs that belong to a character, narrator, or current scene. The second pass identifies compact exposition units: background information delivered as an answer, instruction, explanation, message, artifact, or recall. The comparison pass links needs to exposition units when the text makes the information locally useful rather than merely topically related. The scoring rule then applies a stricter strong-link filter.

\subsubsection{Deterministic scoring rule}
The strong-link filter requires every condition below. The link must either answer a question or identify a needed person, place, or object. Any valid local-need category is acceptable. The exposition must provide a rule or procedure, an identity or role fact, or a location or resource fact. It must be delivered through dialogue, either as an answer or as an instruction. Its placement must be in the same paragraph as the need, immediately after it, or otherwise near after it. Finally, the paragraph gap, computed as exposition paragraph minus need paragraph, must be 0 or 1. This paragraph-gap rule is stricter than the placement label: even a near-after link must still occur in the same paragraph as the need or exactly one paragraph after it.

The score fails if no local need is extracted; this check has priority over other empty-list checks. It fails if at least one local need exists but no exposition unit exists. It fails if both local needs and exposition units exist but no link connects them. It fails if links exist but no resolved link satisfies every strong-link filter. It passes when at least one resolved link satisfies all filters.

\subsubsection{LLM extraction prompts}

\promptlistingtitle{System prompt: Local information-need extraction.}
\begin{Verbatim}[fontsize=\scriptsize,breaklines=true,breakanywhere=true,frame=single,framerule=0.8pt,framesep=8pt,rulecolor=\color{promptframe}]
# Task
Extract concrete local information needs from the First Chapter Text.

# Rules
- Use only explicit evidence from the supplied text or supplied artifact IDs.
- Do not infer from genre knowledge or likely later story events.
- Do not return a metric score, pass/fail, verdict, rating, or quality judgment.
- Keep evidence fields short: one clause or sentence, not a full paragraph.
- Prefer compact high-signal artifact lists over exhaustive inventories.
- Empty arrays are correct when the text has no evidence for a requested artifact.
- Return corrected JSON only if asked to repair a previous response.
- Use First Chapter Text paragraphs 1 through 12 only.
- A local need must belong to a character, narrator, or current scene: a question, request, blocked action, decision, warning, rule or permission need, missing identity or location, or current confusion.
- Do not extract broad reader curiosity, genre premise, mood, or a general need to understand the world.
- Do not extract background information itself; extract the scene-owned need that makes information useful now.
- Return at most six local needs total.
- Do not decide whether exposition is motivated by local need.
- Use need IDs like n1, n2, n3.

# Output Rules
Return exactly one JSON object matching the Required JSON Schema below. Do not wrap it in Markdown.

# Required JSON Schema
```json
{
  "additionalProperties": false,
  "properties": {
    "local_needs": {
      "items": {
        "additionalProperties": false,
        "properties": {
          "evidence": {
            "maxLength": 90,
            "minLength": 1,
            "type": "string"
          },
          "need_id": {
            "maxLength": 96,
            "minLength": 1,
            "type": "string"
          },
          "need_kind": {
            "enum": [
              "explicit_question",
              "request_for_information",
              "blocked_action",
              "decision_need",
              "danger_warning_need",
              "rule_permission_need",
              "missing_identity_or_location",
              "current_relationship_confusion",
              "current_emotional_confusion"
            ],
            "type": "string"
          },
          "need_text": {
            "maxLength": 130,
            "minLength": 1,
            "type": "string"
          },
          "owner_text": {
            "maxLength": 90,
            "minLength": 1,
            "type": "string"
          },
          "paragraph_index": {
            "maximum": 12,
            "minimum": 1,
            "type": "integer"
          }
        },
        "required": [
          "need_id",
          "need_kind",
          "owner_text",
          "need_text",
          "paragraph_index",
          "evidence"
        ],
        "type": "object"
      },
      "maxItems": 6,
      "minItems": 0,
      "type": "array"
    }
  },
  "required": [
    "local_needs"
  ],
  "type": "object"
}
```
\end{Verbatim}
\par\medskip

\promptlistingtitle{System prompt: Exposition-unit extraction.}
\begin{Verbatim}[fontsize=\scriptsize,breaklines=true,breakanywhere=true,frame=single,framerule=0.8pt,framesep=8pt,rulecolor=\color{promptframe}]
# Task
Extract exposition units from the First Chapter Text.

# Rules
- Use only explicit evidence from the supplied text or supplied artifact IDs.
- Do not infer from genre knowledge or likely later story events.
- Do not return a metric score, pass/fail, verdict, rating, or quality judgment.
- Keep evidence fields short: one clause or sentence, not a full paragraph.
- Prefer compact high-signal artifact lists over exhaustive inventories.
- Empty arrays are correct when the text has no evidence for a requested artifact.
- Return corrected JSON only if asked to repair a previous response.
- Use First Chapter Text paragraphs 1 through 12 only.
- An exposition unit is concrete background information delivered as an answer, instruction, explanation, message, artifact, or recall about prior events, rules, identity, roles, locations, resources, constraints, relationships, or customs.
- Do not extract present-moment action unless it explicitly supplies background information.
- Do not extract encyclopedia-like lists, atmospheric lore, or facts that only decorate the setting.
- Prefer compact units that could plausibly answer or enable one local passage.
- Return at most six exposition units total.
- Do not decide whether exposition is motivated by local need.
- Use exposition IDs like e1, e2, e3.

# Output Rules
Return exactly one JSON object matching the Required JSON Schema below. Do not wrap it in Markdown.

# Required JSON Schema
```json
{
  "additionalProperties": false,
  "properties": {
    "exposition_units": {
      "items": {
        "additionalProperties": false,
        "properties": {
          "delivery_mode": {
            "enum": [
              "dialogue_answer",
              "dialogue_instruction",
              "narrator_explanation",
              "document_or_message",
              "observed_artifact",
              "interior_recall"
            ],
            "type": "string"
          },
          "evidence": {
            "maxLength": 100,
            "minLength": 1,
            "type": "string"
          },
          "exposition_id": {
            "maxLength": 96,
            "minLength": 1,
            "type": "string"
          },
          "exposition_kind": {
            "enum": [
              "answer_or_reveal",
              "rule_or_procedure",
              "backstory_fact",
              "world_rule",
              "identity_or_role_fact",
              "location_or_resource_fact",
              "constraint_explanation",
              "relationship_explanation",
              "social_custom"
            ],
            "type": "string"
          },
          "information_text": {
            "maxLength": 150,
            "minLength": 1,
            "type": "string"
          },
          "paragraph_index": {
            "maximum": 12,
            "minimum": 1,
            "type": "integer"
          }
        },
        "required": [
          "exposition_id",
          "delivery_mode",
          "exposition_kind",
          "information_text",
          "paragraph_index",
          "evidence"
        ],
        "type": "object"
      },
      "maxItems": 6,
      "minItems": 0,
      "type": "array"
    }
  },
  "required": [
    "exposition_units"
  ],
  "type": "object"
}
```
\end{Verbatim}
\par\medskip

\promptlistingtitle{System prompt: Need-to-exposition link comparison.}
\begin{Verbatim}[fontsize=\scriptsize,breaklines=true,breakanywhere=true,frame=single,framerule=0.8pt,framesep=8pt,rulecolor=\color{promptframe}]
# Task
Compare local needs to exposition units and label their text-supported relation.

# Rules
- Use only explicit evidence from the supplied text or supplied artifact IDs.
- Do not infer from genre knowledge or likely later story events.
- Do not return a metric score, pass/fail, verdict, rating, or quality judgment.
- Keep evidence fields short: one clause or sentence, not a full paragraph.
- Prefer compact high-signal artifact lists over exhaustive inventories.
- Empty arrays are correct when the text has no evidence for a requested artifact.
- Return corrected JSON only if asked to repair a previous response.
- Use only the supplied need_id and exposition_id values.
- Create links only when a local need and an exposition unit are textually connected.
- Label placement from the exposition unit's position relative to the local need.
- Use same_topic_only when both artifacts concern a similar topic but the exposition does not answer, enable, identify, warn, constrain, or supply a rule for the local need.
- Use background_only when the exposition is broad background not made necessary by the supplied local need.
- Use unrelated when the artifacts are not connected.
- Do not invent IDs or add new First Chapter Text evidence.
- Do not decide whether exposition is motivated by local need.
- Return at most six links.
- Use link IDs like l1, l2, l3.

# Output Rules
Return exactly one JSON object matching the Required JSON Schema below. Do not wrap it in Markdown.

# Required JSON Schema
```json
{
  "additionalProperties": false,
  "properties": {
    "exposition_need_links": {
      "items": {
        "additionalProperties": false,
        "properties": {
          "basis_evidence": {
            "maxLength": 90,
            "minLength": 1,
            "type": "string"
          },
          "exposition_id": {
            "maxLength": 96,
            "minLength": 1,
            "type": "string"
          },
          "link_id": {
            "maxLength": 96,
            "minLength": 1,
            "type": "string"
          },
          "link_relation": {
            "enum": [
              "answers_question",
              "enables_immediate_action",
              "clarifies_current_risk",
              "explains_current_constraint",
              "identifies_needed_person_place_or_object",
              "supplies_rule_for_current_choice",
              "clarifies_current_relationship",
              "explains_current_emotion",
              "same_topic_only",
              "background_only",
              "unrelated",
              "unclear"
            ],
            "type": "string"
          },
          "need_id": {
            "maxLength": 96,
            "minLength": 1,
            "type": "string"
          },
          "placement": {
            "enum": [
              "before_local_need",
              "same_paragraph",
              "immediately_after",
              "near_after",
              "late_after",
              "unclear"
            ],
            "type": "string"
          }
        },
        "required": [
          "link_id",
          "need_id",
          "exposition_id",
          "placement",
          "link_relation",
          "basis_evidence"
        ],
        "type": "object"
      },
      "maxItems": 6,
      "minItems": 0,
      "type": "array"
    }
  },
  "required": [
    "exposition_need_links"
  ],
  "type": "object"
}
```
\end{Verbatim}
\par\medskip

\clearpage
\section{Additional Attention Interpretability Analysis}%
\label{sec:appendix-attention}

\subsection{Repetition-Failure Routing Dynamics}%
\label{subsec:appendix-repetition-failure}

Some failed generations enter a repetitive state in which the chapter begins to repeat the same words, phrases, or sentence structures. Figure~\ref{fig:appendix-failure-routing} shows the corresponding attention dynamics for one such example. The prior-component attention curve begins to oscillate once the repetition loop emerges. Because this curve measures attention allocated to components outside the current chapter, the complementary within-chapter attention exhibits the same periodic behavior.

\begin{center}
\begin{minipage}[t]{0.495\linewidth}
\centering
\includegraphics[width=\linewidth,height=0.23\textheight,keepaspectratio]{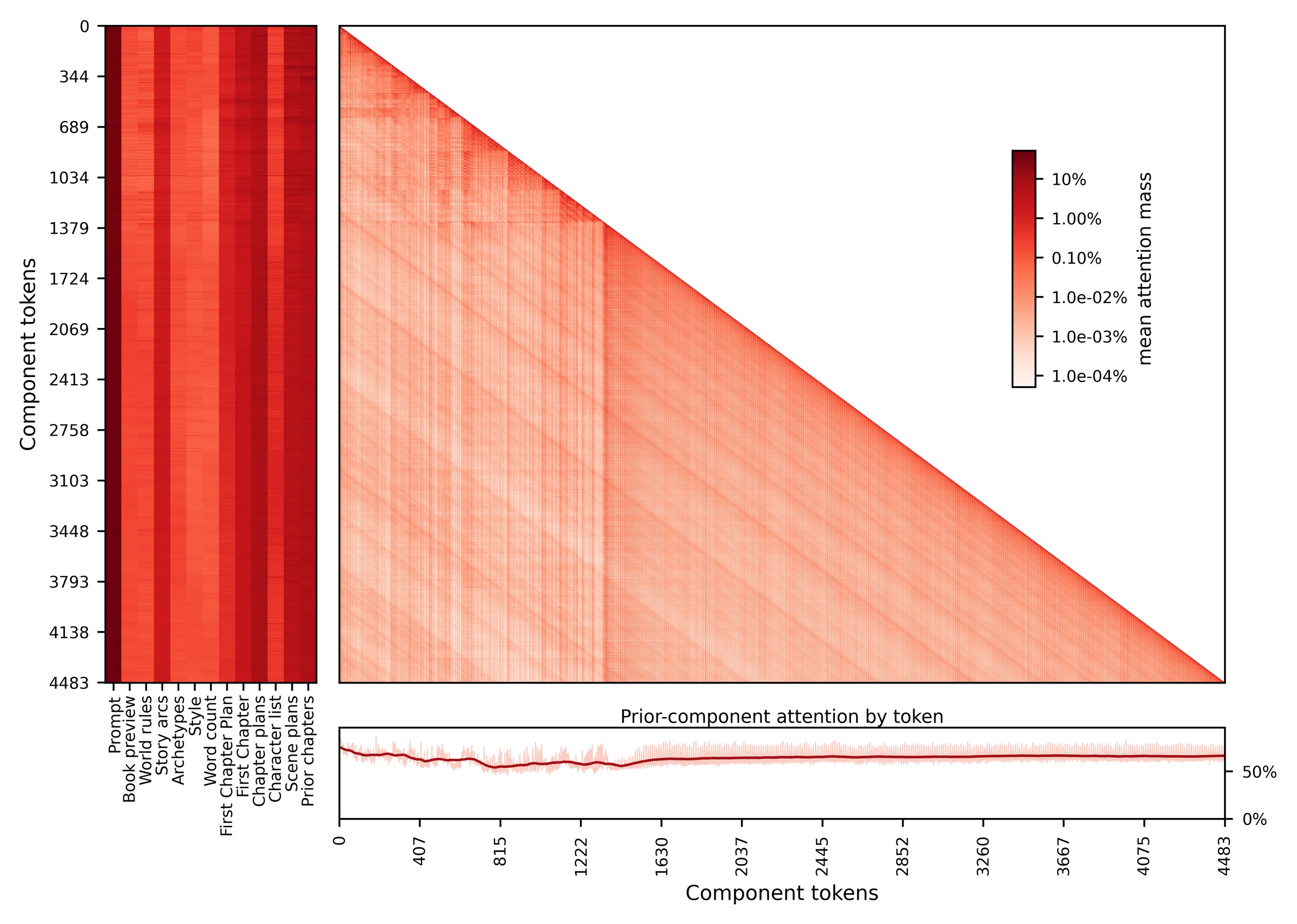}
\end{minipage}\hfill
\begin{minipage}[t]{0.495\linewidth}
\centering
\includegraphics[width=\linewidth,height=0.23\textheight,keepaspectratio]{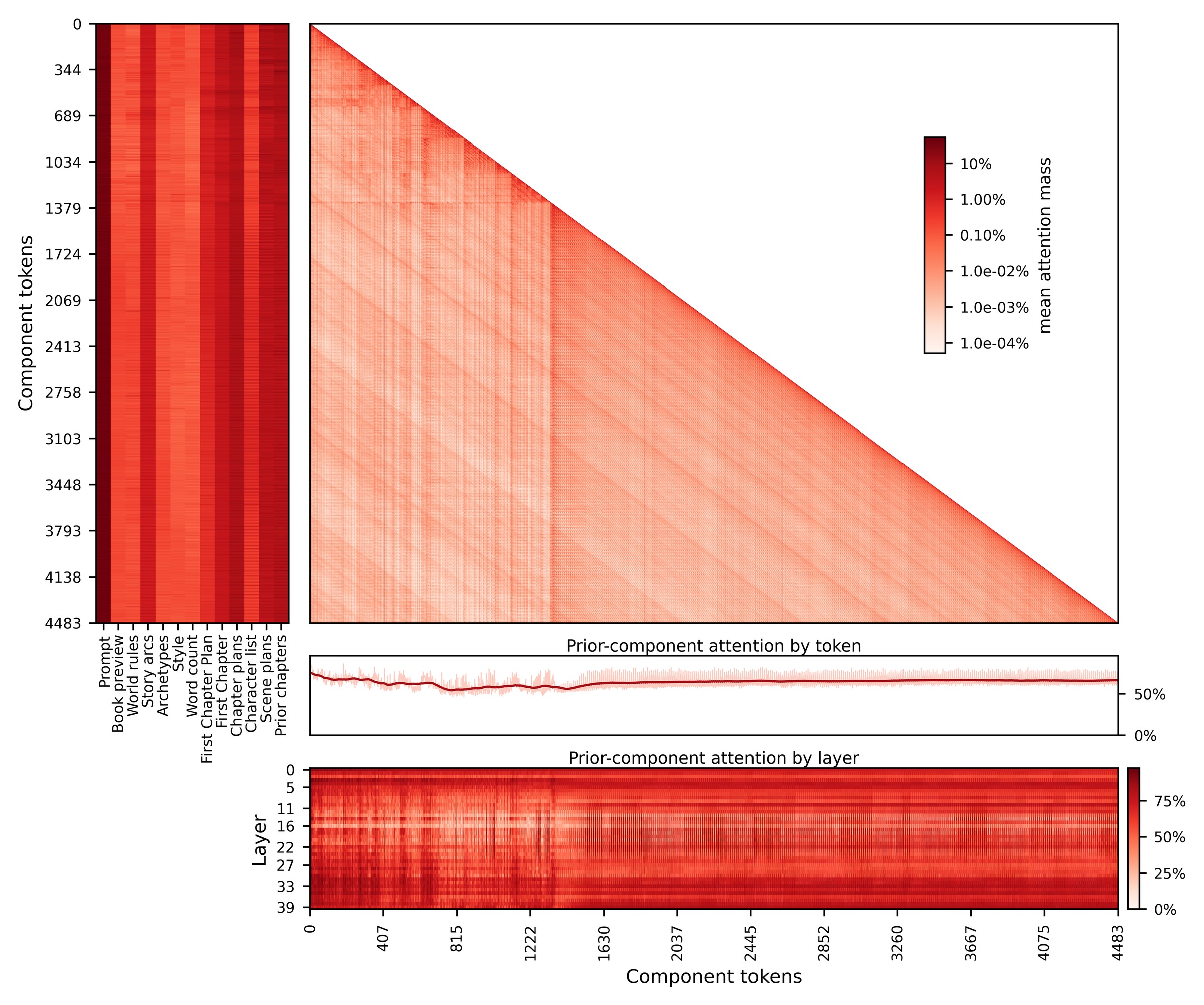}
\end{minipage}
\captionof{figure}{Attention patterns during a repetition failure. Left: compact prior-component view. Right: corresponding layer-wise view. The prior-component attention curve oscillates in phase with the repeated text pattern, indicating periodic shifts between local continuation and retrieval from earlier components.}%
\label{fig:appendix-failure-routing}
\vspace{-0.5em}
\end{center}

In contrast to the stable routing patterns observed in successful generations, the repetition failure exhibits a periodic alternation between local continuation and prior-component retrieval. Access to earlier scaffold components remains available throughout the failure. Instead, the routing pattern becomes coupled to the repetition loop, coinciding with the breakdown of normal chapter progression.

\subsection{Additional Visualization Variants}%
\label{subsec:appendix-duplicate-variants}

The figures below provide alternate views of the first-chapter and later-chapter attention patterns discussed in Section~\ref{sec:mechanistic}.

\begin{center}
\includegraphics[width=0.72\linewidth,height=0.28\textheight,keepaspectratio]{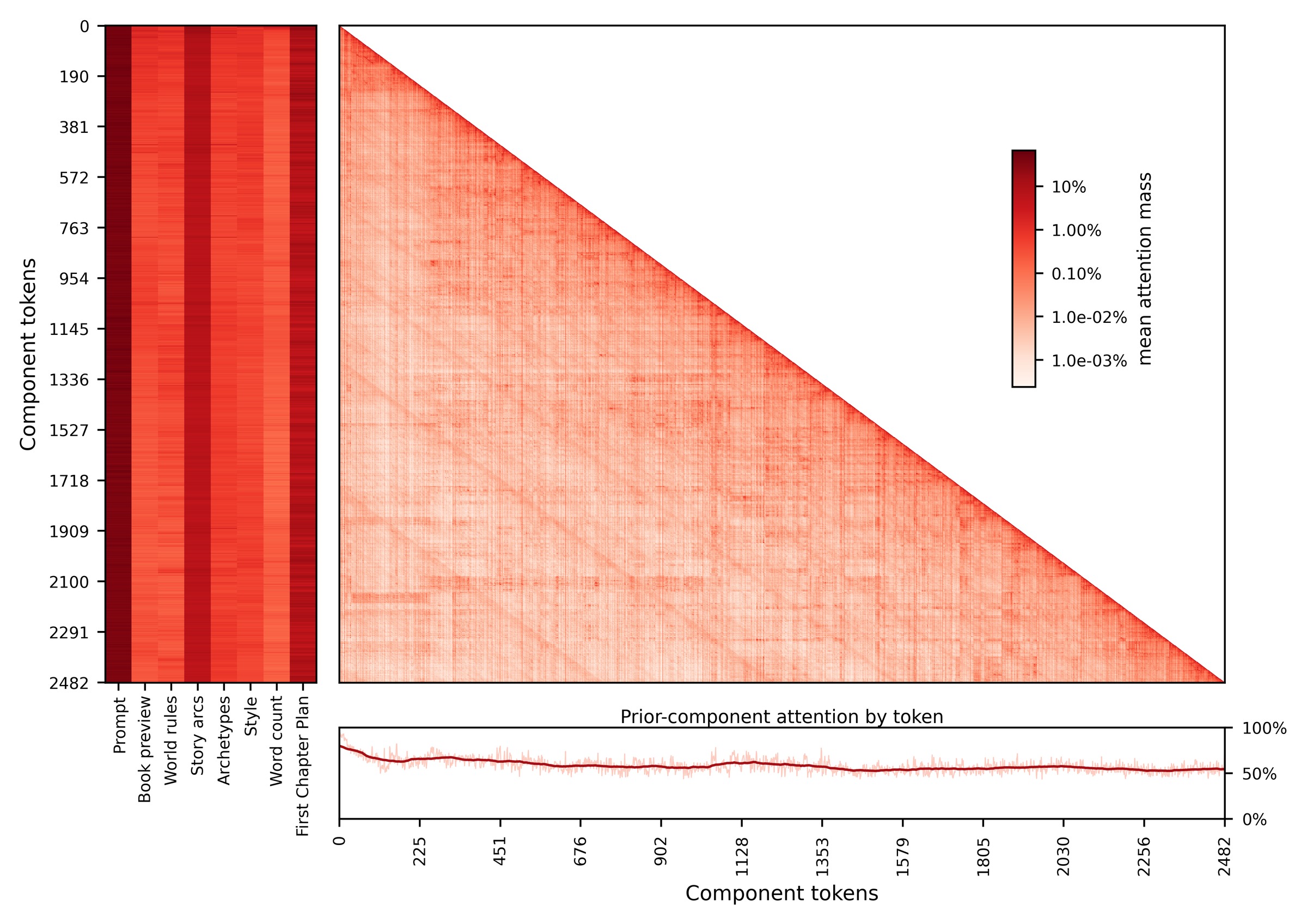}
\captionof{figure}{Compact first-chapter attention view. The corresponding layer-wise visualization in the main text contains the same prior-component attention trace together with a decomposition across model depth.}%
\label{fig:appendix-first-chapter-compact}
\vspace{-0.5em}
\end{center}

\begin{center}
\includegraphics[width=0.72\linewidth,height=0.30\textheight,keepaspectratio]{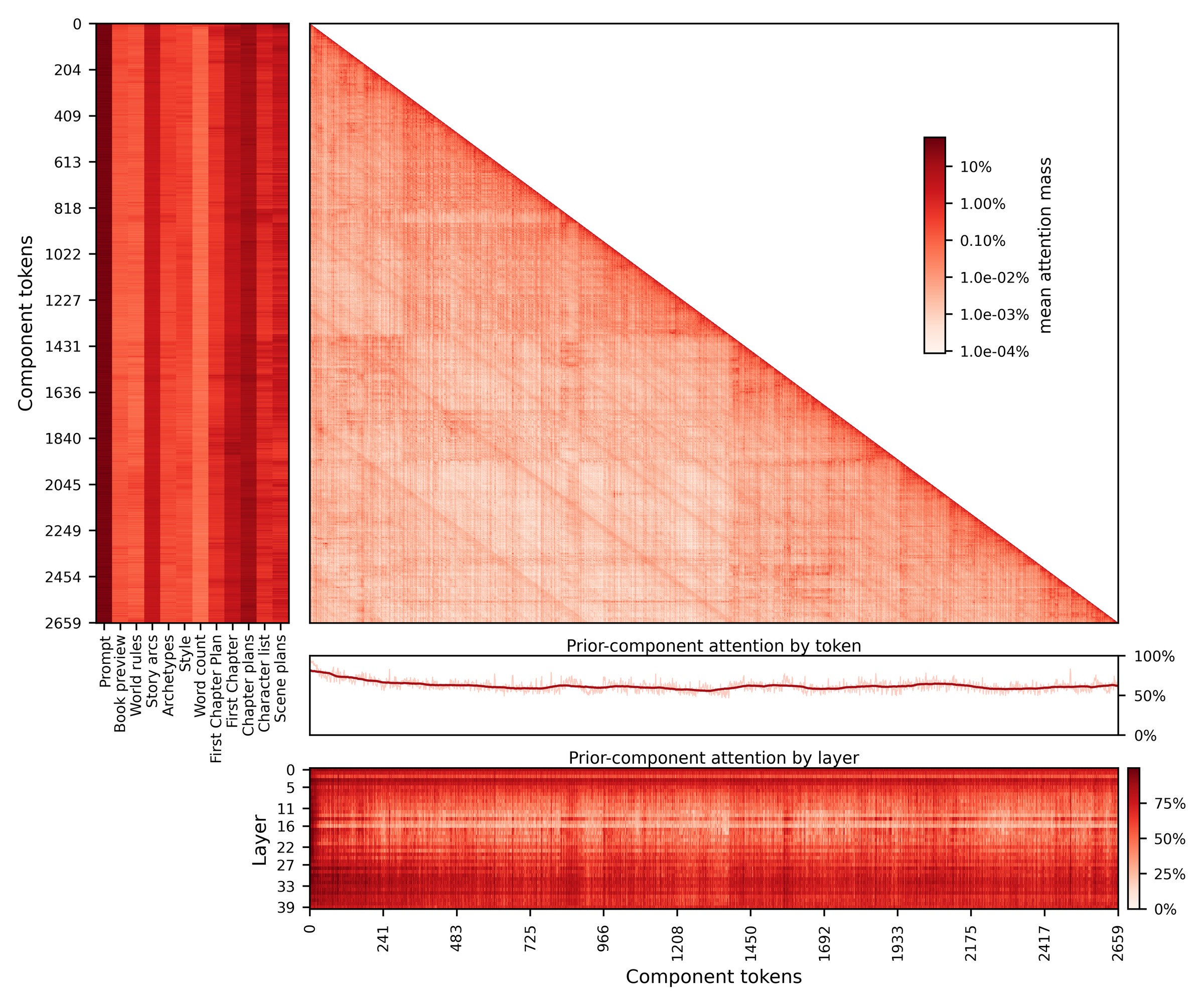}
\captionof{figure}{Layer-wise later-chapter attention view. The main text presents the compact variant because the continuity analysis focuses on the aggregate prior-component attention pattern.}%
\label{fig:appendix-later-chapter-layer}
\vspace{-0.5em}
\end{center}

\FloatBarrier{}

\end{document}